\documentclass[10pt,twocolumn,letterpaper]{article}
\pdfoutput=1

\usepackage[pagenumbers]{cvpr} 

\usepackage{graphicx}
\usepackage{algorithm}
\usepackage{algorithmic}
 \usepackage{xspace}

\usepackage{isl-shortcuts}

\usepackage[pagebackref,breaklinks,colorlinks]{hyperref}

\def\cvprPaperID{xxx} 
\def\confName{xxx}
\def\confYear{2022}

\newcommand{\footremember}[2]{%
    \thanks{#2}
    \newcounter{#1}
    \setcounter{#1}{\value{footnote}}%
}
\newcommand{\footrecall}[1]{%
    \footnotemark[\value{#1}]%
}

\begin{document}

\title{Unsupervised Contrastive Domain Adaptation for Semantic Segmentation}

\author{Feihu Zhang$^1$ \quad  Vladlen Koltun$^2$ \quad  Philip Torr$^1$ \quad  Ren\'{e} Ranftl$^2$\footremember{adv}{Equal advising.} \quad  Stephan R. Richter$^2$\footrecall{adv}\\
$^1$ University of Oxford \quad\qquad\qquad $^2$ Intel Labs}


\maketitle

\begin{abstract}
Semantic segmentation models struggle to generalize in the presence of domain shift. In this paper, we introduce contrastive learning for feature alignment in cross-domain adaptation. We assemble both in-domain contrastive pairs and cross-domain contrastive pairs to learn discriminative features that align across domains. Based on the resulting well-aligned feature representations we introduce a label expansion approach that is able to discover samples from hard classes during the adaptation process to further boost performance.
The proposed approach consistently outperforms state-of-the-art methods for domain adaptation.
It achieves 60.2\% mIoU on the Cityscapes dataset when training on the synthetic GTA5 dataset together with unlabeled Cityscapes images.
\end{abstract}

\section{Introduction}
\label{sec:intro}

\begin{figure}[t]
	\begin{subfigure}{0.49\linewidth}
		\includegraphics[width=1\linewidth,height=0.6\linewidth]{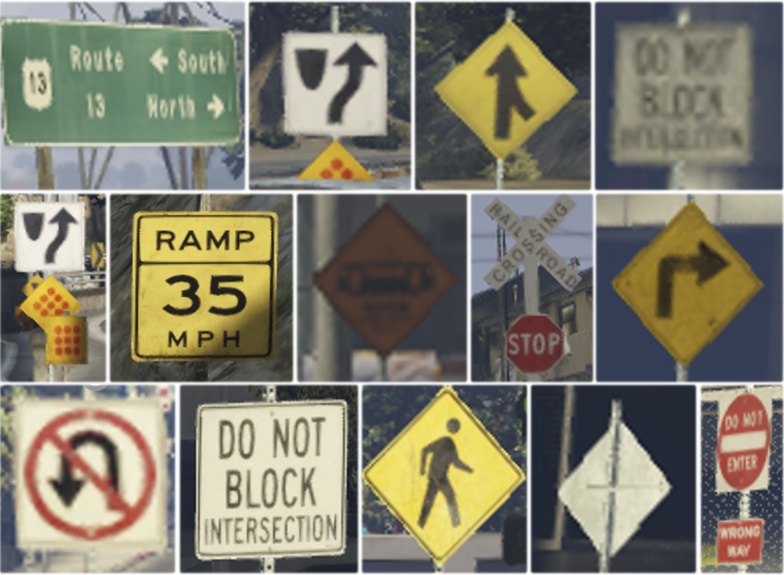}
        \caption{GTA5 signs}
        \label{subfig:gta_sign}
	\end{subfigure}
	\begin{subfigure}{0.49\linewidth}
		\includegraphics[width=1\linewidth,height=0.6\linewidth]{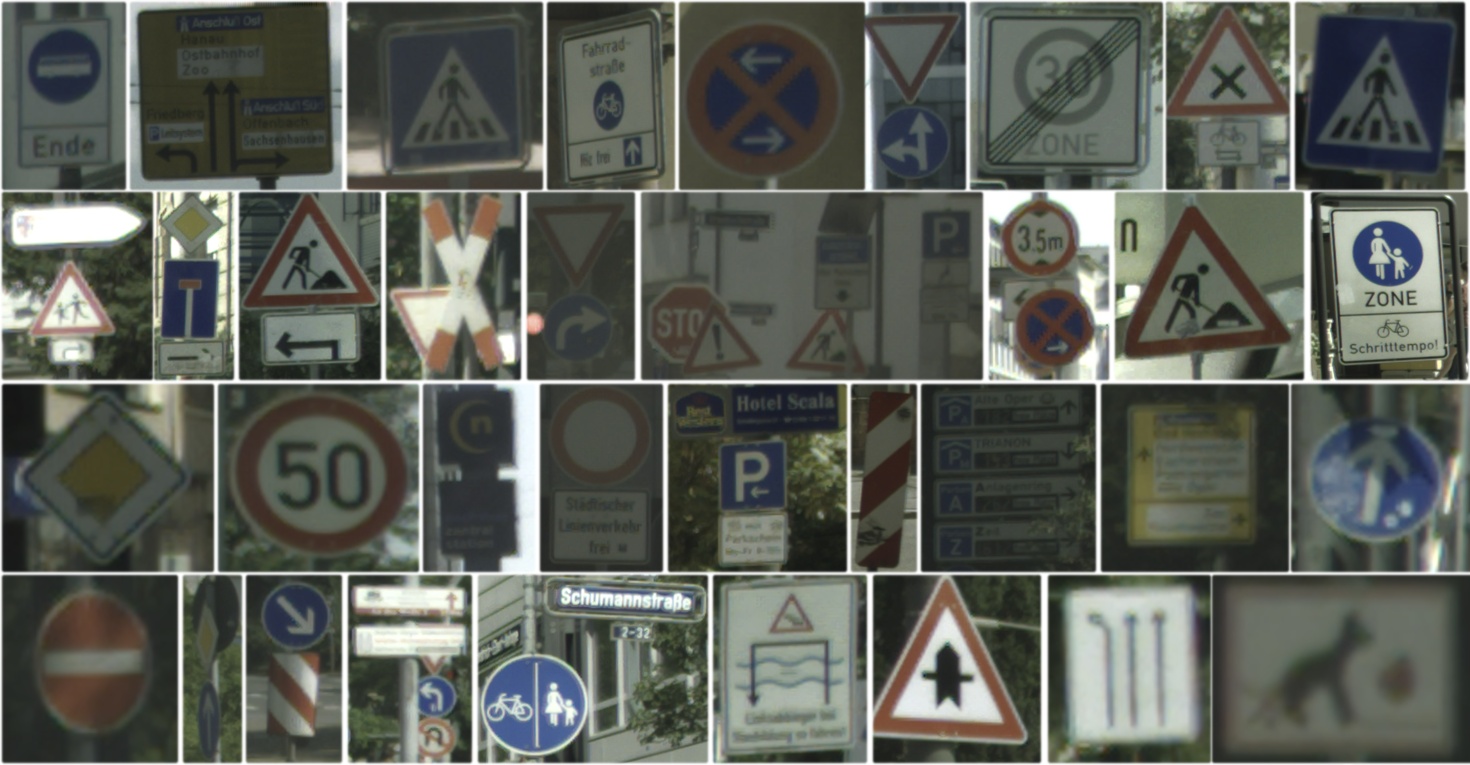}
		\caption{Cityscapes signs}
	\label{subfig:city_sign}
	\end{subfigure}\\
	\begin{subfigure}{0.49\linewidth}
		\includegraphics[height=0.5\linewidth,width=1\linewidth]{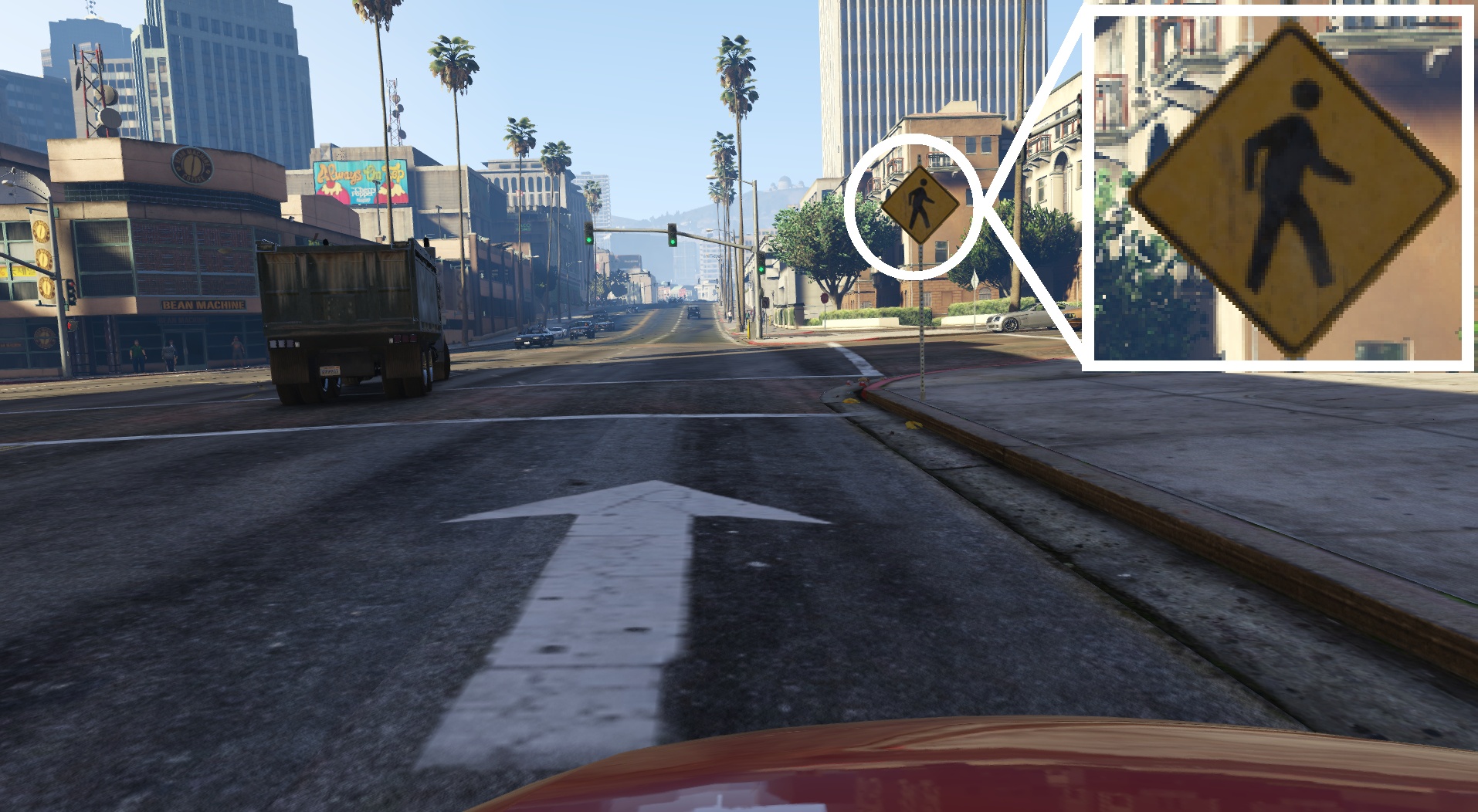}
		\caption{GTA5 scene}
		\label{subfig:gta_scene}
	\end{subfigure}
	\begin{subfigure}{0.49\linewidth}
		\includegraphics[height=0.5\linewidth,width=1\linewidth]{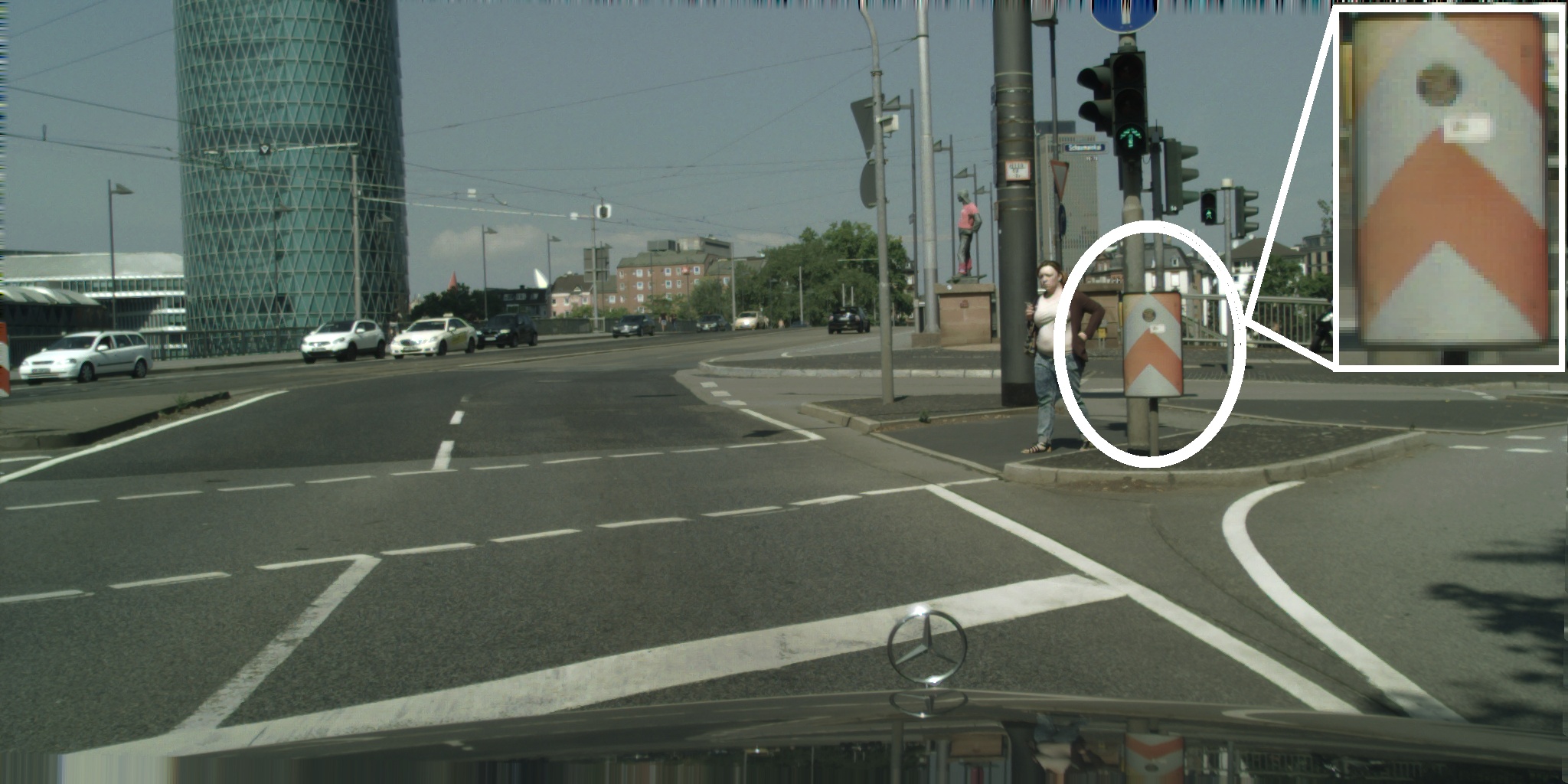}
        \caption{Cityscapes scene}
        \label{subfig:city_scene}
	\end{subfigure}\\
	\begin{subfigure}{0.49\linewidth}
		\includegraphics[height=0.5\linewidth,width=1\linewidth]{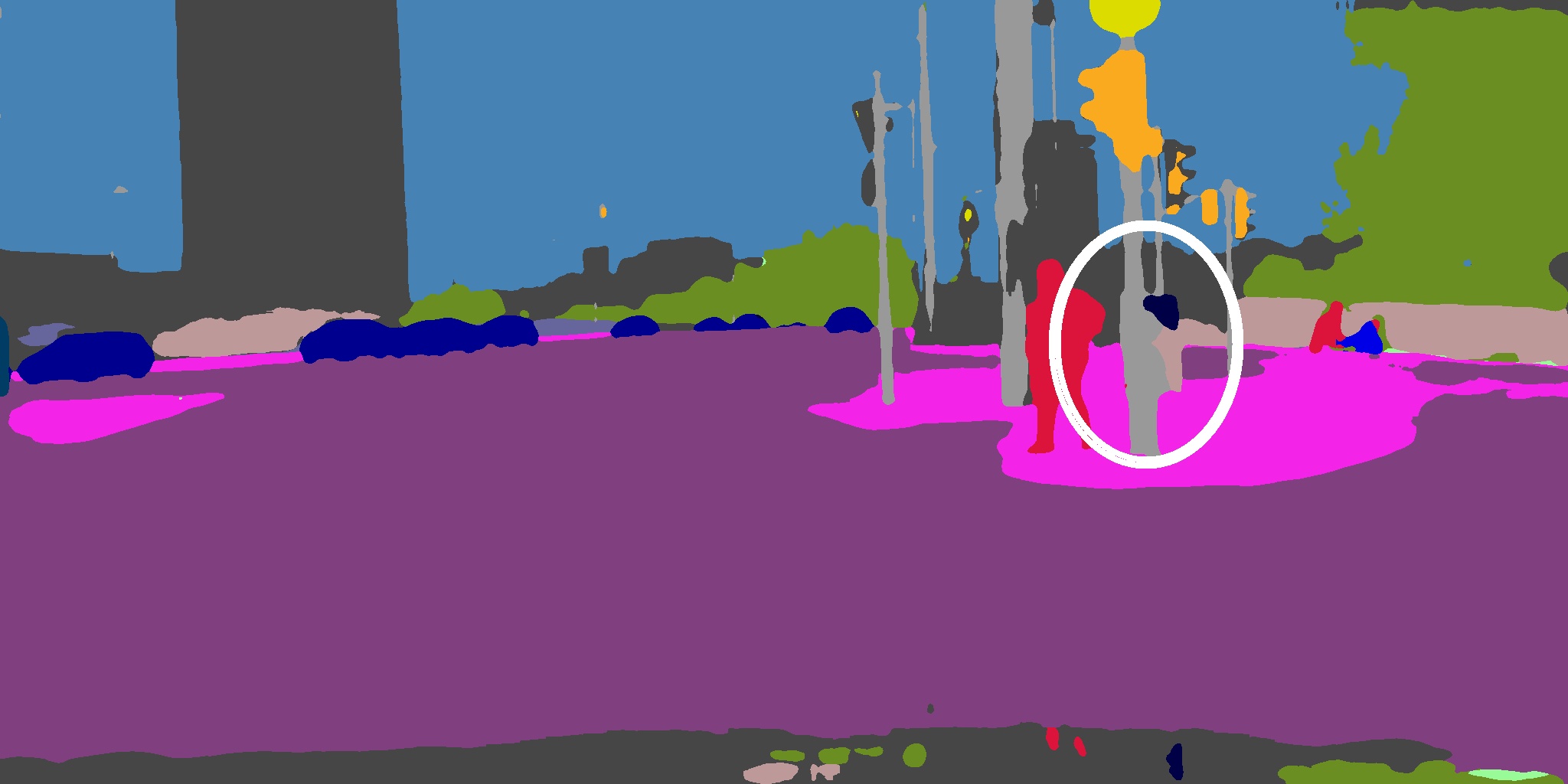}
        \caption{ Result of \cite{Ma2021:CVPR}}
        \label{subfig:sota_result}
	\end{subfigure}
	\begin{subfigure}{0.49\linewidth}
		\includegraphics[height=0.5\linewidth,width=1\linewidth]{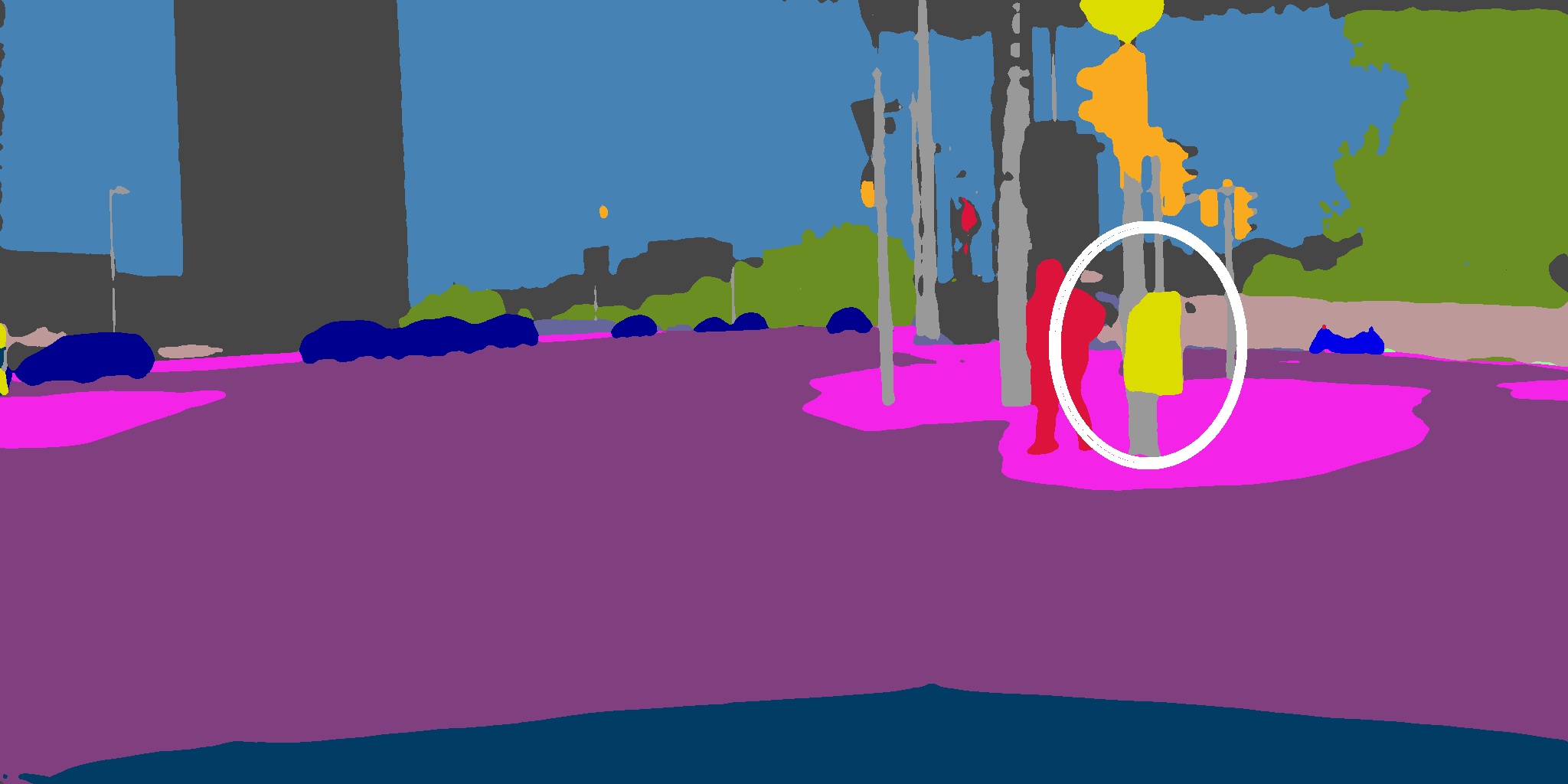}
        \caption{Our result}
		\label{subfig:our_result}
	\end{subfigure}
	\vspace{-0.5em}
	\caption{\small Problem illustration and example result.  (a) Traffic signs in the GTA5\cite{gta5} dataset. (b) Traffic signs in the Cityscapes dataset~\cite{cordts2016cityscapes}. Strong domain shift within a semantic class can lead to poor performance when
	transferring a model.
	(c-d) Domain shift can also include low-level style and overall scene layout. (e)~Results of a state-of-the-art domain adaptation method\cite{Ma2021:CVPR}. (f)~Result of our approach. Our method learned to recognize a traffic sign that didn't appear in source dataset.}
	\label{fig:teaser} 
	\vspace{-0.5em}
\end{figure}

Semantic segmentation is the task of dense per-pixel classification of the content of an input image and is one of the fundamental problems of computer vision. Deep networks have achieved tremendous advances towards solving this problem when a large amount of labeled data is available. In the supervised training paradigm it is crucial that the labeled training data is sufficiently similar to the data that will be encountered in the target application. When
this fundamental assumption is violated, heavy performance drops are frequently observed, rendering  such models impractical for deployment in real-world applications. Unsupervised domain adaptation aims at mitigating these performance drops by augmenting existing labeled training data from a source domain with unlabeled images from the target domain.

The underlying issues resulting in a performance loss when switching domains
are loosely summarized by the term \emph{domain shift}. It includes apparent low-level difference between images 
that might be due to different camera setups or weather conditions. Several existing works apply style transfer networks with adversarial training ~\cite{dagan3,Wang2020:CVPR,BDL,Huang2018:ECCV} to align the low-level image statistics of the source and target domain in order to improve transfer. A second, more subtle, source of domain shift is the distribution of semantic content in the images. This issue includes, but is not limited to, a mismatch between the actual objects depicted in the source and the target domain. While it is apparent that a model cannot learn to segment an object if it was never encountered in the labeled source data, more subtle issues can lead to poor performance in the target domain. An example is \Figs~\ref{fig:teaser} (a) and (b), where we show various traffic signs in two different domains. It is apparent that there is a strong class-specific domain shift that can not be mitigated by low-level style transfer alone. 

An underlying assumption of many domain adaptation approaches is that training on different datasets (domains) will lead to different visual representations, even if the set of classification labels is the same. Thus, many works explicitly target the alignment of feature representations between domains. What makes this particularly challenging is the lack of labeled data for the target domain, and hence no directly available supervision for learning a representation in the target domain to begin with. 

In this work, we introduce contrastive learning for unsupervised adaptation of semantic segmentation models across domains. We build both in-domain contrastive pairs  (in the source and the target domain) as well as cross-domain contrastive pairs (from source to target domain) to improve feature alignment between domains while simultaneously ensuring a highly discriminative representation.
We propose a student-teacher approach that iteratively updates a set of pseudo labels in the target domain, which allows us 
to draw corresponding samples from both domains.
Contrastive learning allows us to naturally balance frequent easy classes and infrequent hard classes by controlling the sampling of contrastive pairs. We further propose a feature-based pseudo-label expansion strategy to discover more pixels that correspond to hard classes. 


Our approach consistently outperforms the state of the art for domain adaptation across multiple datasets.  
It achieves 60.2\% mIoU on the Cityscapes validation set when trained on the synthetic GTA5 dataset~\cite{gta5} together with unlabeled Cityscapes training images.  We observe a state-of-the-art 56.5\% mIoU when leveraging the synthetic SYNTHIA dataset~\cite{synthia} in the same setting. Our experiments show that our approach significantly improves performance particularly on hard classes that have few labeled pixels in the source data or that potentially undergo a strong domain shift between source and target domain.

\section{Related Work}

\mypara{Adversarial training.}
In an adversarial setup, a discriminator is trained to distinguish images~\cite{Chen2019:CVPR,Choi2019:ICCV,Hoffman2018:ICLR, Huang2018:ECCV, Zhu2017:ICCV}, intermediate representations~\cite{Chen2017:ICCV,Chen2019:CVPR,Wang2020:CVPR, Wang2020:ECCV}, or predicted labels~\cite{Tsai2018:CVPR,Tsai2019:ICCV} from different domains. The discriminator then provides a supervisory signal for aligning the distributions of its inputs and thus the different domains.
For more fine-grained supervision, one line of work trained the discriminator to additionally distinguish individual classes~\cite{Chen2017:ICCV, Wang2020:ECCV}.
To leverage local spatial invariances, ROAD aligns fixed image regions~\cite{Chen2018:CVPR}, and Tsai~\etal's work predicts patches mined from the source domain~\cite{Tsai2019:ICCV}.
SIM aligns the feature representation from stuff and things of the target domain to specific samples in the source domain~\cite{Wang2020:CVPR}. ADVENT finds that the entropy of predictions is higher on the target domain and introduces an entropy minimization loss~\cite{Vu2019:CVPR}. This can lead to exploding gradients for confident predictions, which can be addressed via a maximum square loss~\cite{Chen2019:ICCV}. CrDoCo trains separate networks in the task domains and leverages a cross-domain consistency~\cite{Chen2019:CVPR}.

Kim~\etal~ \cite{Musto2020:BMVC} apply random styles using style transfer to learn a texture-invariant representation~\cite{Kim2020:CVPR}. Yang~\etal propose to reconstruct images from inferred label maps~\cite{Yang2020:ECCV}.
FDA transfers low-frequency image components~\cite{Yang2020:CVPRa} and PCEDA introduces a phase-consistency loss to retain semantic content~\cite{Yang2020:CVPRb}.
DISE allows for domain-specific, non-transferable features~\cite{Chang2019:CVPR} and CSCL learns non-transferable content via reinforcement learning~\cite{Dong2020:ECCV}.
Zhang \etal propose several domain-invariant regularizers to improve the consistency of predictions~\cite{Zhang2020:CVPR}. 
Yang~\etal introduce an attention mechanism to learn transferable contextual relations across domains~\cite{Yang2021:WACV}.
\begin{figure*}[t]
    \centering
    \includegraphics[width=0.73\linewidth]{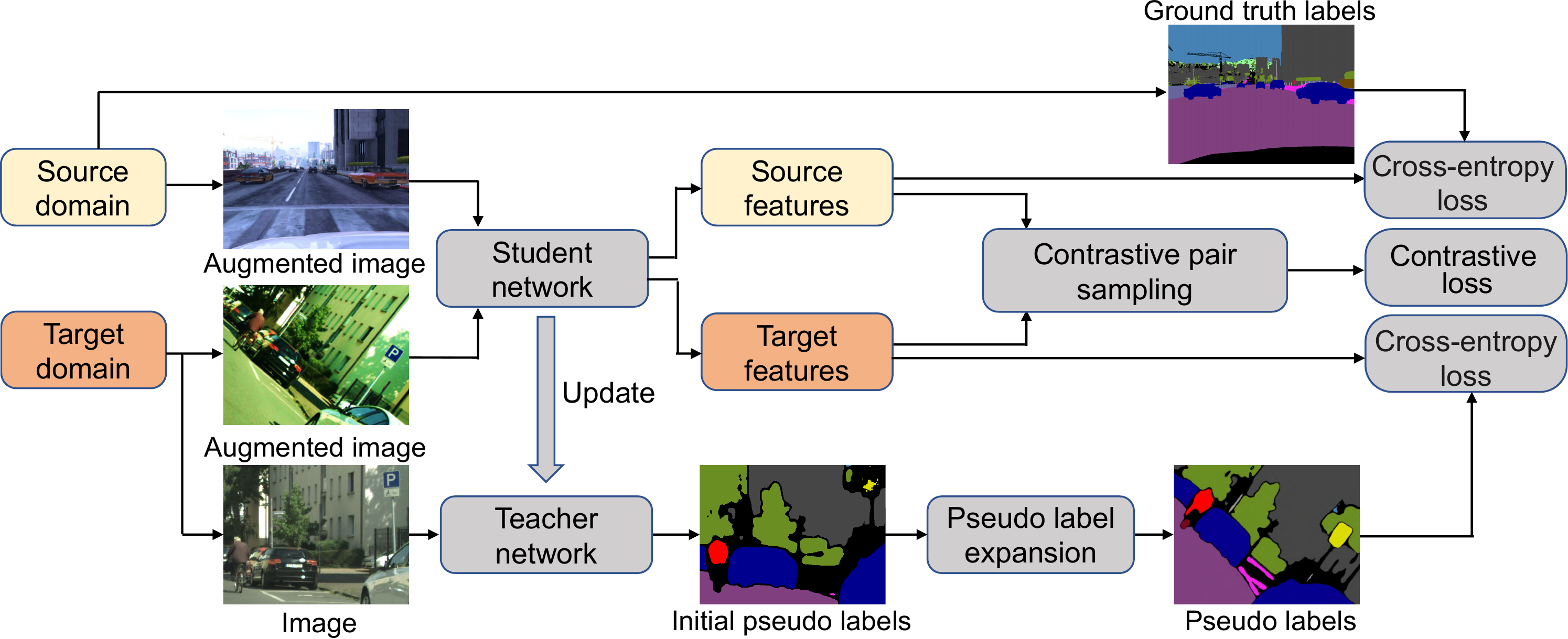}
    \vspace{-0.5em}
    \caption{\small Overview of self-training stage. We train a semantic segmentation network (student) with augmented images from source and target domains. It is supervised via ground truth labels from the source domain and pseudo labels generated by a teacher network from unmodified target domain images. We improve the pseudo labels via a novel expansion mechanism. Further details in the text.}
    \label{fig:overview}
\end{figure*}

%

\mypara{Self-training.}
CBST introduced self-training for domain adaptation~\cite{Tang2012:NIPS} to semantic segmentation~\cite{Zou2018:ECCV}. State-of-the-art self-training relies on pseudo labels for the target domain, which are generated by earlier versions (a \emph{teacher}) of the network (the \emph{student}) being trained.
To balance the selection of samples to harder cases in pseudo labels, adaptive thresholds are commonly used for classes~\cite{Araslanov2021:CVPR, Subhani2020:ECCV} or even instances~\cite{Mei2020:ECCV}. 
The predicted label distributions are frequently sharpened during training~\cite{Berthelot2019:NeurIPS}.
Subhani and Ali propose to leverage the expected consistency of predictions across scales to generate pseudo labels~\cite{Subhani2020:ECCV}. 
SAC extends this idea to a broader range of image transformations~\cite{Araslanov2021:CVPR}, and TGCF-DA  as well as BiSIDA to different stylizations~\cite{Choi2019:ICCV,Wang2021:AAAI}.
Another line of work leverages spatially related samples~\cite{Kang2020:NeurIPS,Lv2020}.
IAST smooths pseudo labels in high-confidence regions and sharpens them for low-confidence regions~\cite{Mei2020:ECCV}.
Zheng and Yang exploit auxiliary classifiers to obtain a confidence measure for the pseudo labels~\cite{Zheng2021:IJCV}.
ProDA weights pseudo labels by their distance to prototypical features to reduce the influence of outliers~\cite{Zhang2021:CVPR}. Ma~\etal propose a triplet loss to align average feature representations for different categories~\cite{Ma2021:CVPR}. 
While we also aim to align feature representations, we formulate the alignment as a more powerful contrastive objective.

\mypara{Contrastive training.}
After impressive performance demonstrations for image classification~\cite{He2020:CVPR,Chen2020:arxiv,Chen2020:ICML,Chen2020:NeurIPS}, contrastive representation learning has found its way to semantic segmentation as an alternative to the ubiquitous pre-training on ImageNet~\cite{Zhang2020:NeurIPS,Zhao2021:ICCV} as well as a supporting training objective~\cite{Wei2020:arxiv}.
A number of recent works have adapted the contrastive learning framework to semantic segmentation, but restricted the sampling of positive correspondences to the same image~\cite{Chaitanya2020:NeurIPS, Chen2021:arxiv, Wang2021:CVPRa, Xie2021:CVPR, Xiao2021:ICCV, Zoph2020:NeurIPS, Van2021:ICCV}.

Wang \etal exploited ground truth labels to construct contrastive pairs for fully supervised semantic segmentation~\cite{Wang2021:ICCV}.
SSC retains high-quality features to build contrastive pairs in a semi-supervised setting~\cite{Alonso2021:ICCV}. 
For domain adaptation, Liu~\etal~\cite{Liu2021:arxiv} identify patches to pair via simplified spatial pyramid matching. ASSUDA uses a contrastive loss to minimize the distance of predictions to adversarial samples~\cite{Yang2021:ICCV}. Masden~\etal construct contrastive pairs from mean feature representations in the source and target domain to features averaged over target images~\cite{Marsden2021:arxiv}.
Our work employs three complementary strategies for sampling contrastive pairs, which exploit supervision in the source domain, compress feature representations, and align them across domains. 
\begin{figure}
    \centering
    \includegraphics[width=0.95\linewidth]{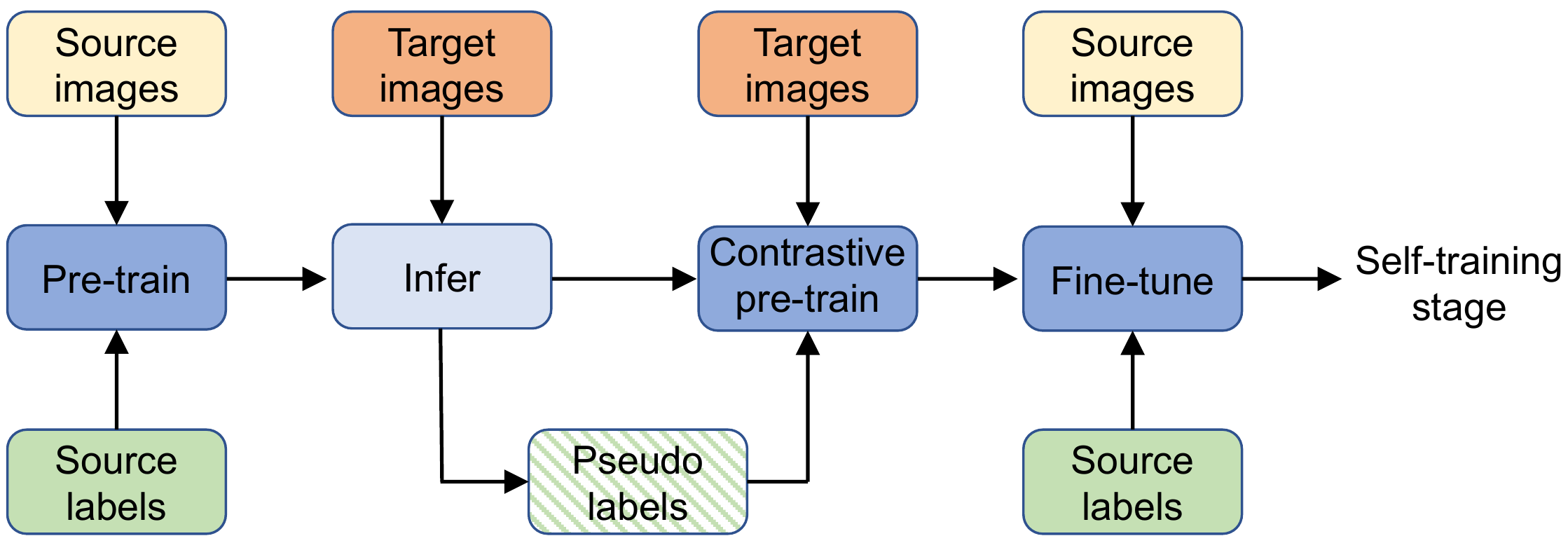}
    \vspace{-0.6em}
    \caption{Pre-training stage. We start by training on the source domain with direct supervision and obtain pseudo labels for the target domain to guide contrastive training. Finally, we fine-tune again on the source domain before progressing to self-training.}
    \label{fig:pretraining}
\end{figure}

\section{Overview}
Our approach consists of a pre-training stage (as illustrated in \Fig~\ref{fig:pretraining}) and a self-training stage (\Fig~\ref{fig:overview}).  

We pre-train our segmentation network in the pre-training stage (\Sec~\ref{sec:pretraining}), and the pre-trained network is then used to initialize the teacher and student networks in the self-training stage (\Sec~\ref{sec:self_training}). 
Here, the teacher network generates pseudo labels as supervision for the student network. 
The student network is trained via a standard cross-entropy loss for direct supervision and a novel contrastive training objective (\Sec~\ref{sec:contrastive_objective}). Furthermore, we propose a label expansion mechanism (\Sec~\ref{sec:label_expansion}) to extend the set of samples for which pseudo labels are generated.

\section{Pre-training}\label{sec:pretraining}
Before training, we first translate images from the source domain to the style of the target domain via an off-the-shelf image-to-image translation network~\cite{Huang2018:ECCV}.
This translation step helps bootstrapping our method as it reduces the visual difference between domains. Note that we only apply the translation during pre-training and later operate directly on unmodified images from the source domain, which we augment for robustness.
Figure~\ref{fig:pretraining} illustrates the pre-training phase. 
We start by training a semantic segmentation network via direct supervision on the translated images and source domain labels. 
%
Training the network on the source domain with direct supervision serves as a strong initialization. Since the images have been translated to the target domain, the initially learned visual representations are already partially aligned with the target domain.
Next, we train the network solely on the target domain with a contrastive objective. 
The same objective is used for both pre-training and self-training. We defer a detailed discussion to Section~\ref{sec:contrastive_objective}.
\begin{figure*}[t]
    \centering
    \begin{tabular}{@{\hspace{0.03\linewidth}}c@{\hspace{0.035\linewidth}}c@{\hspace{0.035\linewidth}}c@{\hspace{0.03\linewidth}}}
        \multicolumn{3}{l}{ \includegraphics[height=0.03\linewidth]{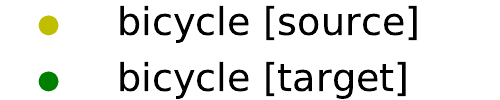}
        \includegraphics[height=0.03\linewidth]{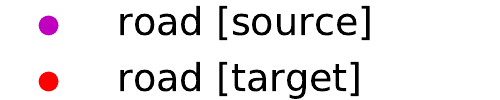}
        \includegraphics[height=0.03\linewidth]{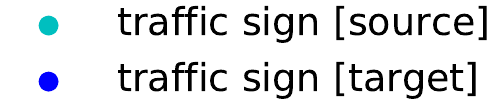}
        }\\
        \includegraphics[width=0.28\linewidth]{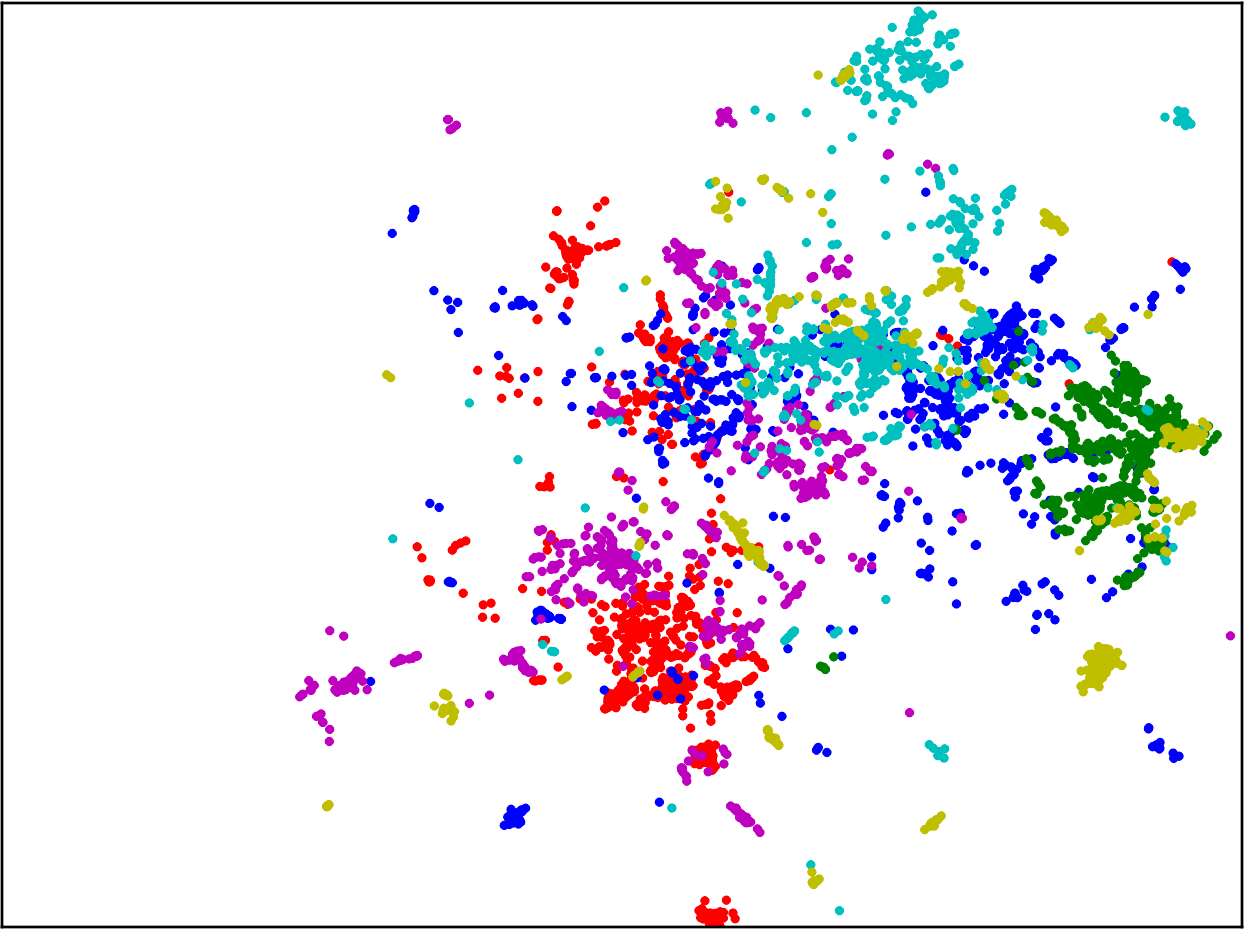}&
        \includegraphics[width=0.28\linewidth]{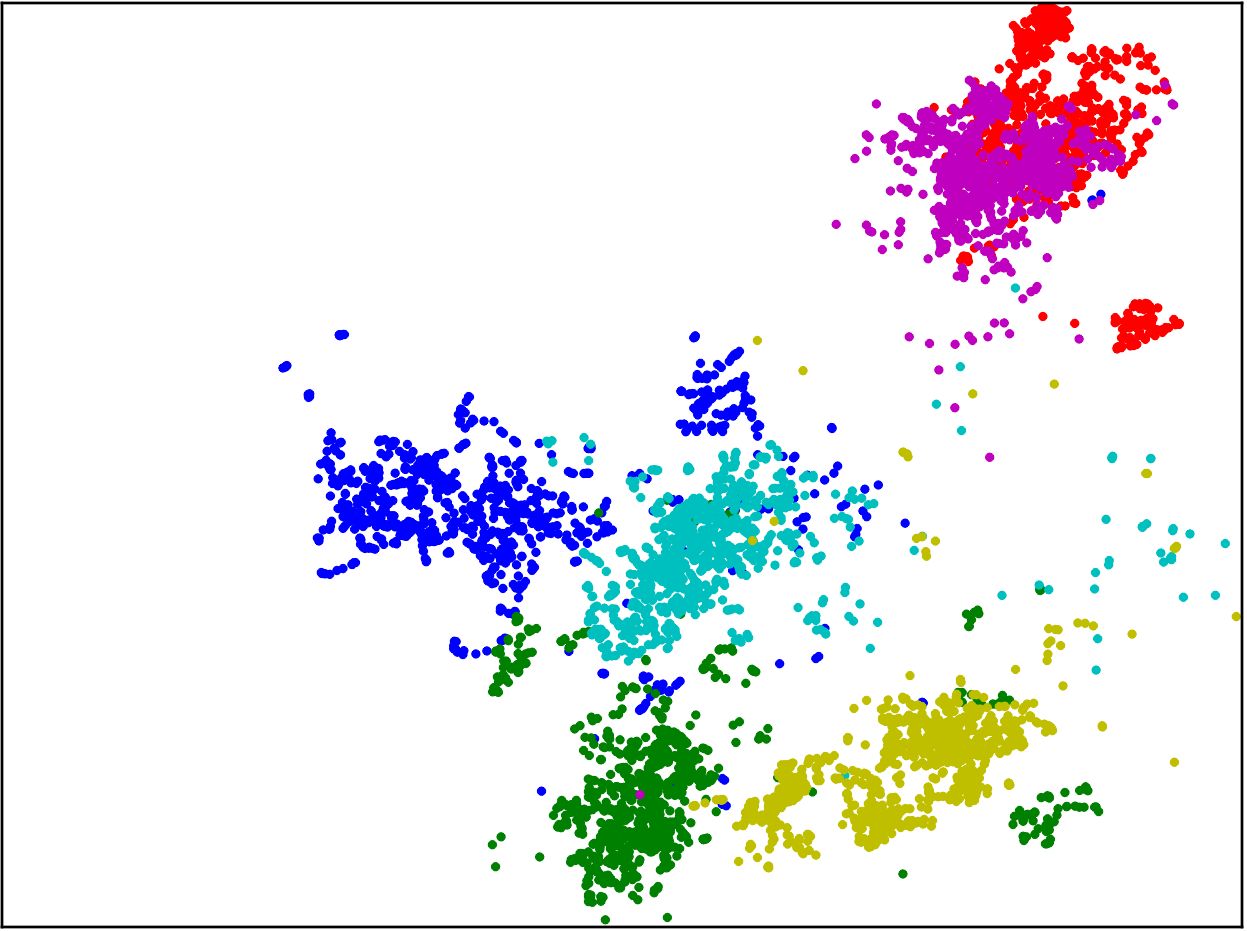}&
        \includegraphics[width=0.28\linewidth]{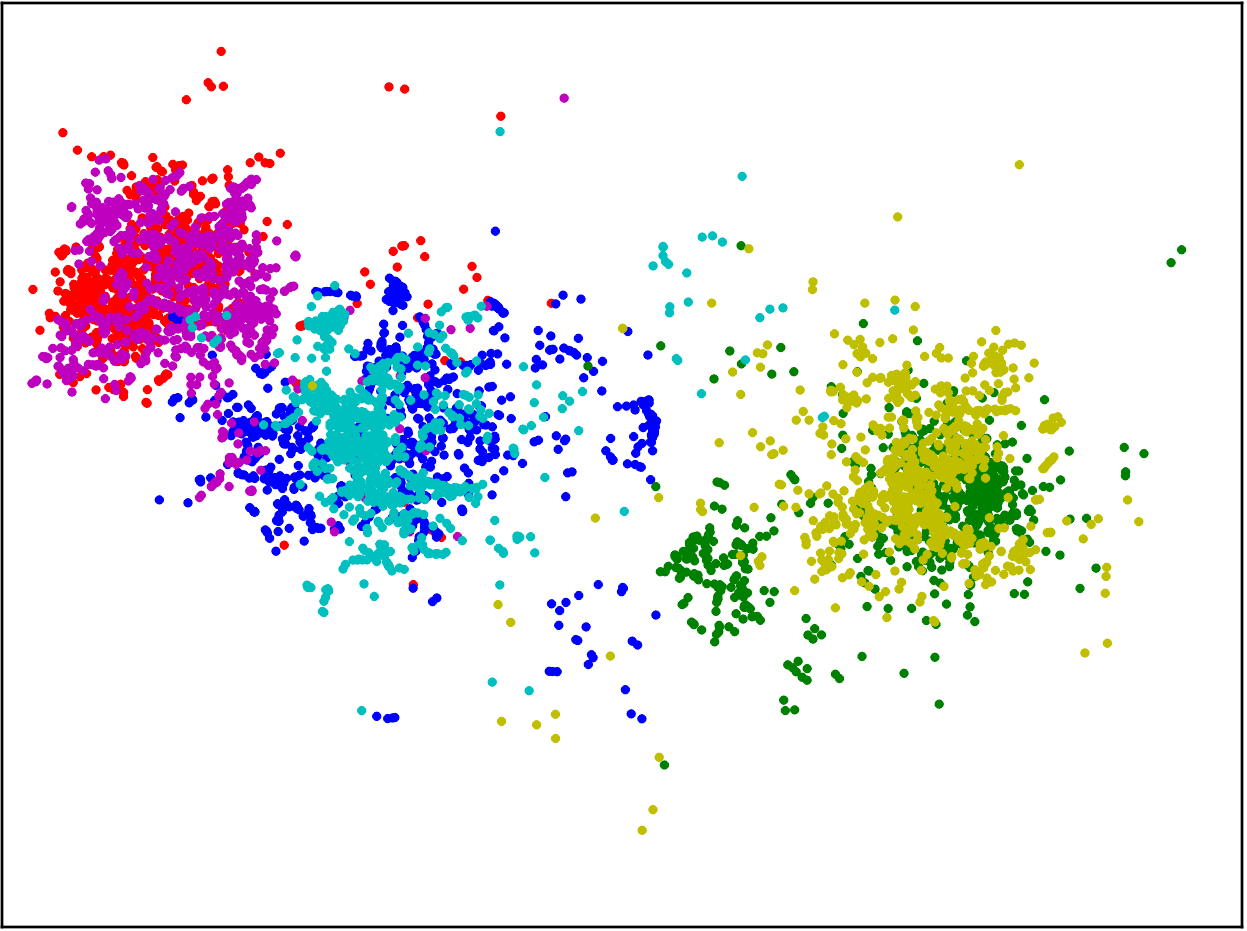}\\
        \footnotesize a) w/o contrastive learning & \footnotesize b) only in-domain   & \footnotesize c) in-domain \& cross-domain\\
    \end{tabular}
    \vspace{-0.5em}
    \caption{Visualization of representation compression and alignment. We visualize the learned feature space for a subset of classes from the source (GTA5) and target (Cityscapes) domains with UMAP~\cite{mcinnes2020umap}. a) Without any contrastive objective, representations for individual classes overlap. b) Applying a contrastive objective within domains leads to more discriminative representations for individual classes, but representations are misaligned
    between domains. c) Adding a cross-domain contrastive objective aligns the classes from different domains while keeping the representation discriminative.} 
    \label{fig:feature_align}
\end{figure*}
After adapting the visual representations, we fine-tune the network again on translated images of the source domain and ground truth labels. For this, we fix the backbone  network and train only the segmentation head with a reduced learning rate of 0.001 for 5 epochs.
%

\section{Self-training}\label{sec:self_training}
%
Following the self-training paradigm~\cite{Tarvainen2017:NIPS}, 
we employ 
a teacher network for generating pseudo labels. The teacher network is initialized with the same weights as the student network. While we update the weights of the student network in every training iteration, the weights of the teacher network are only updated every 200 iterations by copying from the student network.

\subsection{Pseudo label generation}\label{sec:pseudo_labels}
To generate pseudo labels for images from the target domain, we randomly sample $713 \times 713$ crops (without augmentation) from these images and pass them through the teacher network. The same crop is augmented before it is ingested by the student network (see~\Sec~\ref{sec:details}).
As typical for semantic segmentation networks, our teacher network predicts class probabilities. As in prior work, we take the most confident prediction per pixel as pseudo label and employ the output probability as confidence measure~\cite{Zhang2021:CVPR,Ma2021:CVPR}.
We follow Ma~\etal~\cite{Ma2021:CVPR} and ignore predictions below an adaptively computed, class-dependent confidence threshold (detailed in the supplement). To increase the robustness of the generated pseudo labels, we 
apply the teacher network at multiple scales (0.5,0.75,1.0,1.25,1.5,1.75) and average predictions over scales.
Besides supervising the student network with the pseudo labels directly via a cross-entropy loss, we further use them to formulate a contrastive objective.

\subsection{Contrastive objective}\label{sec:contrastive_objective}
Contrastive learning aims to learn a representation space in which similar entities cluster closely together, while simultaneously pushing apart dissimilar entities~ \cite{Chen2020:ICML,He2020:CVPR,Chen2020:arxiv}. Importantly, the notion of what constitutes similarity between entities is entirely defined by data and can be steered by an appropriate selection of similar (positive) pairs of samples together with dissimilar (negative) pairs.

\begin{figure}[t]
    \centering
    \begin{tabular}{@{}c@{}c@{}}
        \multicolumn{2}{l}{\includegraphics[width=0.6\linewidth]{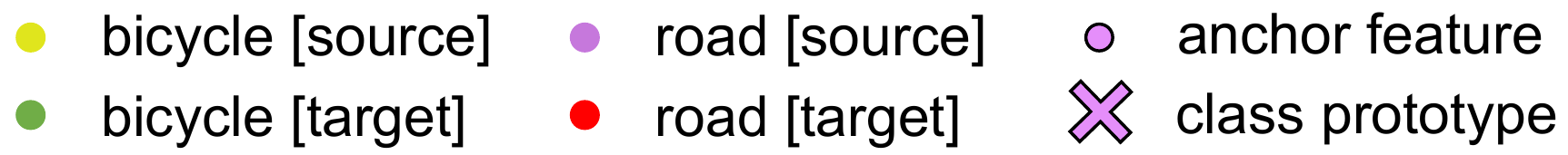}}\\
        \includegraphics[width=0.47\linewidth]{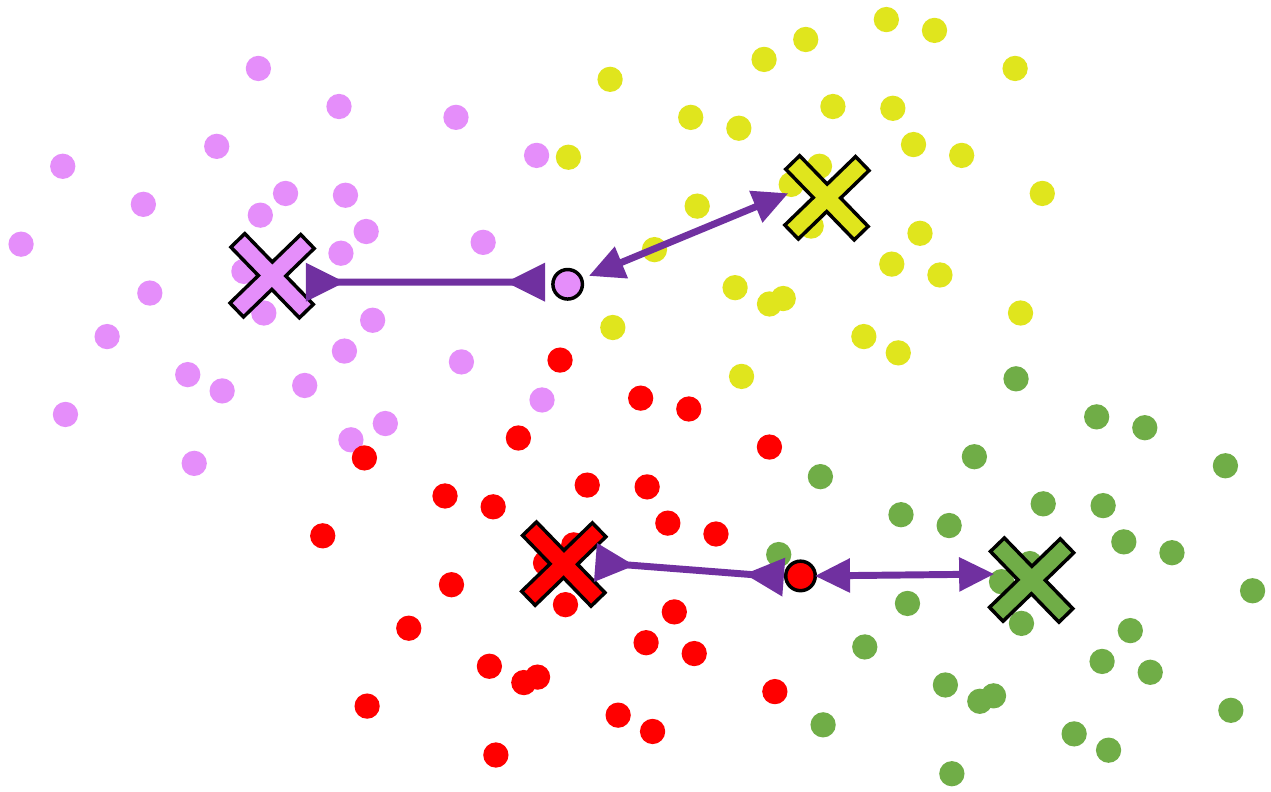} &     
        \includegraphics[width=0.47\linewidth]{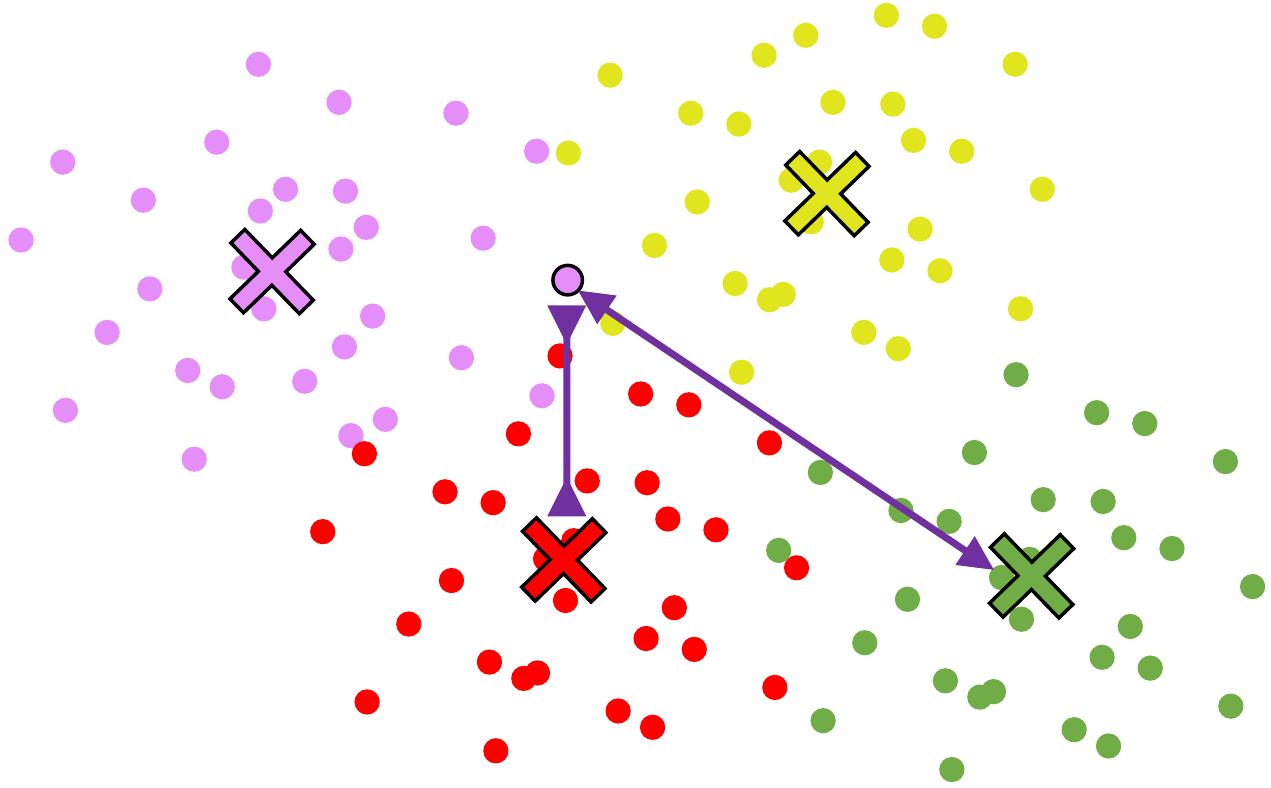} \\
        \footnotesize a) within domain &
        \footnotesize b) across domains \\
    \end{tabular}
    \vspace{-0.5em}
    \caption{We employ different sampling strategies for obtaining corresponding contrastive pairs. Starting from a sampled anchor feature, we match class prototypes within domain to compress representations (a), and across domains to align them (b). Figure~\ref{fig:feature_align} illustrates the effect on the representations learned.}
    \label{fig:contrasts}
    \vspace{-0.5em}
\end{figure}

We follow SimCLR~\cite{Chen2020:ICML} and add a 2-layer MLP head 
to our student network to learn a 128-dimensional latent feature representation $f$ per pixel. We leverage the contrastive loss proposed by SupContrast~\cite{Khosla2020:NeurIPS}:
\begin{align}
    \mathcal{L}_{ctr}(\bm{v},\bm{v}^+)=& -\log\frac{\exp(\frac{\bm{v}\cdot\bm{v}^+}{\tau})}{\exp(\frac{\bm{v}\cdot\bm{v}^+}{\tau}) + \sum\limits_{\bm{v}^-}\exp(\frac{\bm{v}\cdot\bm{v}^-}{\tau})},
    \label{EQ:contrast}
\end{align}
which maximizes similarity between positive feature pairs ($\bm{v}$/$\bm{v}^+$, \ie of the same class), and minimizes it between negative pairs ($\bm{v}$/$\bm{v}^-$, \ie of different classes). We set the temperature $\tau = 0.07$ in our experiments. 
\begin{figure*}[t]
    \centering
    \begin{tabular}{@{}c@{\hspace{0.005\linewidth}}c@{\hspace{0.005\linewidth}}c@{\hspace{0.005\linewidth}}c@{}}
         \includegraphics[width=0.241\linewidth, trim={35cm 10cm 10cm 3cm}, clip]{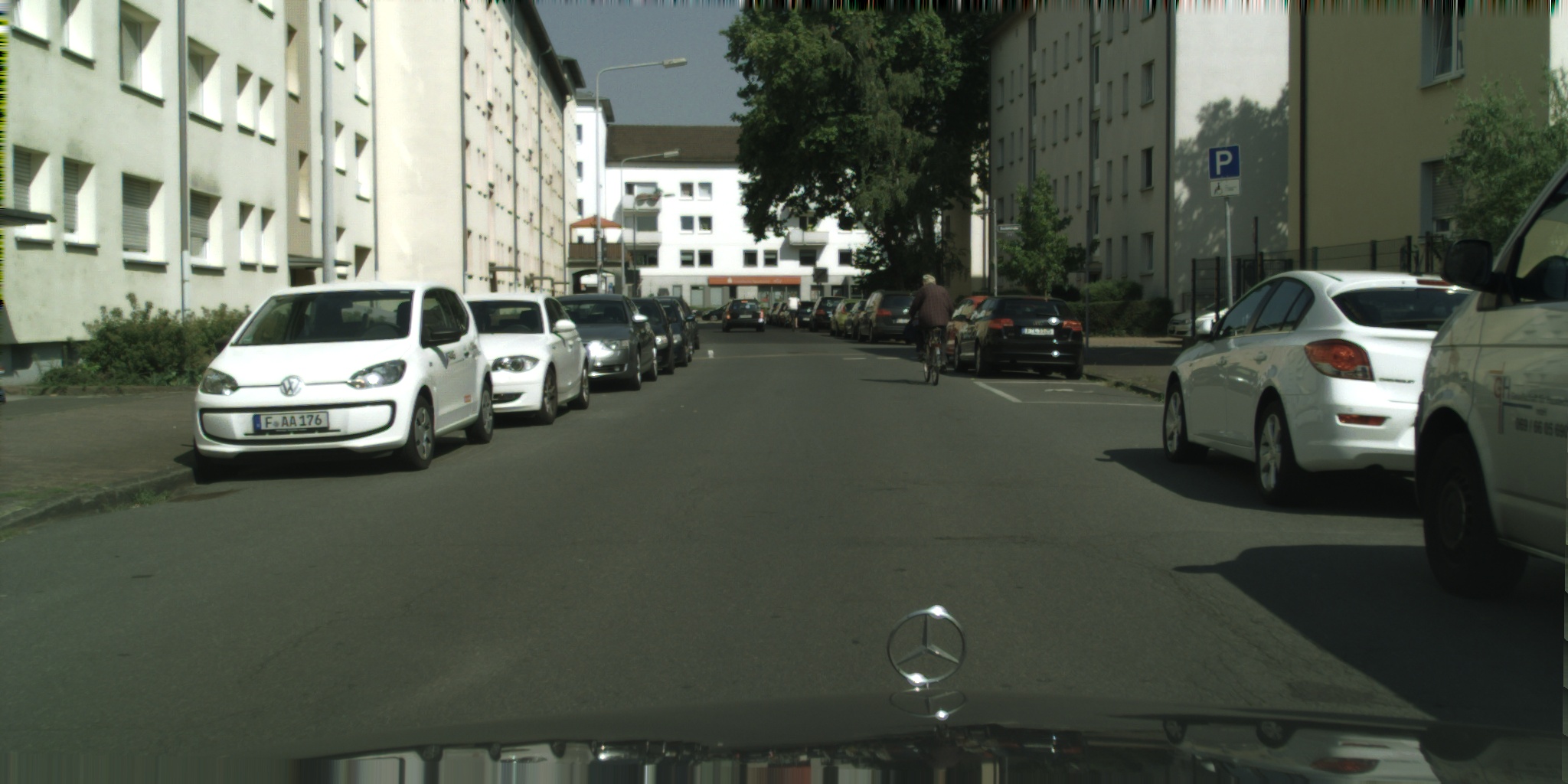}&
         \includegraphics[width=0.241\linewidth, trim={35cm 10cm 10cm 3cm}, clip]{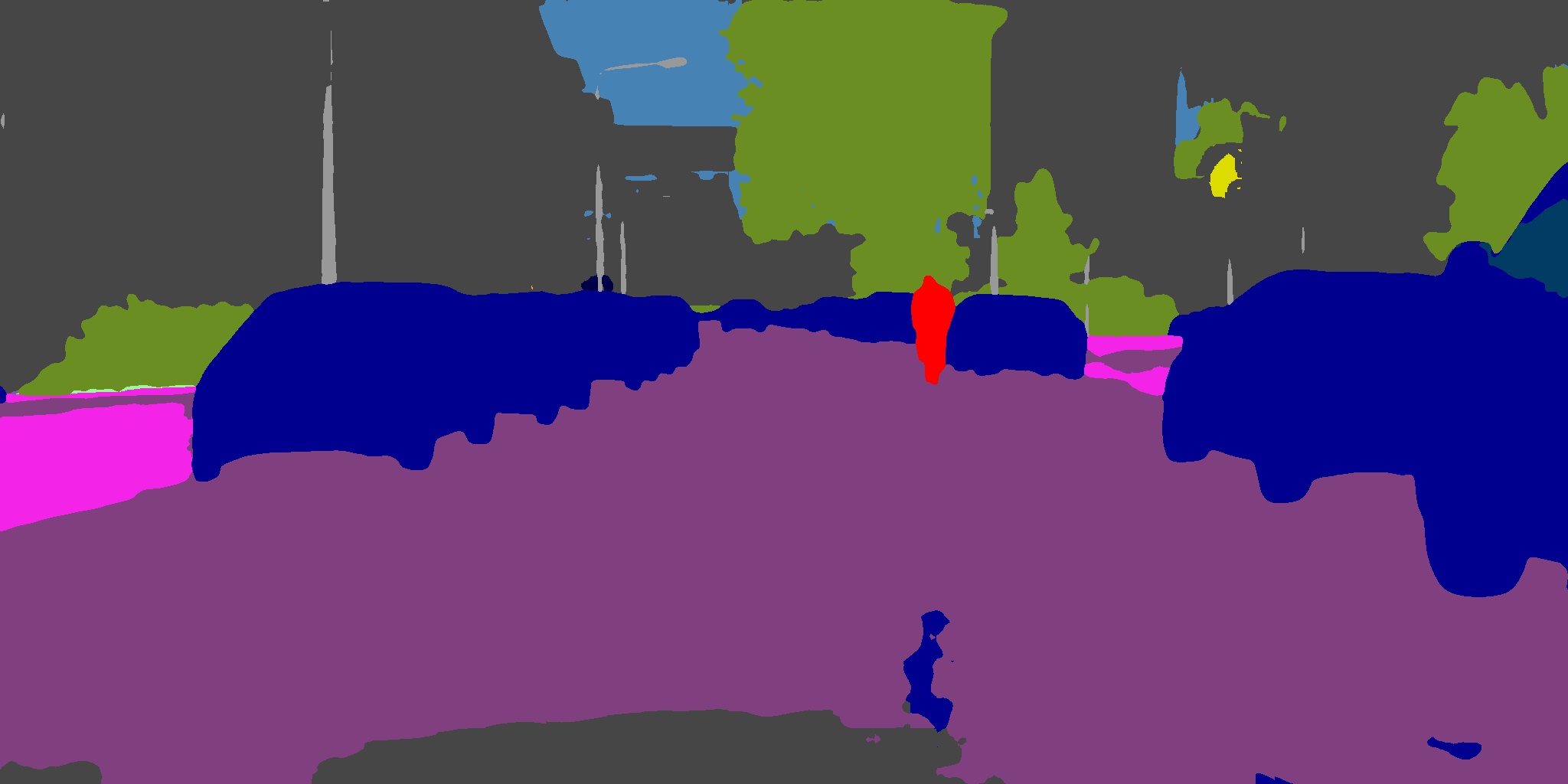}&
         \includegraphics[width=0.241\linewidth, trim={35cm 10cm 10cm 3cm}, clip]{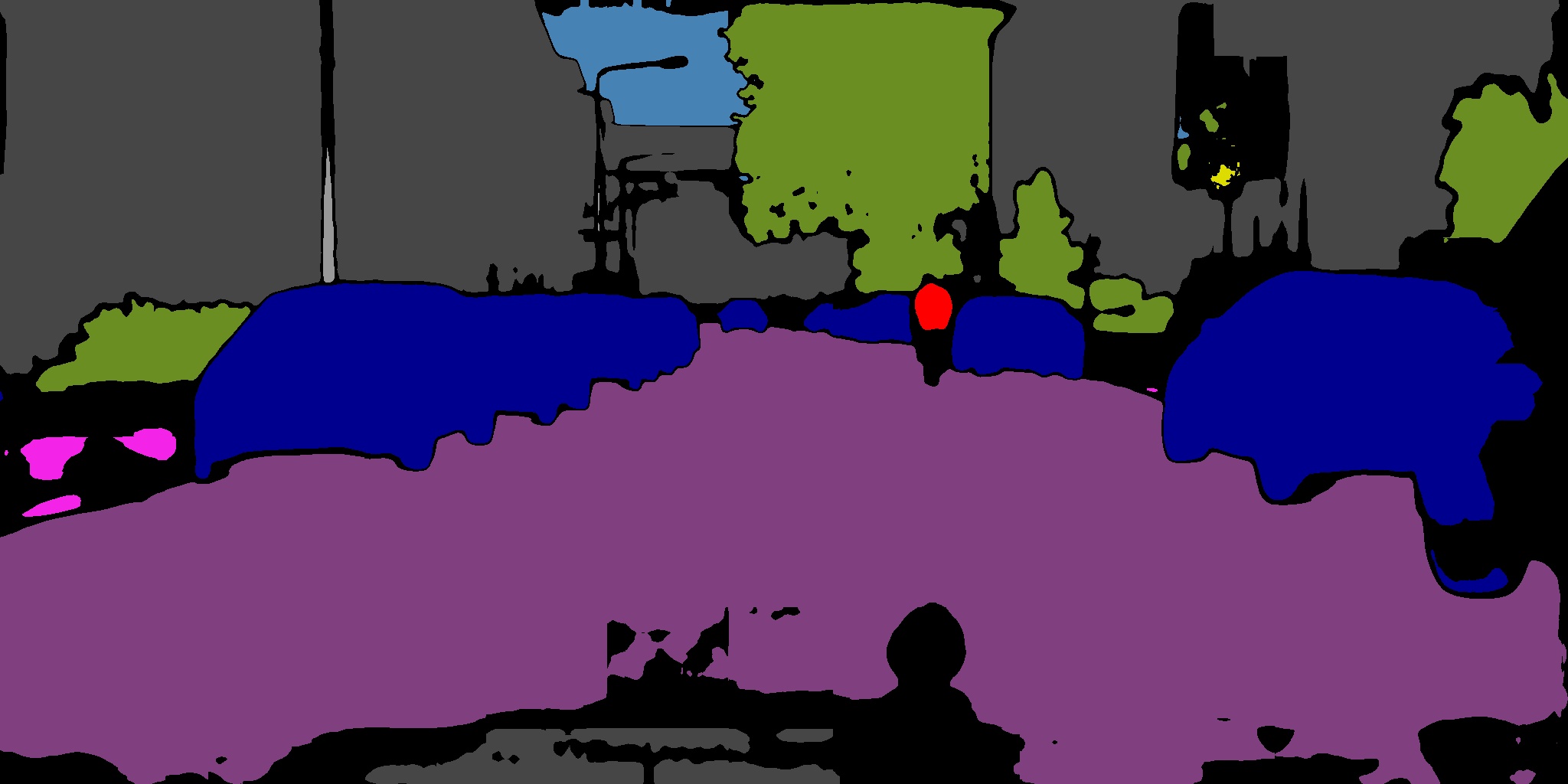}&
         \includegraphics[width=0.241\linewidth, trim={35cm 10cm 10cm 3cm}, clip]{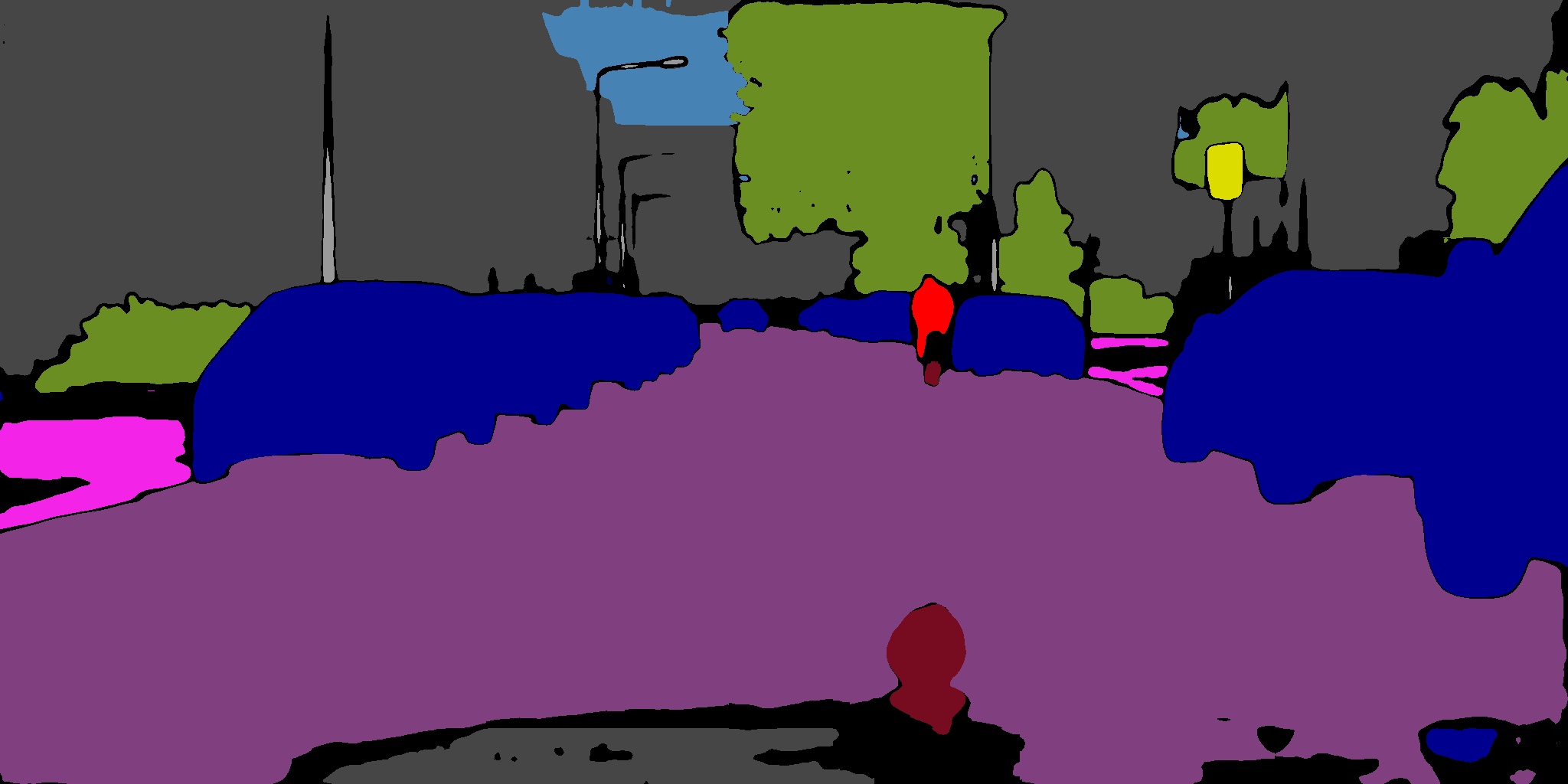}\\
         \footnotesize a) Input image &
         \footnotesize b) Initial pseudo labels &
         \footnotesize c) W/o low confidence predictions &
         \footnotesize d) After alignment to prototypes

    \end{tabular}

    \vspace{-0.5em}
    \caption{Pseudo label expansion. For a target domain image (a), our teacher network predicts an initial pseudo label map (b), from which we retain only high-confidence predictions (c). For low-confidence predictions, we assign the label of the closest pseudo label prototype and obtain our final pseudo label map (d). The expansion commonly corrects the labels of ``hard classes`` such as \emph{bicycle} \& \emph{traffic sign} here.}
    \label{fig:pseudo_labels}
\end{figure*}

We employ three different strategies to assemble contrastive pairs.
First, as ground truth labels are available in the source domain, we exploit this knowledge to sample positive and negative pairs in the source domain, as proposed for supervised segmentation~\cite{Wang2021:ICCV}. 
Second, we construct class-dependent feature prototypes $\overline{f}_c$ for each domain to compress the learned representations.
Third, we leverage the prototypes to further align source and target domain.

A prototype for a class $c$ is assembled as a weighted average of features:
\begin{align}
    \overline{f_c} = \frac{1}{\sum_{f \in F_c}\omega(f,c)} \sum_{f \in F_c} \omega(f,c) \cdot f,
    \label{eq:avg_feature}
\end{align}
where $F_c$ is a set of admissible features $f$.
We weight the contribution of individual features by $\omega(f,c)$, which is the softmax probability of the respective pixel as provided by the teacher network.
For computing prototypes on the images of the source domain, we do not need to resort to the teacher network as ground truth is available, and we simply set $\omega(f,c) \coloneqq 1$.
The set $F_c$ contains all feature vectors for which weights exceed a class-dependent threshold $\tau_c$:
\begin{equation}
    F_c = \{f\;|\;c = \argmax_c\: \omega(f,c) \wedge \omega(f,c) > \tau_c\}.
\end{equation}
We build the prototypes for each mini-batch during training.

To compress feature representations with the prototypes in our contrastive objective, we sample anchor features $\bm v$ and pair them with respective class prototypes from the same domain -- 
we create positive pairs with the class prototype $\overline{f_c}$ of the corresponding class and negative pairs with other prototypes. This is visualized in Figure~\ref{fig:contrasts} (a), and the effect on the learned feature representations can be seen in Figure~\ref{fig:feature_align} (b). Our class-dependent prototypes resemble the category-centers in the regularization approach of Ma~\etal~\cite{Ma2021:CVPR}. Different from their work, however, we incorporate the prototypes into a contrastive framework, where we use them to align learned representations across domains explicitly. 

To enforce alignment of feature representations across domains, we further pair anchors from the source domain with the corresponding class prototypes from the target domain, see Figure~\ref{fig:contrasts} (b). This has the effect of pulling the representations learned for specific classes on images from different features together and can be observed in the learned representations in Figure~\ref{fig:feature_align} (c).

\subsection{Pseudo label expansion}\label{sec:label_expansion}
A common issue in self-training is the generation of reliable pseudo labels for a diverse set of samples.
As the teacher network represents an earlier or ensembled version of the student network, it inherits the student network's classification capabilities. It typically classifies easy examples with high confidence, but hard examples, which would be informative for training, only with low confidence. To extend the set of pseudo-labeled examples we could simply lower the confidence threshold, but this inevitably increases the number of mislabeled training examples, which in turn reduces accuracy.

A possible cause for hard examples is a missing clear correspondence between source and target domain~\cite{Chang2019:CVPR,Dong2020:ECCV}. We show an example for adapting traffic signs from GTA 5 to Cityscapes in Figure~\ref{fig:teaser}. While traffic signs are present in both domains, they typically belong to a set of hard classes, which are misclassified to a large extent. 
Prior work attributed this to a smaller pixel footprint in the dataset~\cite{Chen2019:ICCV}, but we empirically found that even after adjusting for this, traffic signs keep getting misclassified. We hypothesize that the underlying reason is a limited diversity in the source domain paired with a high diversity in the target domain.

To mitigate this discrepancy, we propose a pseudo label expansion mechanism. Intuitively, we want to find more samples of hard-to-label classes to which we can assign a label reliably.
For this, we take inspiration from the class-dependent prototypes from Sec.~\ref{sec:contrastive_objective}
(from here on termed \emph{contrastive} prototypes to avoid confusion), and compute prototypes from features of the teacher network here, which we term \emph{pseudo-label prototypes}.

As with the contrastive prototypes, we assemble pseudo-label prototypes from features extracted at the penultimate network layer. To increase robustness, we accumulate the features from which we compute the prototypes over 200 training iterations, instead of a single mini-batch as with the contrastive ones.
Starting from an input image of the target domain (see \Fig~\ref{fig:pseudo_labels} (a)), we obtain an initial pseudo label map (\Fig~\ref{fig:pseudo_labels} (b)). We treat the softmax probabilities of the network as a measure of confidence and retain only labels for which they exceeds a class-dependent threshold (\Fig~\ref{fig:pseudo_labels} (c)). For simplicity, we use the same adaptive mechanism~\cite{Ma2021:CVPR} as for the contrastive prototypes.
We assign pseudo labels to remaining low-confidence predictions.
To that end, we match each low-confidence feature with the closest pseudo-label prototype and assign the respective label. If there exists no suitable pseudo-label prototype within a maximal distance, we do not assign a label. For details on the distance computation we refer to the supplementary material.

As shown in Figure~\ref{fig:pseudo_labels} (d), our pseudo label expansion is able to correct mispredictions of ``hard samples``, typically belonging to classes with few training samples and a considerable difference in diversity between source and target domain. This is further confirmed in the quantitative evaluation we conduct in Section~\ref{sec:evaluation}.

\subsection{Training \& implementation details}\label{sec:details}
\mypara{Augmentations.}
To learn robust features and improve generalization, we augment the images fed to the student network. Specifically, we apply random color transformations, random scaling and rotations, as well as CutOut~\cite{DeVries2017:arxiv}. Geometric transformations are accordingly applied to the pseudo label maps to ensure correspondence between image content and labels. For more details we refer to the supplemental material.

\mypara{Contrastive pairs.}
For each mini-batch, we use ground truth and pseudo labels to sample $K$ features per class in each domain. These features represent anchors $\bm v$. To build contrastive correspondences $\bm{v}^{+/-}$ for these anchors, we employ a momentum strategy~\cite{He2020:CVPR} and store the $K$ most recent feature embeddings per class at the pixel and prototype level. We use $K = 1000$ in our experiments.
Finally, $(\bm{v}, \bm{v}^+)$ from the same category are used as the positive pairs. All samples  $\bm{v}^-$ from other categories are used to build the negative pairs $(\bm{v}, \bm{v}^-)$.
Sampling the same number of contrastive pairs per class mitigates the effect of an unbalanced distribution of samples per class. Note that even if no samples of a particular class are available as anchors in a mini-batch, samples of that class may still participate in the contrastive loss through the momentum strategy applied to the pixel-level and prototype-level embeddings.

\mypara{Loss.}\label{sec:loss_function}
We train our method with a cross-entropy loss and a contrastive loss.
The cross-entropy is applied for direct supervision of the network per pixel:
\begin{equation}
    \mathcal{L}_{cls} = \mathcal{L}_{CE}(s, y_s) + \lambda \mathcal{L}_{CE} (t, \hat{y}_t),
\end{equation}
where $y$ is a ground truth label from the source domain, $\hat y$ a pseudo label, and $s/t$ are the source or target domain predictions by the student network. 
A weight $\lambda = 0.1$, applied to samples from the target domain, accounts for imperfect supervision from pseudo labels.

The in-domain and cross-domain contrastive objectives are based on the contrastive loss of Eq.\eqref{EQ:contrast}. 
The in-domain contrastive losses are given by:
\begin{equation}
    \mathcal{L}_{in} = \mathcal{L}_{ctr}(s,s^+) + \mathcal{L}_{ctr} (t, t^+),
\end{equation}
where $(s, s^+)$ are contrastive pairs from the source domain and $(t, t^+)$ are the contrastive pairs from the target domain. Note that $s^+$ can be either a pixel representation or class prototype. $t^+$ represents the class prototypes in the target domain. 

The cross-domain contrastive loss is given by 
\begin{equation}
\mathcal{L}_{cross} = \mathcal{L}_{ctr}(s,t^+),
\end{equation}
where $s$ are the pixels from the source domain and $t^+$ are the class prototypes computed on the target domain.

The final training loss is given by
\begin{equation}
\mathcal{L} = \mathcal{L}_{cls} + \alpha (\mathcal{L}_{in} + \mathcal{L}_{cross}).
\end{equation}
We set $\alpha = 0.1$ in all experiments.

\mypara{Training details.}
All experiments were conducted on 8 24GB GPUs with 2 samples per GPU. We use momentum of 0.9 and weight decay of 1e-4 for training. All other training settings are the same as \cite{pspnet} (detailed in the supplement). 
We start with an initial learning rate of 0.01 for pre-training on the source domain. 
In the self-training phase, we train for 25K iterations and the learning rate starts at 1e-3.

\section{Experiments}\label{sec:evaluation}
We follow the evaluation protocol that was proposed in \cite{CAG,Ma2021:CVPR} and evaluate our method with source domain datasets GTA5~\cite{gta5} and SYNTHIA\cite{synthia}, and with target domain datasets  Cityscapes~\cite{cordts2016cityscapes} and Mapillary Vistas~\cite{neuhold2017mapillary} (Vistas in the supplement). The GTA5 dataset shares 19 common classes with Cityscapes and we ignore classes that are not shared during training. SYNTHIA shares 16 classes with Cityscapes. Some existing works only train and test on a 13-class subset of SYNTHIA, or train two separate models on the subset and on the full set. 
Here we follow the practice introduced in \cite{intra_domain,Wang2020:ECCV} and train a model on the full label set but test it on both settings (13 \& 16 classes). 
We use the training set (ignoring the labels) from the respective target domain to perform unsupervised adaptation for all experiments. We report results in terms of per-class Intersection over Union (IoU) as well as the mean IoU (mIoU) over all classes on the respective validation sets.
\begin{table*}[t]
	\centering
	\resizebox{\linewidth}{!}{
	\begin{tabular}{l l l l l l l l l l l l l l l l l l l l l l}
		\hline
		& \rotatebox{90}{road} & \rotatebox{90}{sidewalk} & \rotatebox{90}{building} & \rotatebox{90}{wall} & \rotatebox{90}{fence} & \rotatebox{90}{pole} & \rotatebox{90}{light} & \rotatebox{90}{sign} & \rotatebox{90}{veget.} & \rotatebox{90}{terrain} & \rotatebox{90}{sky} & \rotatebox{90}{person} & \rotatebox{90}{rider} & \rotatebox{90}{car} & \rotatebox{90}{truck} & \rotatebox{90}{bus} & \rotatebox{90}{train} & \rotatebox{90}{m.cycle} & \rotatebox{90}{bicycle} & mIoU \\
		\hline
		BDL~\cite{BDL} & 91.0 & 44.7 & 84.2 & 34.6 & 27.6 & 30.2 & 36.0 & 36.0 & 85.0 & {43.6} & 83.0 & 58.6 & 31.6 & 83.3 & 35.3 & 49.7 & 3.3 & 28.8 & 35.6 & 48.5 \\
		IDA~\cite{intra_domain} & 90.6 & 36.1 & 82.6 & 29.5 & 21.3 & 27.6 & 31.4 & 23.1 & 85.2 & 39.3 & 80.2 & 59.3 & 29.4 & 86.4 & 33.6 & 53.9 & 0.0 & 32.7 & 37.6 & 46.3 \\
		BiMaL~\cite{truong2021bimal} & 91.2 & 39.6 &82.7 & 29.4 & 25.2 & 29.6 & 34.3 & 25.5 & 85.4 & 44.0 & 80.8 & 59.7 & 30.4 & 86.6 & 38.5 & 47.6 & 1.2 & 34.0 & 36.8 & 47.3\\
		DTST~\cite{Wang2020:CVPR} & 90.6 & 44.7 & 84.8 & 34.3 & 28.7 & 31.6 & 35.0 & 37.6 & 84.7 & 43.3 & 85.3 & 57.0 & 31.5 & 83.8 & {42.6} & 48.5 & 1.9 & 30.4 & 39.0 & 49.2\\
		FGGAN~\cite{Wang2020:ECCV} & 91.0 & 50.6 & 86.0 & {43.4} & {29.8} & 36.8 & 43.4 & 25.0 & {86.8} & 38.3 & 87.4 & 64.0 & {38.0} & 85.2 & 31.6 & 46.1 & 6.5 & 25.4 & 37.1 & 50.1\\
		FDA~\cite{Yang2020:CVPRa} & {92.5} & 53.3 & 82.3 & 26.5 & 27.6 & 36.4 & 40.5 & 38.8 & 82.2 & 39.8 & 78.0 & 62.6 & 34.4 & 84.9 & 34.1 & {53.1} & 16.8 & 27.7 & 46.4 & 50.4 \\
		CAG~\cite{CAG} & 90.4 & 51.6 & 83.8 & 34.2 & 27.8 & {38.4} & 25.3 & {48.4} & 85.4 & 38.2 & 78.1 & 58.6 & 34.6 & 84.7 & 21.9 & 42.7 & 41.1 & 29.3 & 37.2 & 50.2\\
		Uncertainty~\cite{wang2021uncertainty} & 90.5 & 38.7 & 86.5 & 41.1 & 32.9 & 40.5 & 48.2 & 42.1 & 86.5 & 36.8 & 84.2 & 64.5 & 38.1 & 87.2 & 34.8 & 50.4 & 0.2 & 41.8 & 54.6 & 52.6\\
        SAC~\cite{Araslanov2021:CVPR} &90.4 & 53.9 & 86.6 & 42.4 & 27.3 & 45.1 & 48.5 & 42.7 & 87.4 & 40.1 & 86.1& 67.5 & 29.7 & 88.5 & 49.1 & 54.6 & 9.8 & 26.6 & 45.3 & 53.8\\
        RPT~\cite{Zhang2020:CVPR}$^*$ (FCN-101) & 89.7 & 44.8 & 86.4 & 44.2 & 30.6 & 41.4 & 51.7 & 33.0 & 87.8 & 39.4 & 86.3 & 65.6 & 24.5 & 89.0 & 36.2 & 46.8 & 17.6 & 39.1 & 58.3 & 53.2\\
		coarse-to-fine~\cite{Ma2021:CVPR} $^*$ & {92.5} & {58.3} & {86.5} & 27.4 & 28.8 & 38.1 & {46.7} & 42.5 & 85.4 & 38.4 & {91.8} & {66.4} & 37.0 & {87.8} & 40.7 & 52.4 & {44.6} & {41.7} & {59.0} & {56.1}\\
		BAPA-Net~\cite{liu2021bapa} &94.4 & 61.0 & 88.0 & 26.8 & 39.9 & 38.3 & 46.1 & 55.3 & 87.8 & 46.1 & 89.4 & 68.8 & 40.0 & 90.2 &60.4 & 59.0& 0.00& 45.1& 54.2 &57.4\\
		ProDA~\cite{Zhang2021:CVPR} & 87.8 & 56.0 & 79.7 & 46.3 & 44.8 & 45.6 &53.5 &53.5 &88.6 &45.2 &82.1 &70.7 &39.2 &88.8 &45.5 &59.4 &1.0 &48.9 &56.4 & 57.5\\
		\hline
       Ours & 92.6 & 59.1& {\bf 88.5} &45.8 &40.5& {\bf 52.9} & {\bf 53.6} &  54.1 & 88.0& 41.9 & 86.0 & {\bf 73.5} & {\bf 44.1} & 89.7 & 39.3 & 53.2& 26.8 &{\bf 51.6} & {\bf 61.8} & {\bf 60.2}\\
		\hline
	\end{tabular}}
	\caption{Comparison to prior work on GTA5$\rightarrow$Cityscapes. All methods use a DeepLabV2 (ResNet101) backbone, except Coarse-to-fine~\cite{Ma2021:CVPR} and RPT~\cite{Zhang2020:CVPR}, which use DeepLabV3 (ResNet-101) and FCN (ResNet-101) respectively. } 
	\label{tab:gta2city}
\end{table*}
\begin{table*}[t]
	\centering
	\resizebox{\linewidth}{!}{
	\begin{tabular}{l l l l l l l l l l l l l l l l l l l l l l}
		\hline
		& \rotatebox{90}{road} & \rotatebox{90}{sidewalk} & \rotatebox{90}{building} & \rotatebox{90}{wall} & \rotatebox{90}{fence} & \rotatebox{90}{pole} & \rotatebox{90}{light} & \rotatebox{90}{sign} & \rotatebox{90}{veget.} & \rotatebox{90}{sky} & \rotatebox{90}{person} & \rotatebox{90}{rider} & \rotatebox{90}{car} & \rotatebox{90}{bus} & \rotatebox{90}{m.cycle} & \rotatebox{90}{bicycle} & mIoU & mIoU*\\
		\hline
		BDL~\cite{BDL} & {86.0 } & {46.7 } & 80.3 & - & - & - & 14.1 & 11.6 & 79.2 & 81.3 & 54.1 & 27.9 & 73.7 & 42.2 & 25.7 & 45.3 & - & 51.4 \\
		IDA~\cite{intra_domain} & 84.3 & 37.7 & 79.5 & 5.3 & 0.4 & 24.9 & 9.2 & 8.4 & 80.0 & 84.1 & 57.2 & 23.0 & 78.0 & 38.1 & 20.3 & 36.5 & 41.7 & 48.9 \\
		DTST~\cite{Wang2020:CVPR} & 83.0 & 44.0 & 80.3 & - & - & - & 17.1 & 15.8 & 80.5 & 81.8 & 59.9 & 33.1 & 70.2 & 37.3 & 28.5 & 45.8 & - & 52.1 \\
		FGGAN~\cite{Wang2020:ECCV} & 84.5 & 40.1 & 83.1 & 4.8 & 0.0 & 34.3 & {20.1 } & 27.2 & 84.8 & 84.0 & 53.5 & 22.6 & {85.4 } & {43.7 } & 26.8 & 27.8 & 45.2 & 52.5 \\
		FDA~\cite{Yang2020:CVPRa}  & 79.3 & 35.0 & 73.2 & - & - & - & 19.9 & 24.0 & 61.7 & 82.6 & 61.4 & 31.1 & 83.9 & 40.8 & {38.4 } & 51.1 & - & 52.5 \\
		CAG (13 classes)~\cite{CAG}  & 84.8 & 41.7 & {85.5 } & - & - & - & 13.7 & 23.0 & {86.5 } & 78.1 & {66.3 } & 28.1 & 81.8 & 21.8 & 22.9 & 49.0 & - & 52.6 \\
		CAG (16 classes)~\cite{CAG} & 84.7 & 40.8 & 81.7 & 7.8 & 0.0 & 35.1 & 13.3 & 22.7 & 84.5 & 77.6 & 64.2 & 27.8 & 80.9 & 19.7 & 22.7 & 48.3 & 44.5 & - \\ 
		BiMaL \cite{truong2021bimal}&92.8 & 51.5 & 81.5 & 10.2 & 1.0 & 30.4 & 17.6 & 15.9 & 82.4 & 84.6 & 55.9 & 22.3 & 85.7 & 44.5 & 24.6 & 38.8 & 46.2 & 53.7\\
        Uncertainty\cite{wang2021uncertainty} & 79.4 & 34.6 & 83.5 & 19.3 & 2.8 & 35.3 & 32.1 & 26.9 & 78.8 & 79.6 & 66.6 & 30.3 & 86.1 & 36.6 & 19.5 & 56.9 & 48.0 & 54.6\\
		coarse-to-fine\cite{Ma2021:CVPR} & 75.7 & 30.0 & 81.9 & {11.5} & {2.5} & {35.3} & 18.0 & {32.7} & 86.2 & {90.1} & 65.1 & {33.2} & 83.3 & 36.5 & 35.3 & {54.3} & {48.2 } & {55.5} \\
    	BAPA-Net\cite{liu2021bapa} &91.7 & 53.8 & 83.9 & 22.4 & 0.8 & 34.9 & 30.5 & 42.8 & 86.6 & 88.2 & 66.0 & 34.1 & 86.6 & 51.3 & 29.4 & 50.5 & 53.3 & 61.2\\
		ProDA\cite{Zhang2021:CVPR} & 87.8& 45.7 & 84.6 & 37.1 & 0.6 &44.0 & 54.6&37.0 & 88.1 & 84.4 & 74.2 & 24.3 & 88.2  &51.1 & 40.5 &45.6 & 55.5 & 62.0\\
		\hline
		Ours &85.2& 46.5& 83.3& {\bf 39.2}&{\bf 6.1}& 37.3& 50.1& 40.1& 87.9& 88.2& 70.1& 29.8& 85.4& 45.4& {\bf 59.1} & 49.8& {\bf 56.5} & {\bf 63.1}\\
		\hline
	\end{tabular}}
	\caption{Comparison to prior work on adapting SYNTHIA$\rightarrow$Cityscapes (mIoU: 16-class; mIoU*: 13-class). }
	\label{tab:synthia2city}
\end{table*}

\begin{figure*}[t]
    \begin{subfigure}{0.245\linewidth}
        \begin{minipage}[b]{1\linewidth}
            \includegraphics[width=1\linewidth]{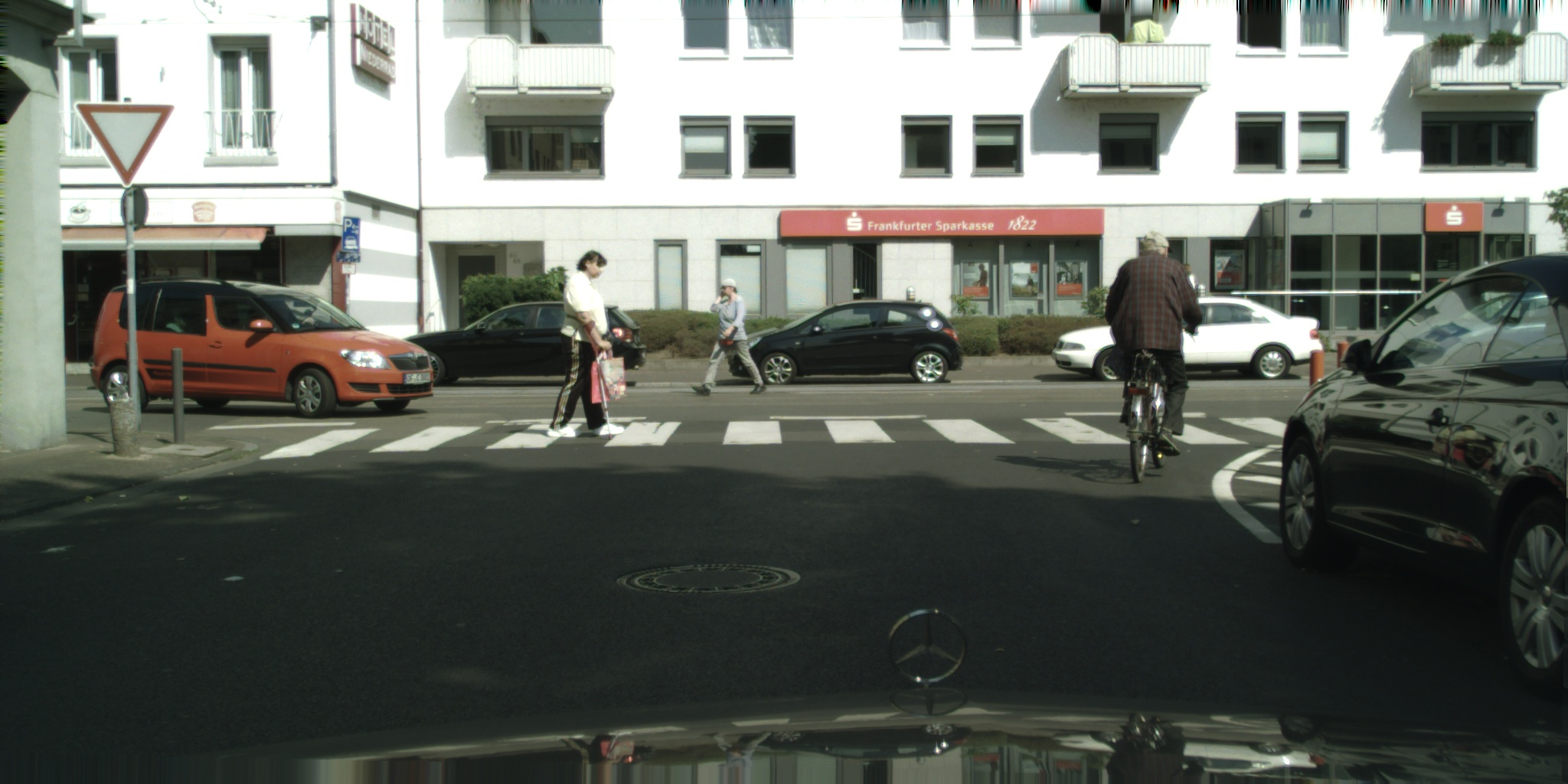}\\
            \includegraphics[width=1\linewidth]{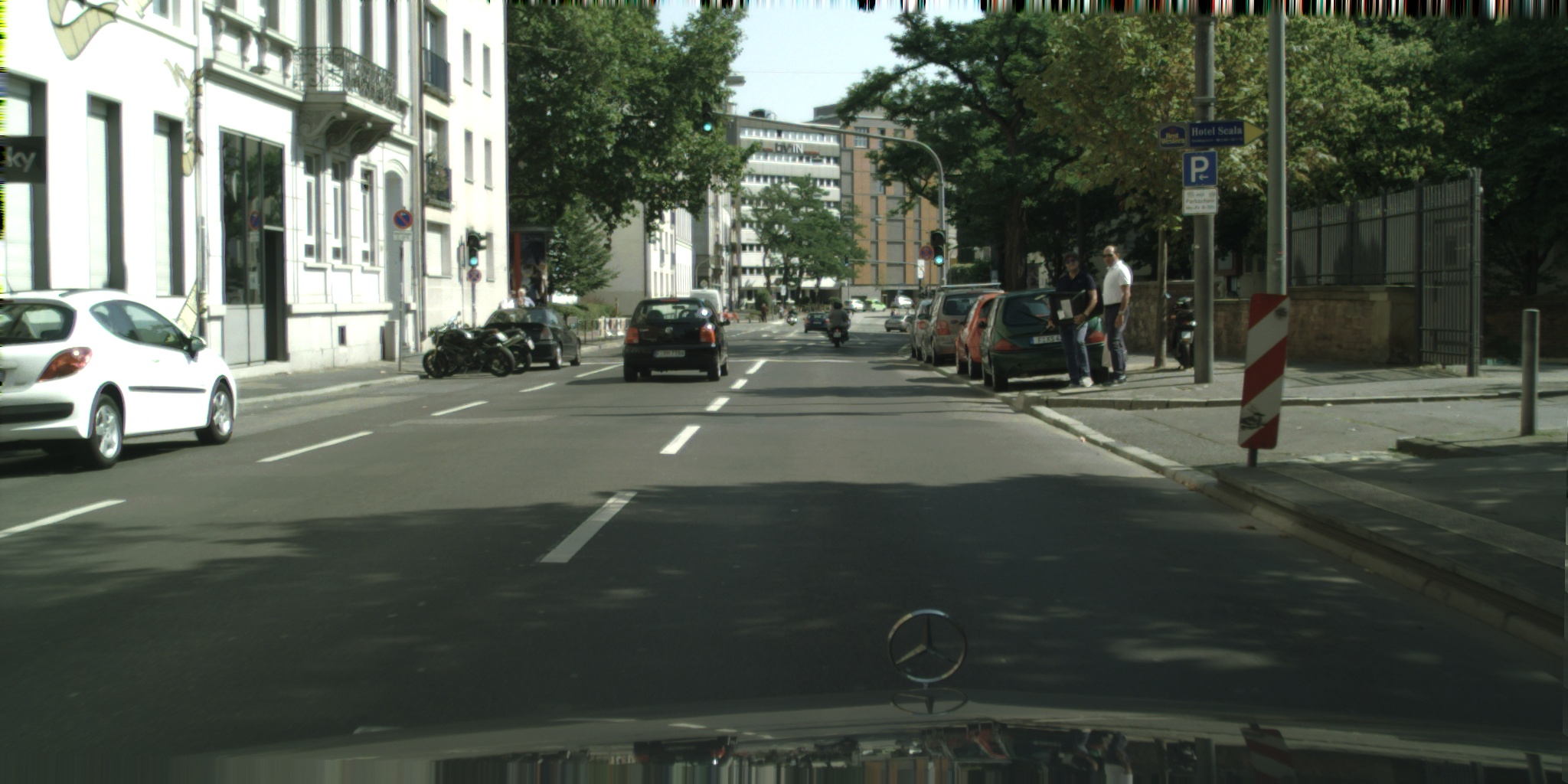}\\
            \includegraphics[width=1\linewidth]{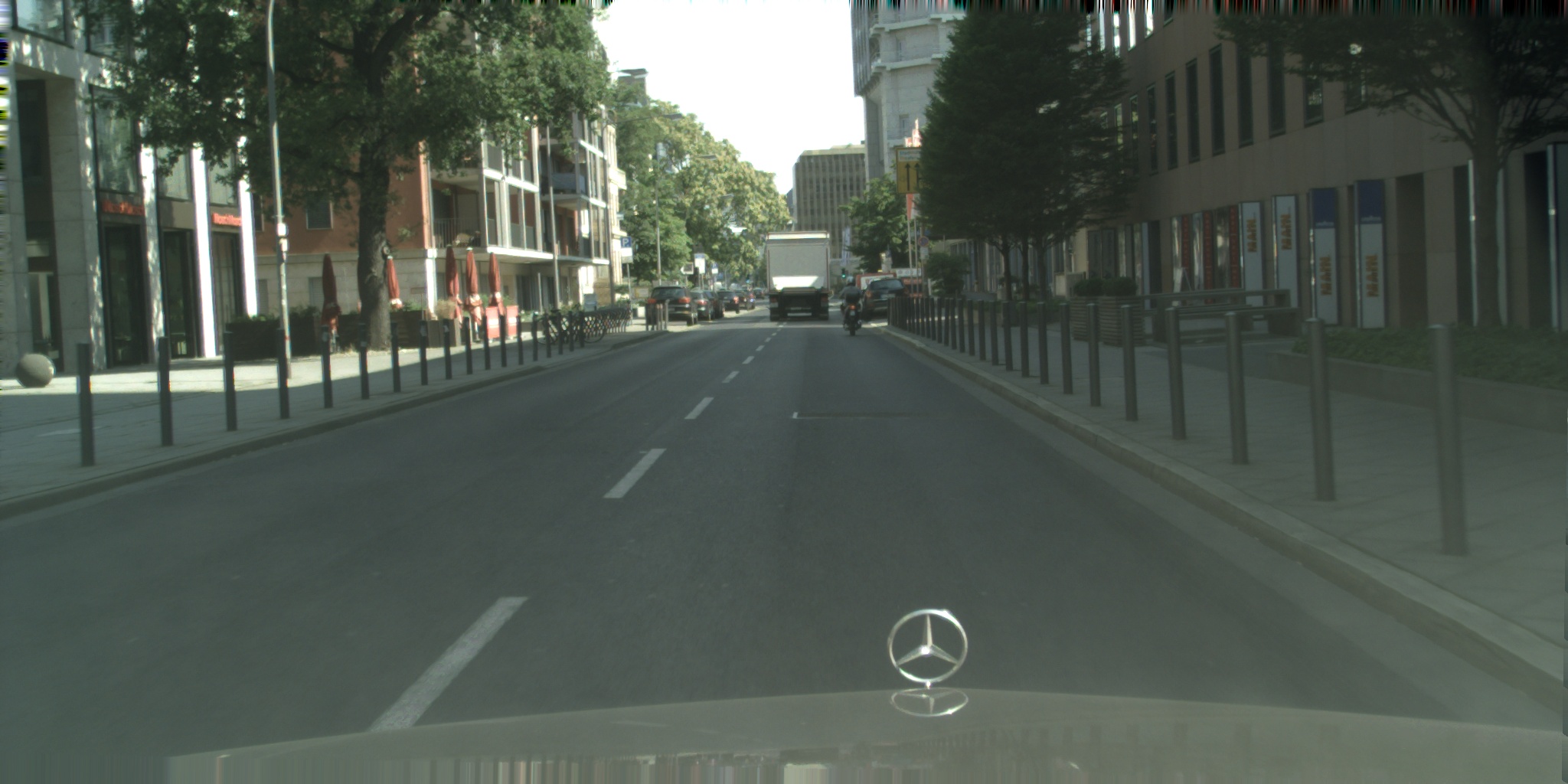}\\
            \includegraphics[width=1\linewidth]{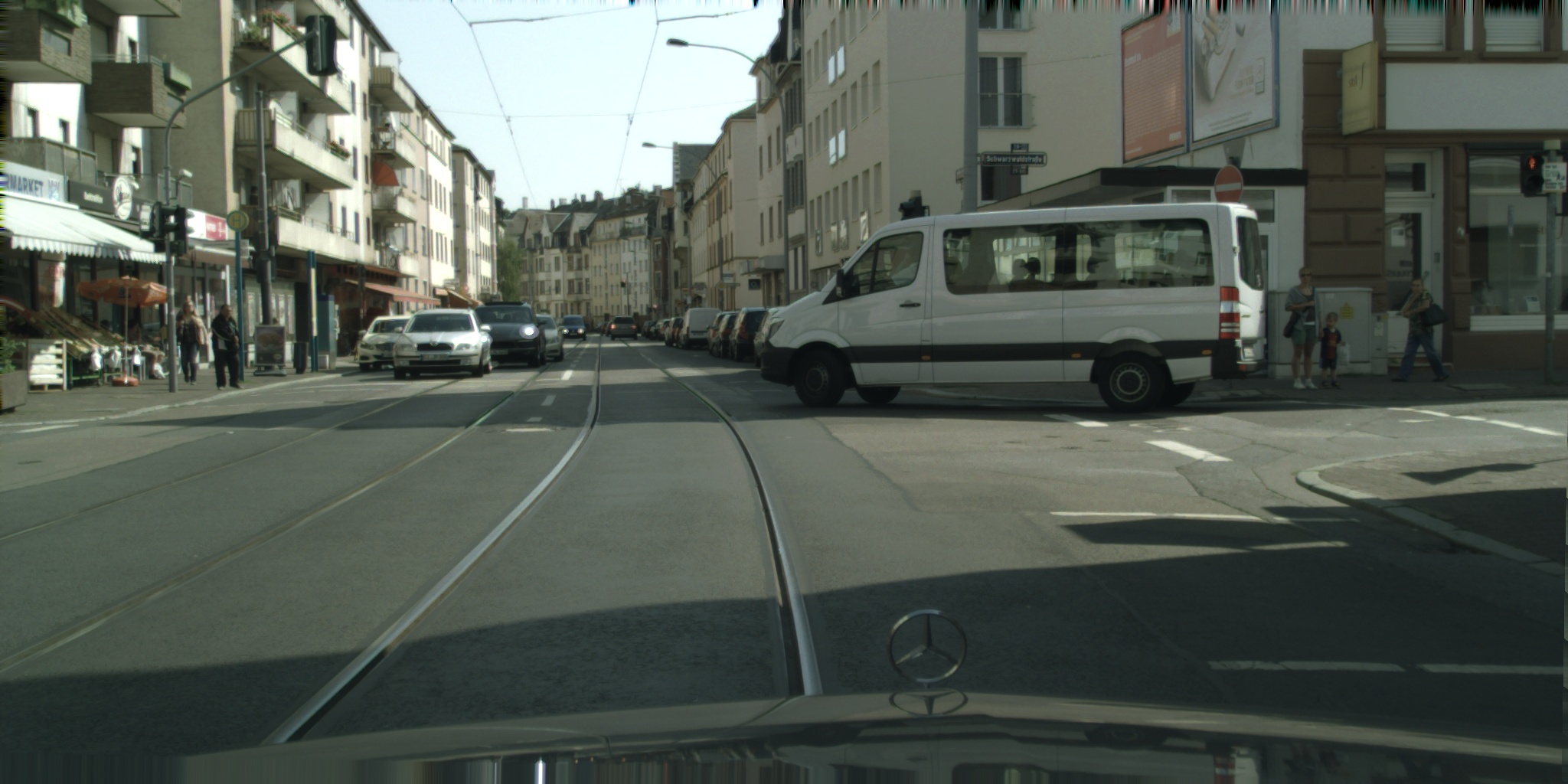}\\
            \includegraphics[width=1\linewidth]{images/results/input_f0_576.jpg}\\
            \includegraphics[width=1\linewidth]{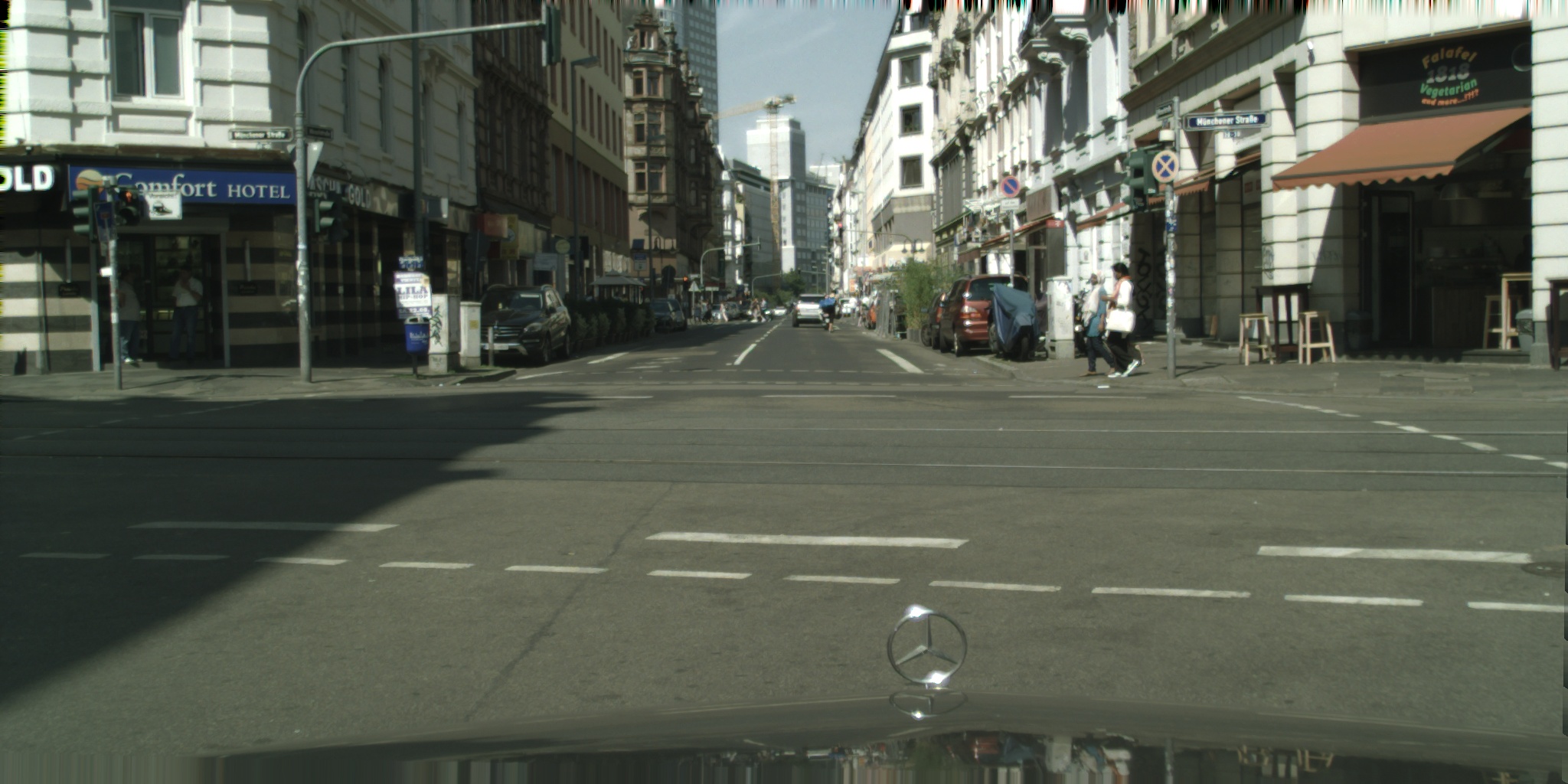}\\
            \includegraphics[width=1\linewidth]{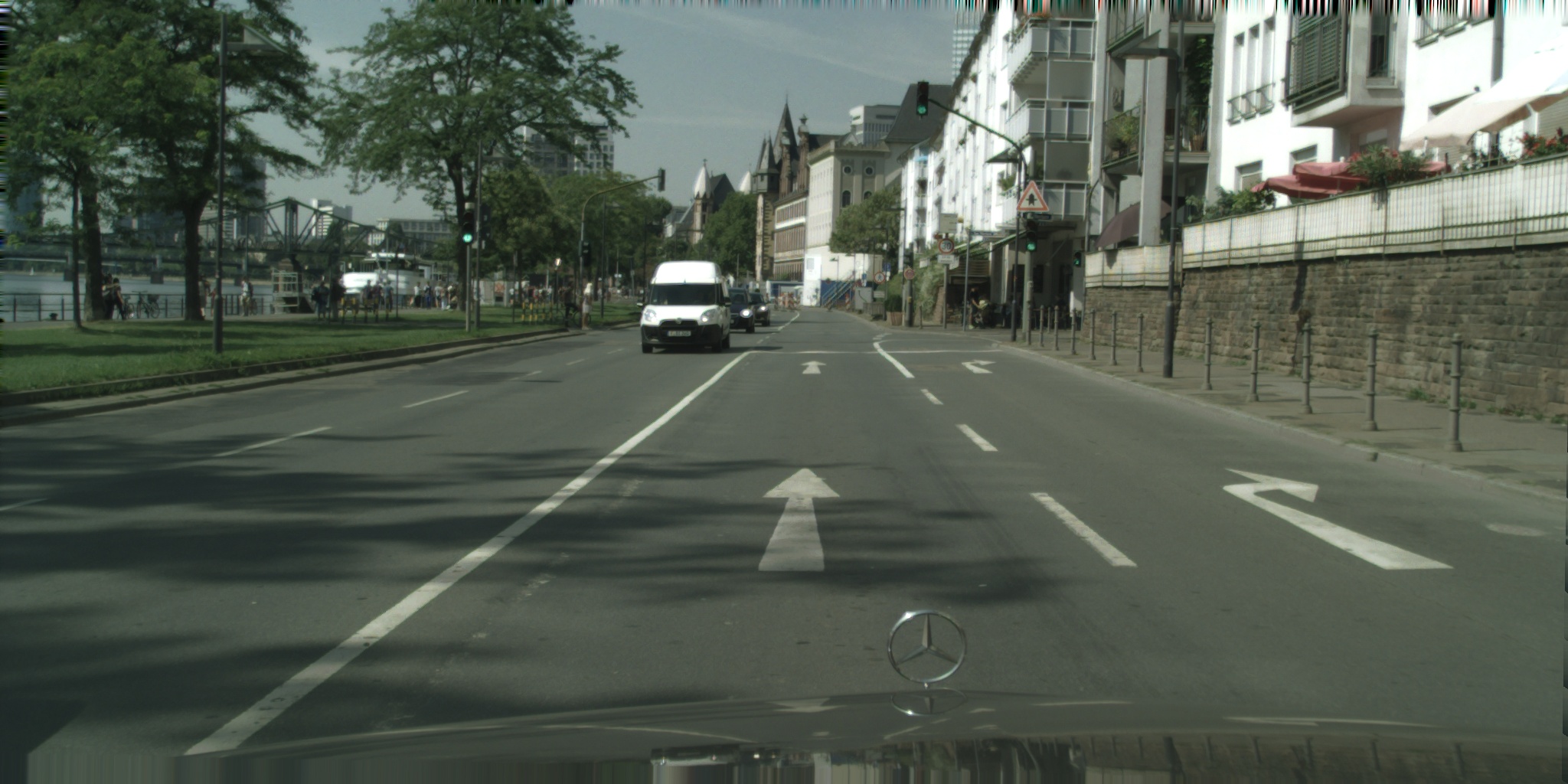}
        \end{minipage}
        \caption{Inputs}
    \end{subfigure}
    \hspace{-1.2mm}
    \begin{subfigure}{0.245\linewidth}
        \begin{minipage}[b]{1\linewidth}
            \includegraphics[width=1\linewidth]{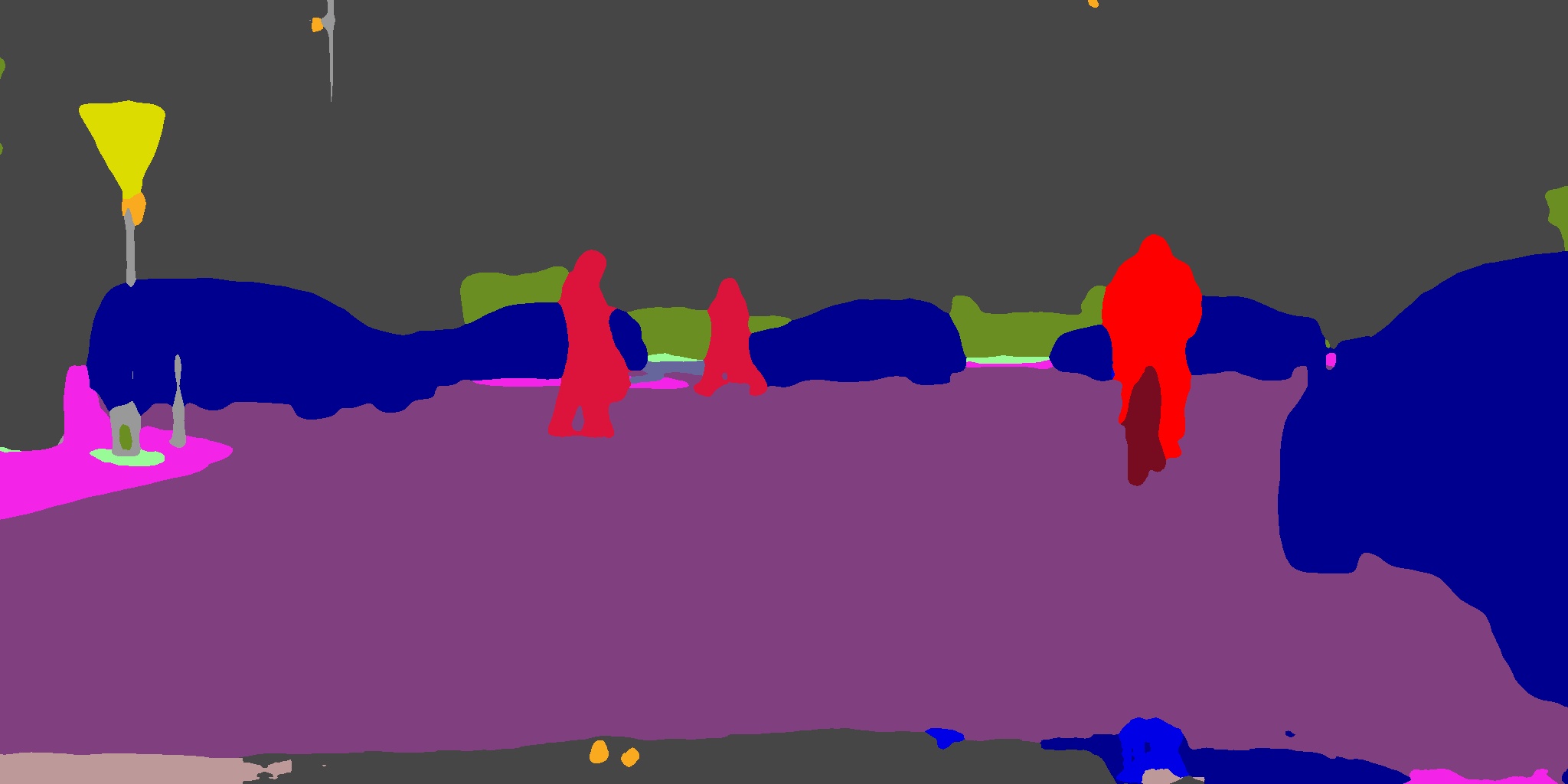}\\
            \includegraphics[width=1\linewidth]{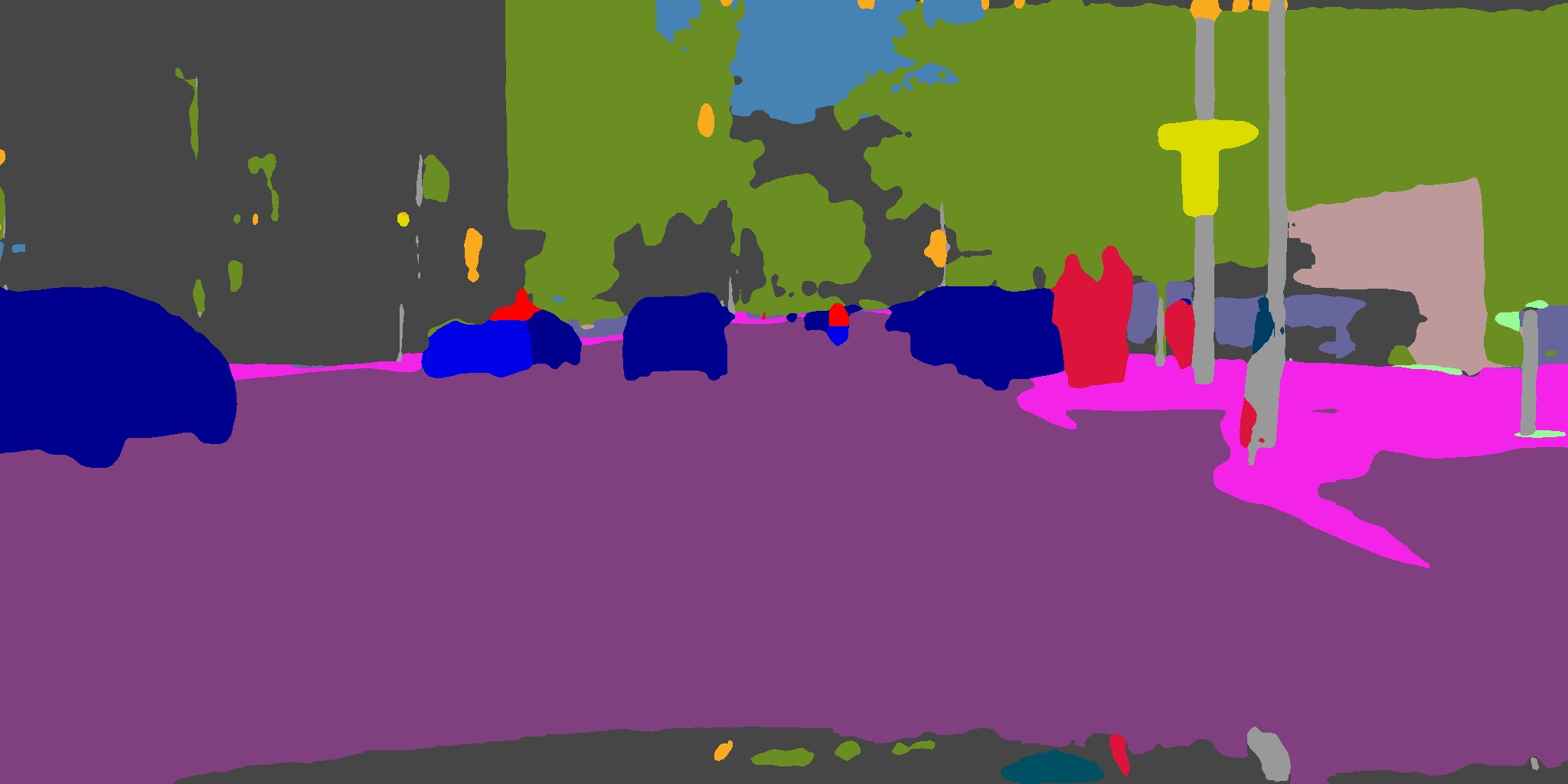}\\
            \includegraphics[width=1\linewidth]{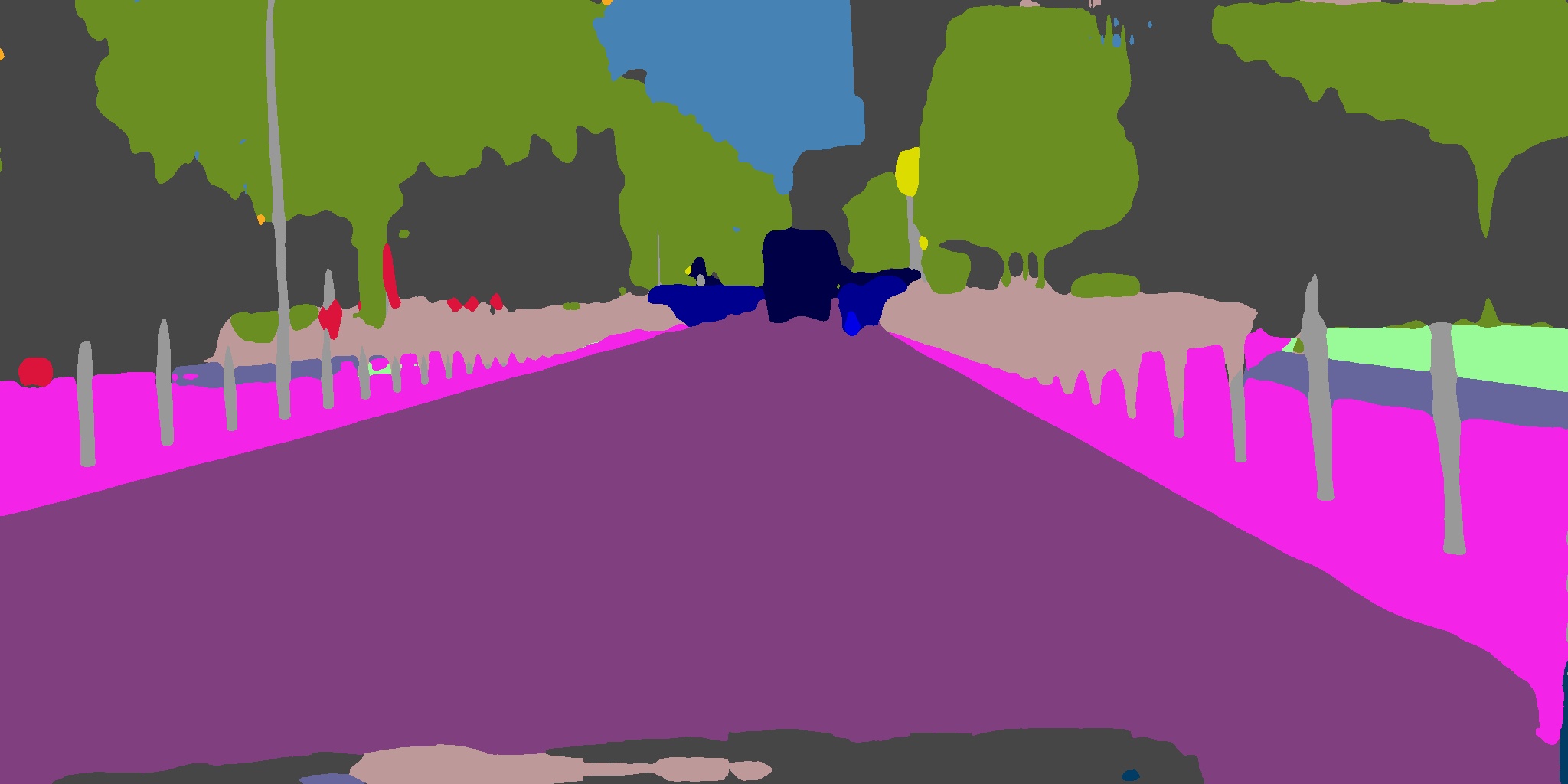}\\
            \includegraphics[width=1\linewidth]{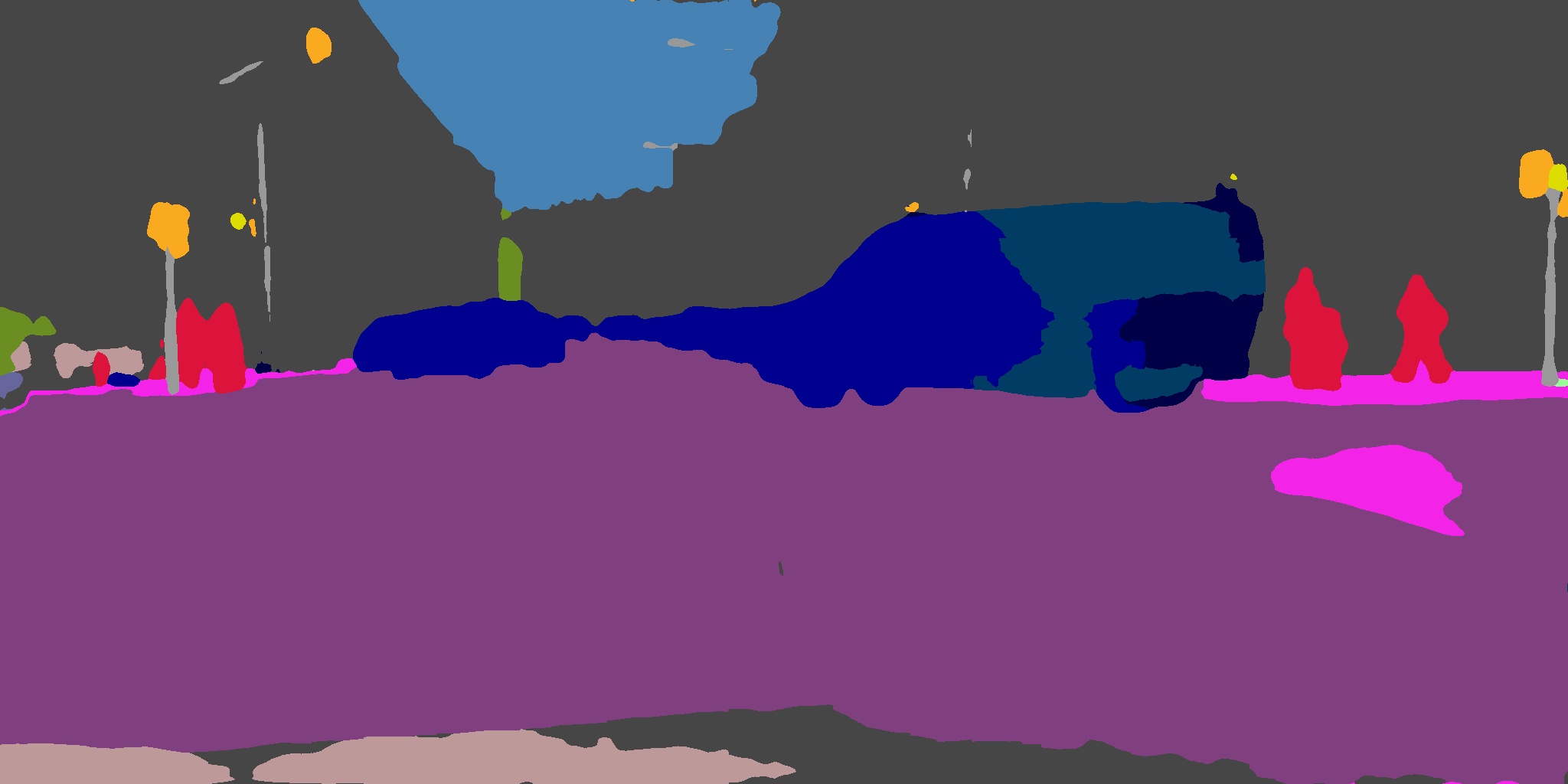}\\
            \includegraphics[width=1\linewidth]{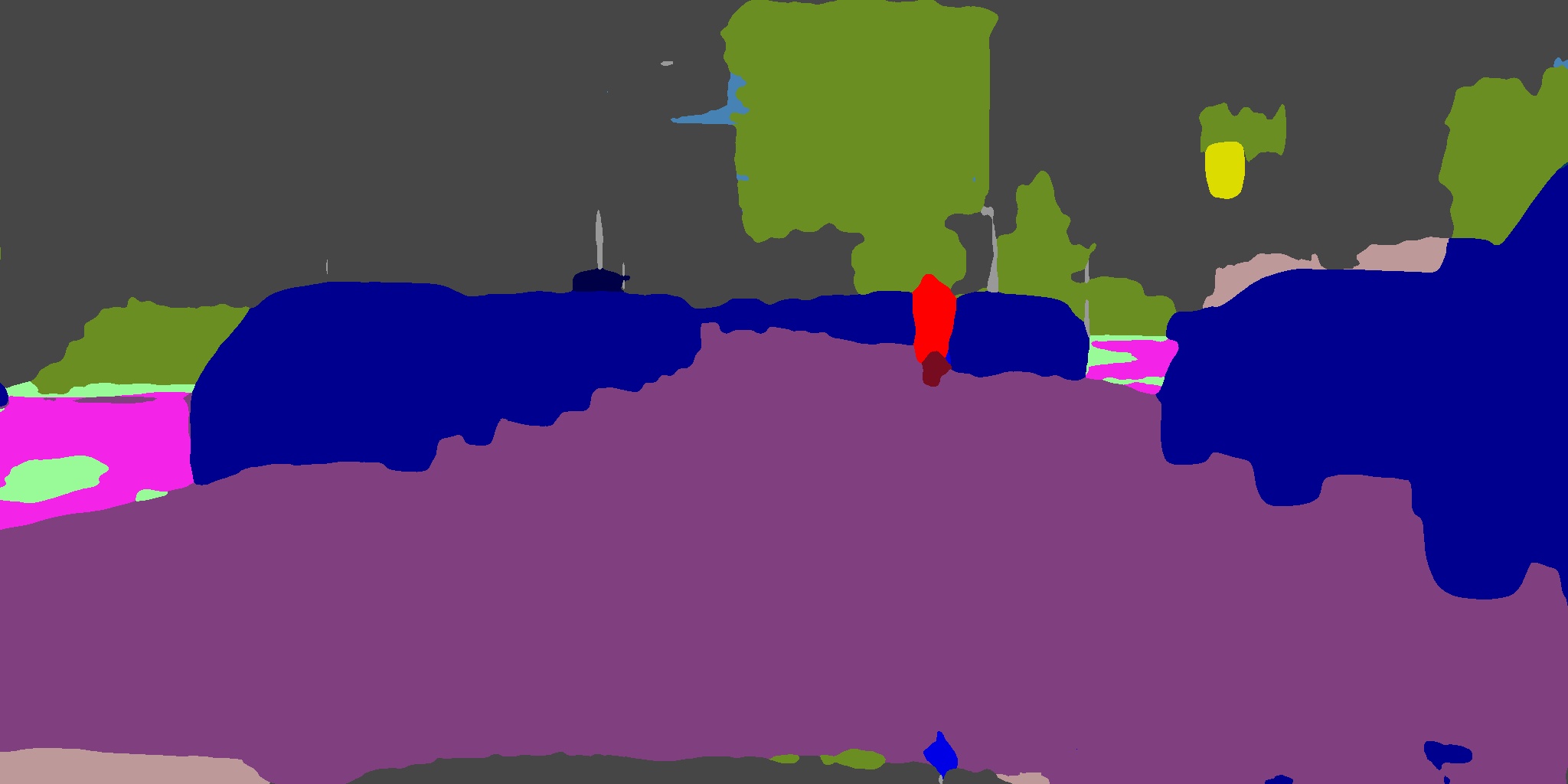}\\
            \includegraphics[width=1\linewidth]{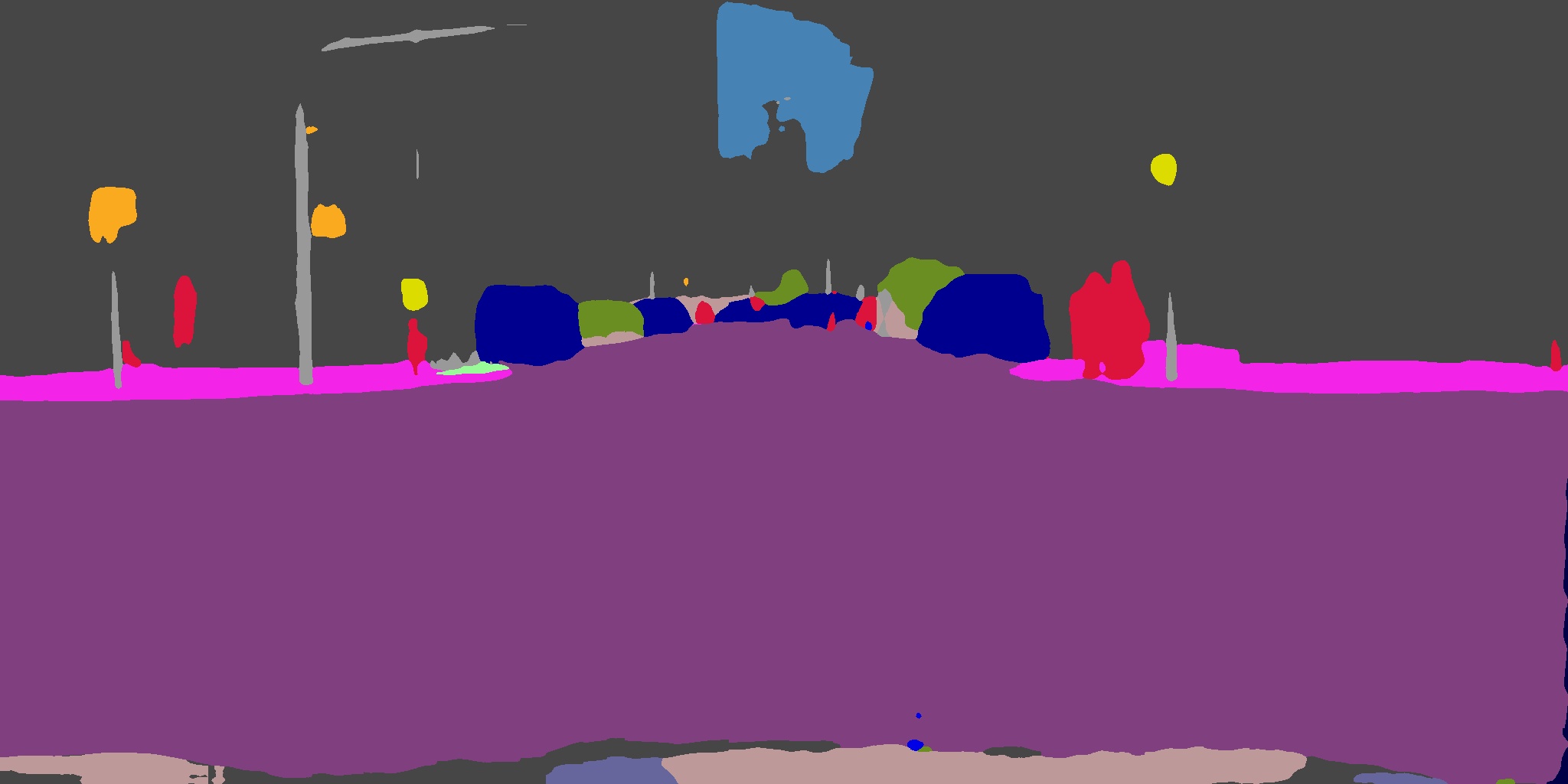}\\
            \includegraphics[width=1\linewidth]{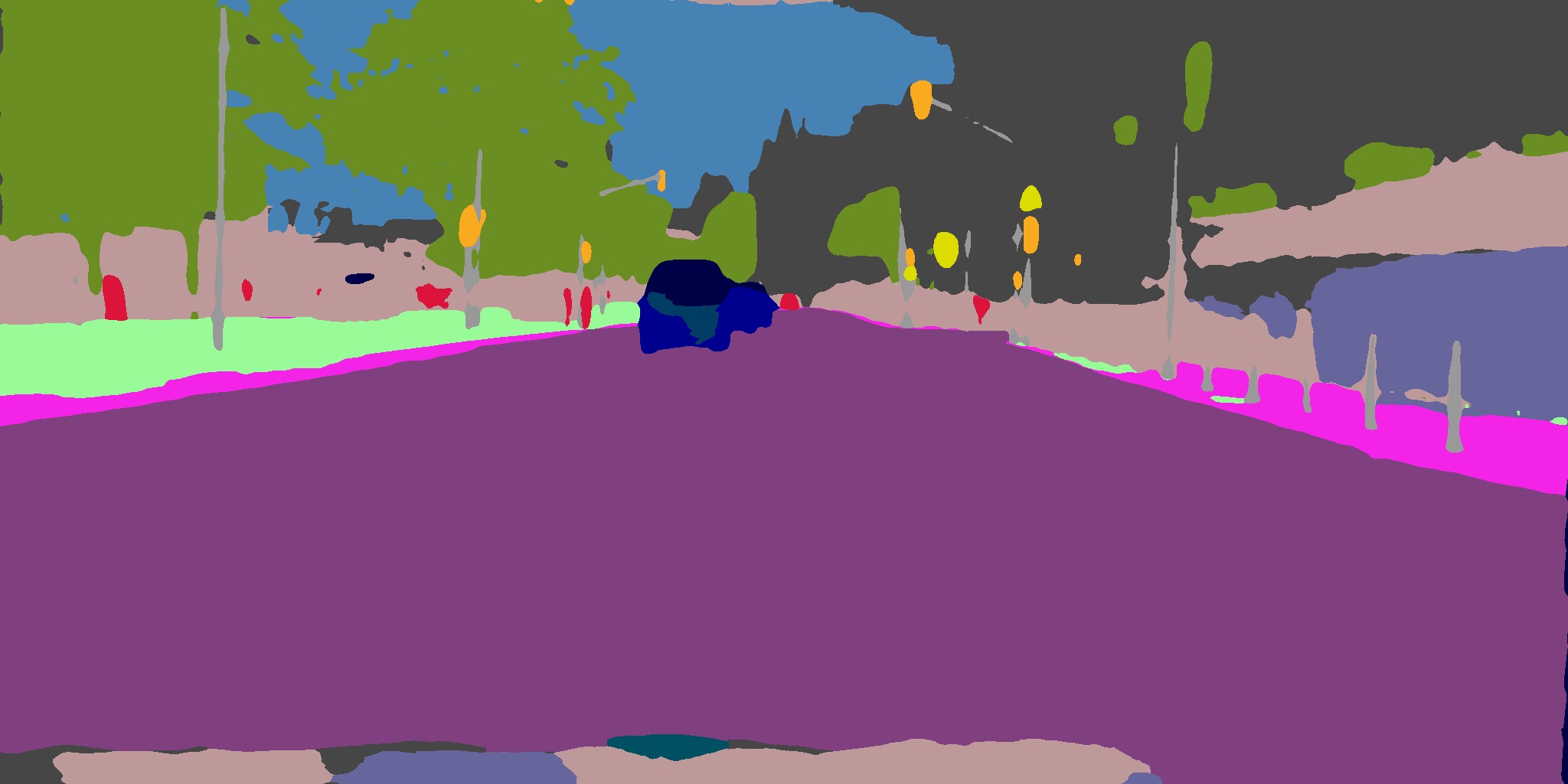}
        \end{minipage}
        \caption{ProDA}
    \end{subfigure}
    \hspace{-1.2mm}
    \begin{subfigure}{0.245\linewidth}
        \begin{minipage}[b]{1\linewidth}
            \includegraphics[width=1\linewidth]{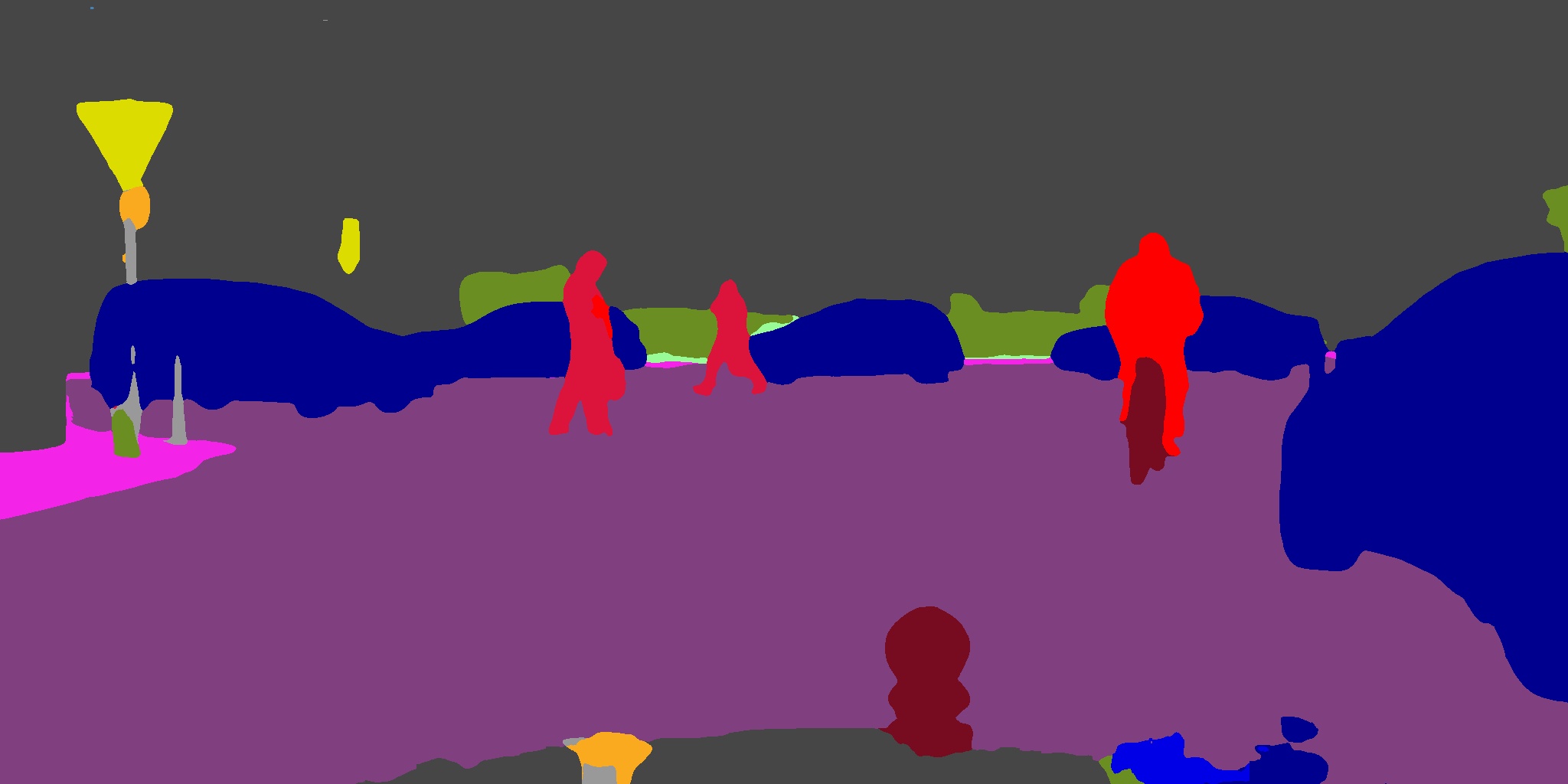}\\
            \includegraphics[width=1\linewidth]{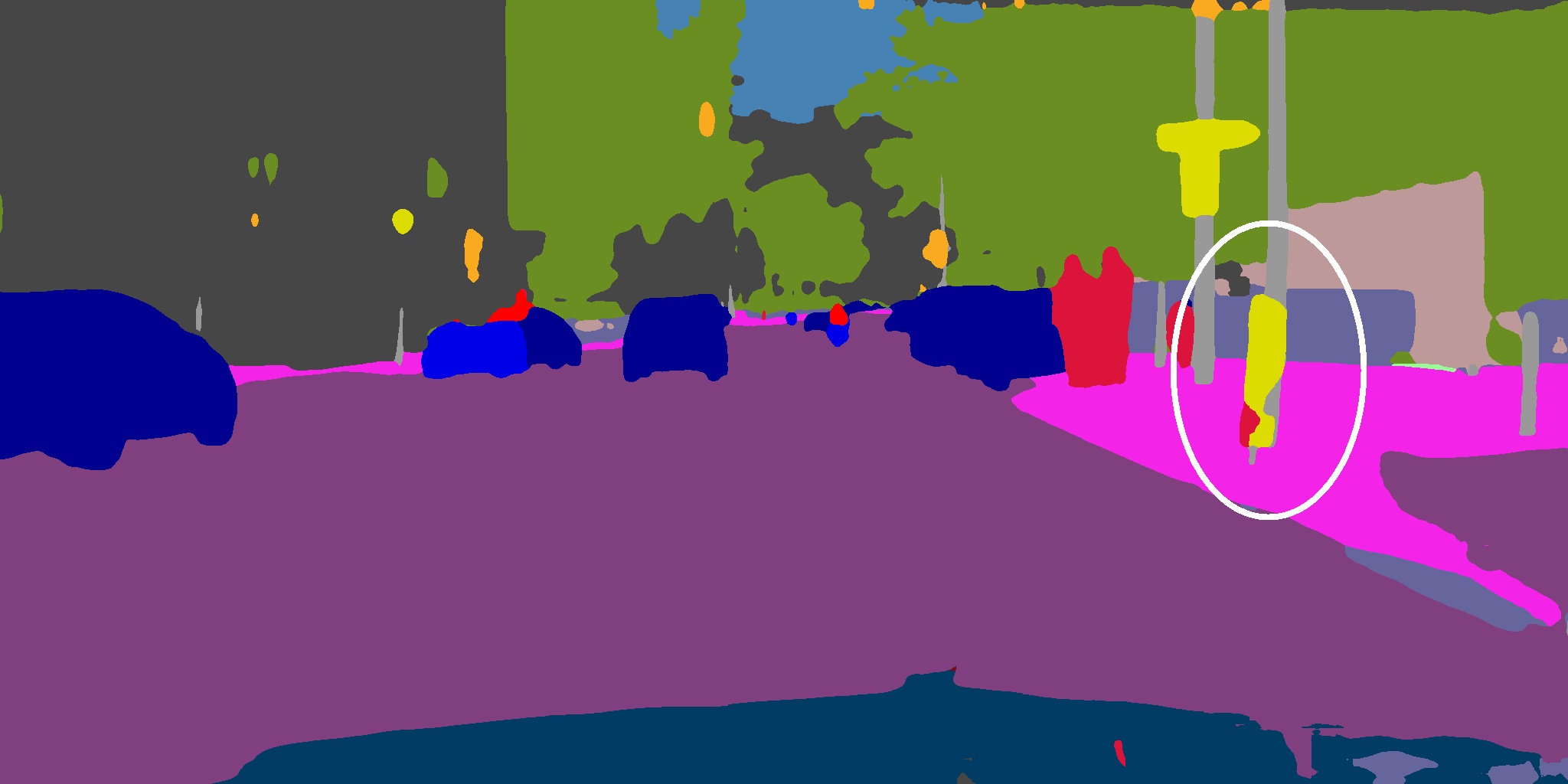}\\
            \includegraphics[width=1\linewidth]{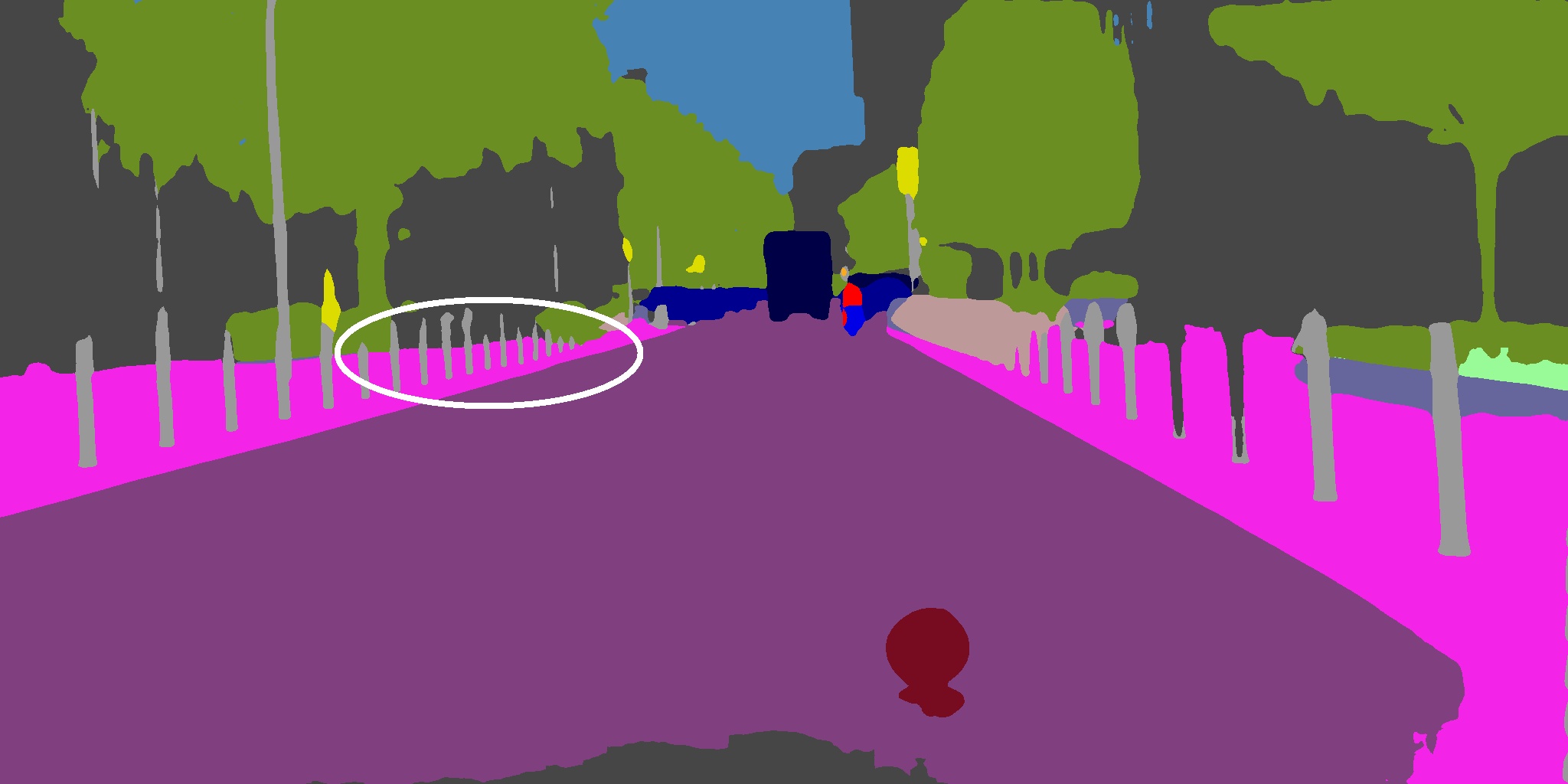}\\
            \includegraphics[width=1\linewidth]{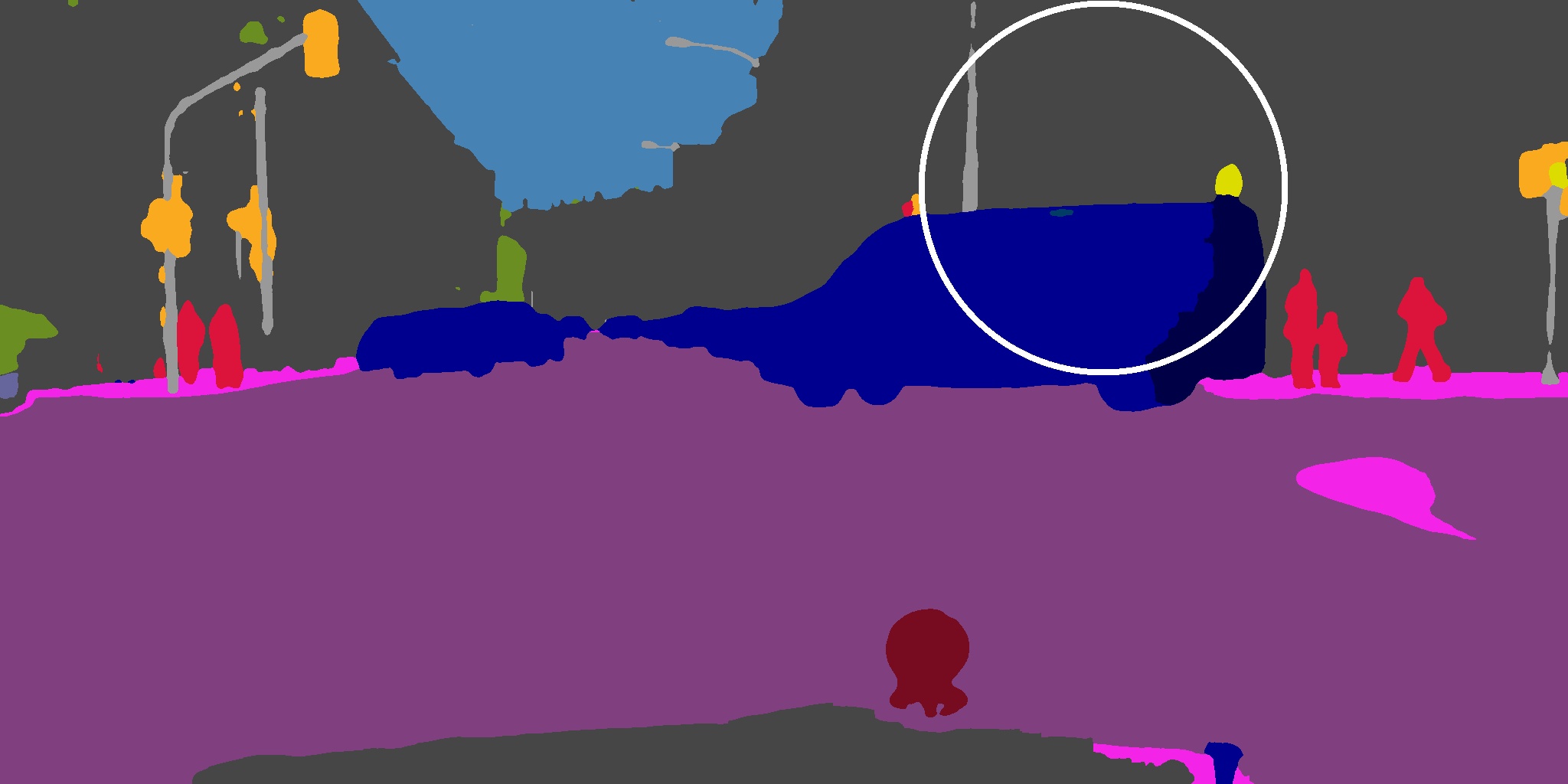}\\
            \includegraphics[width=1\linewidth]{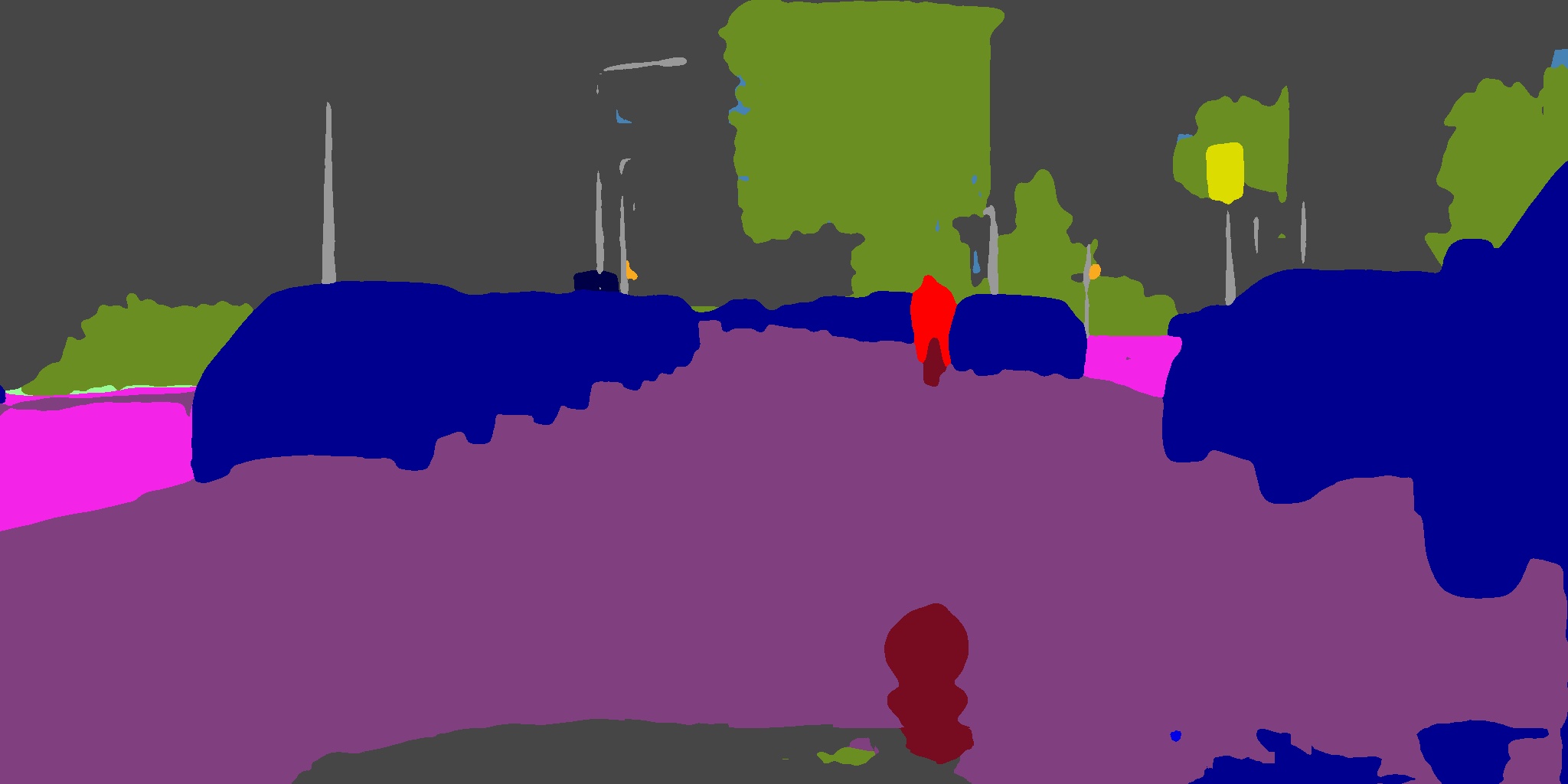}\\
            \includegraphics[width=1\linewidth]{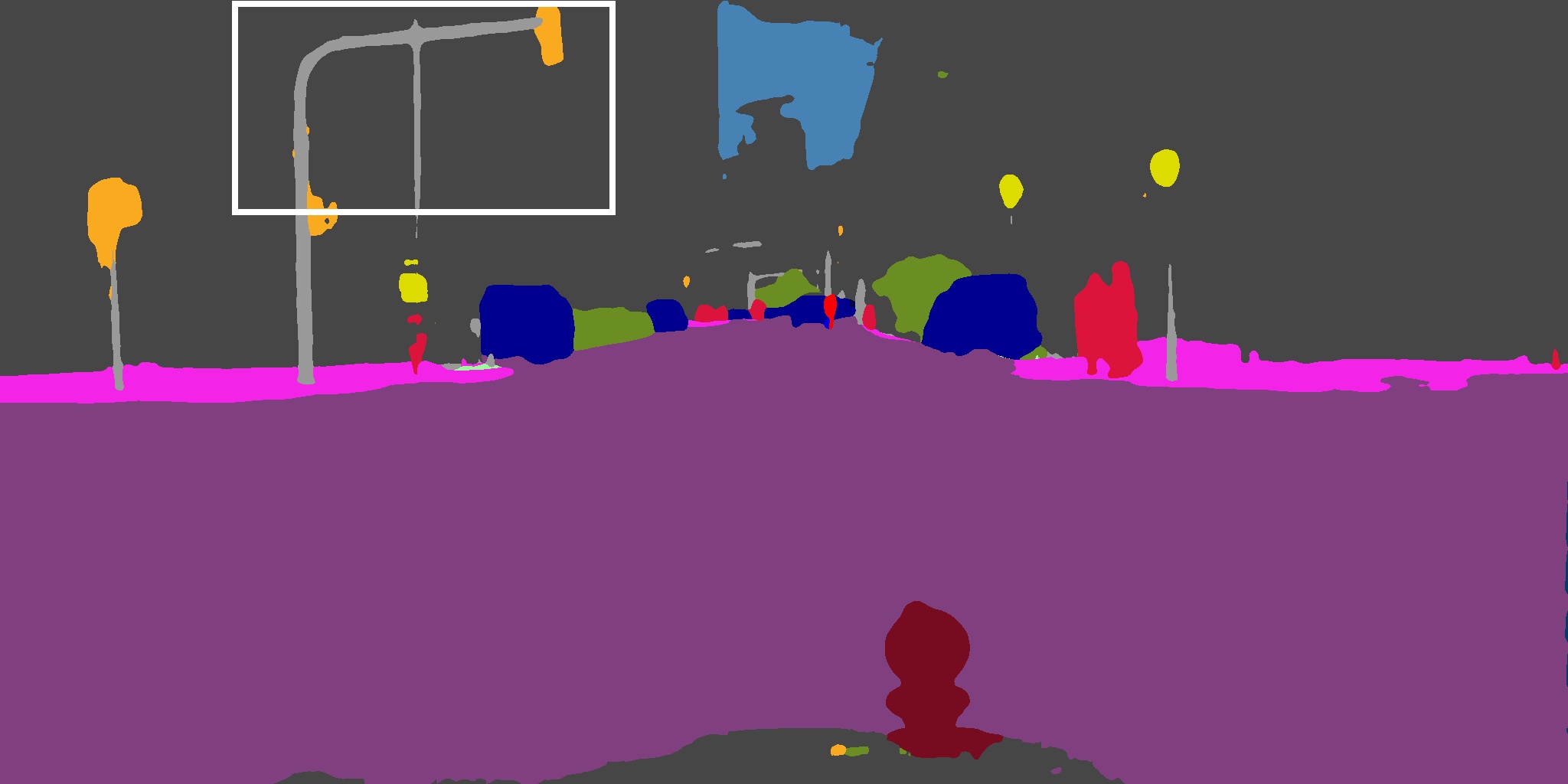}\\
            \includegraphics[width=1\linewidth]{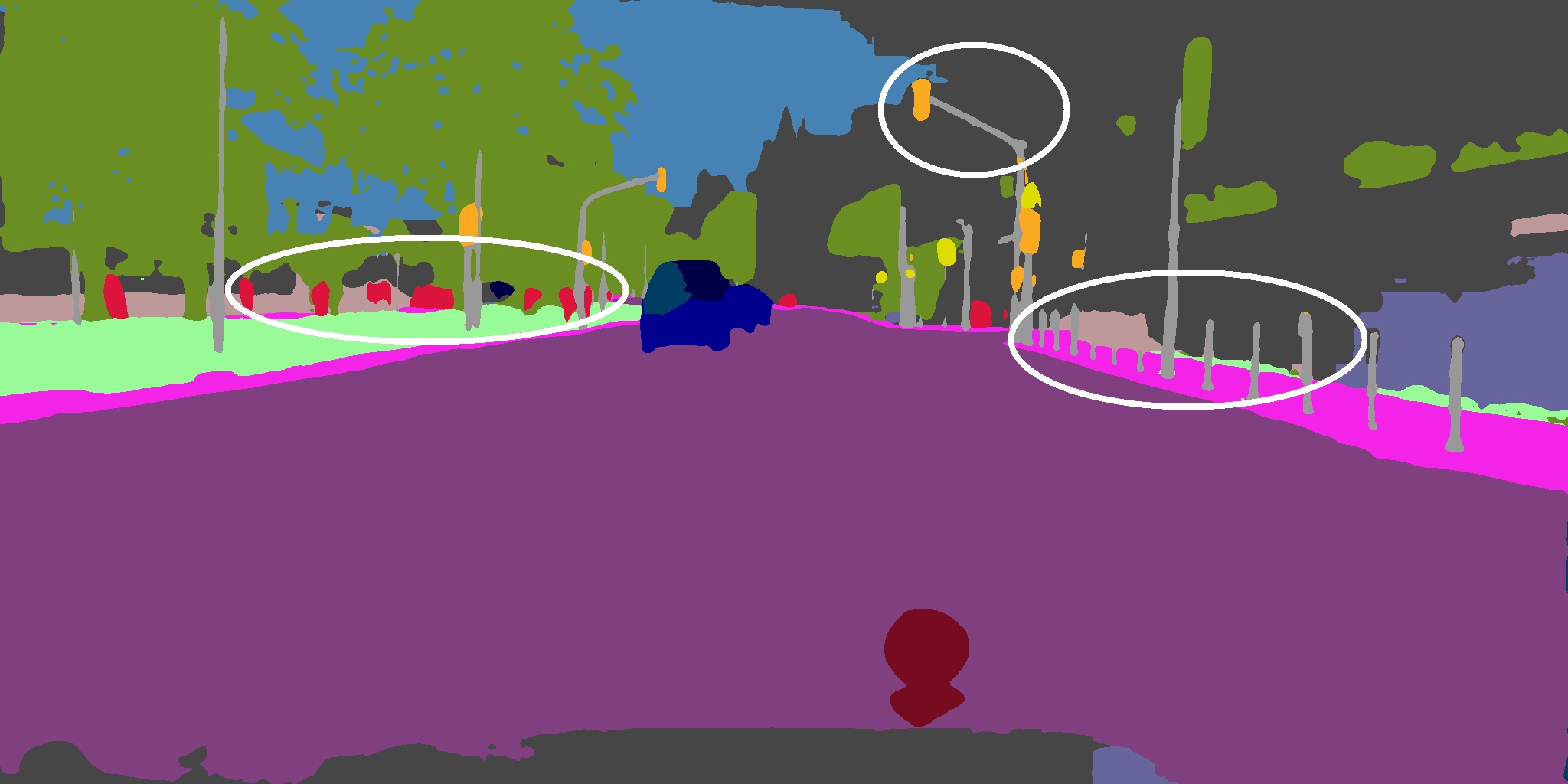}
        \end{minipage}
        \caption{Ours}
    \end{subfigure}
    \hspace{-1.2mm}
    \begin{subfigure}{0.245\linewidth}
        \begin{minipage}[b]{1\linewidth}
            \includegraphics[width=1\linewidth]{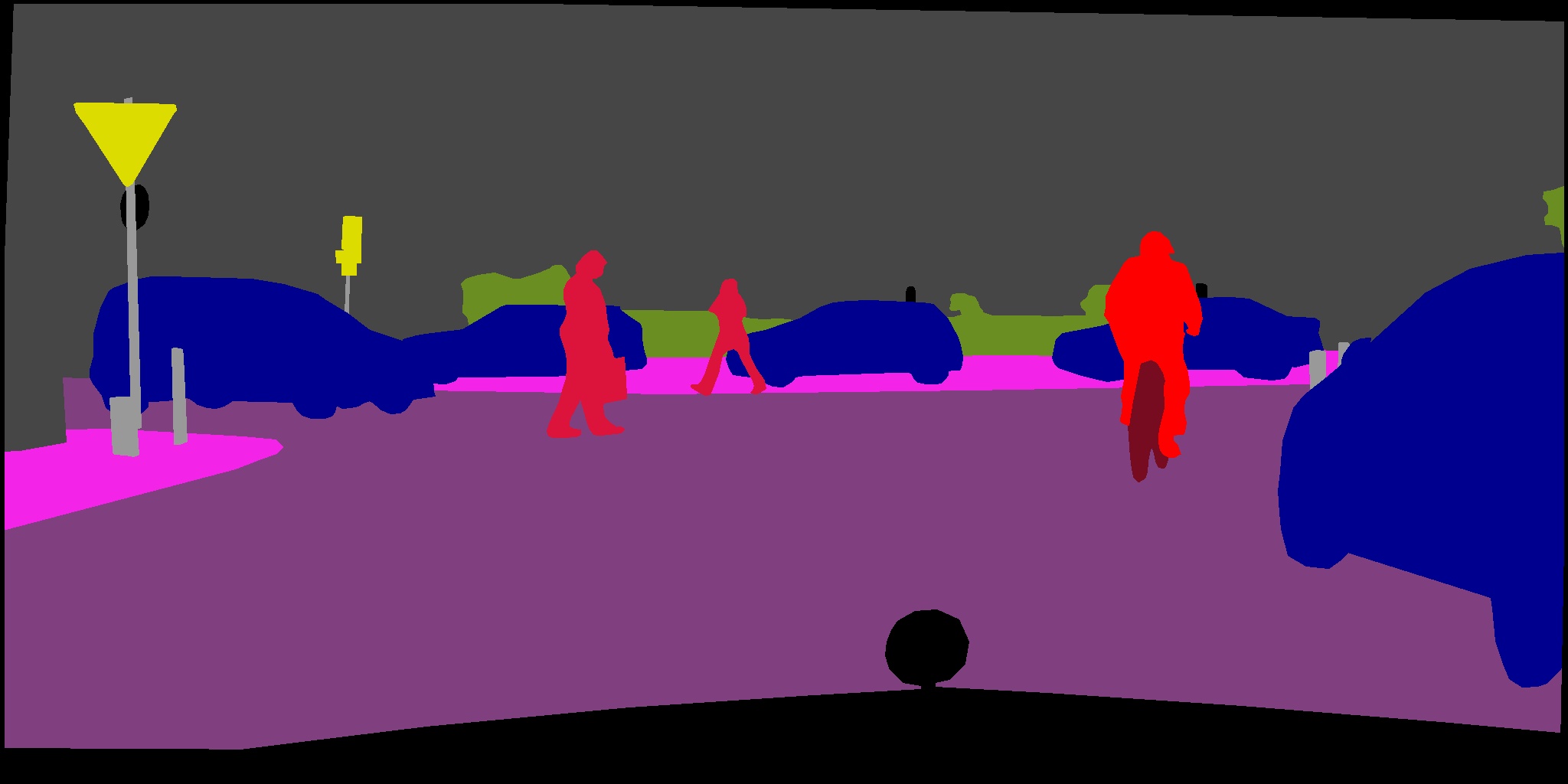}\\
            \includegraphics[width=1\linewidth]{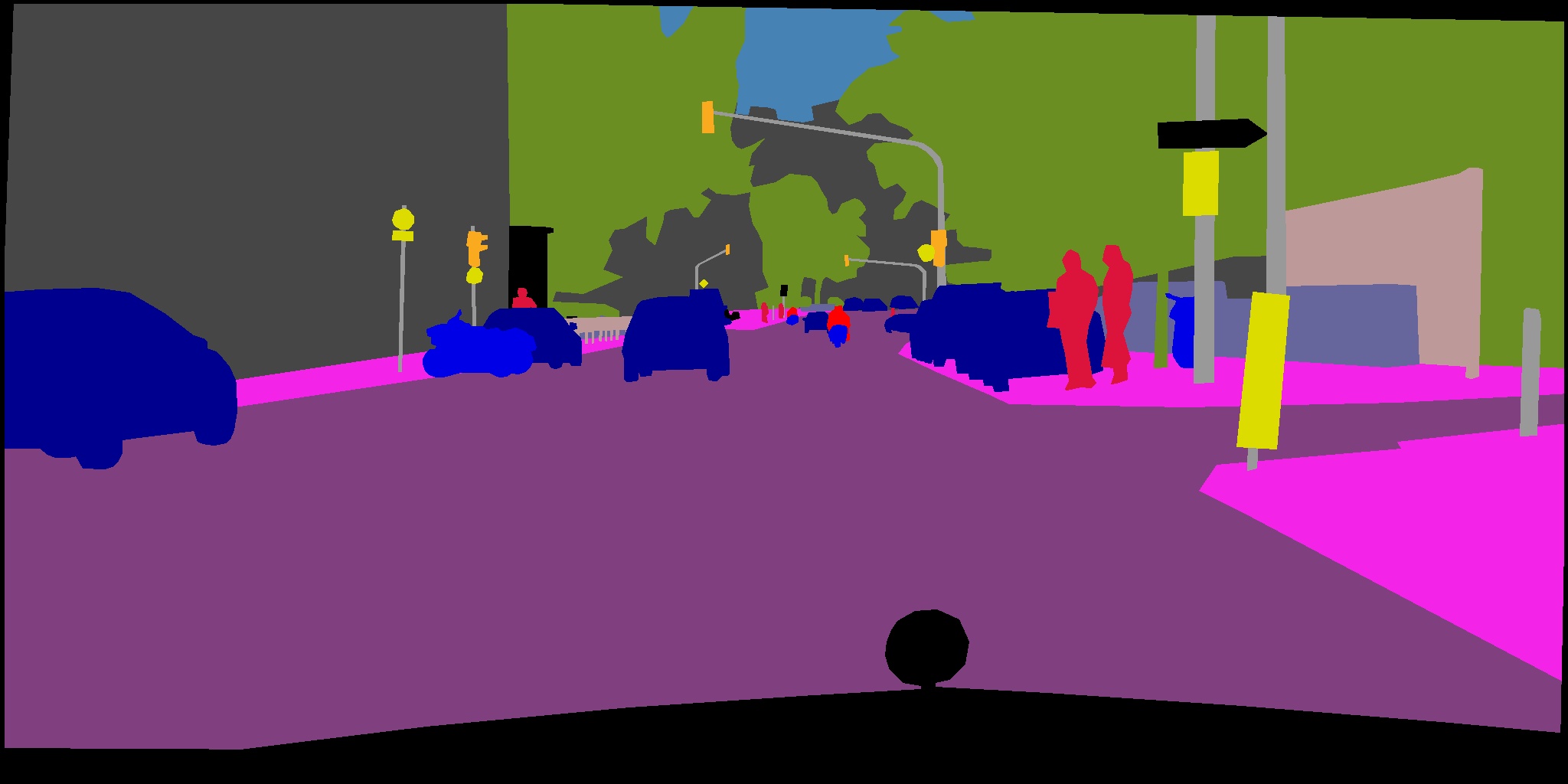}\\
            \includegraphics[width=1\linewidth]{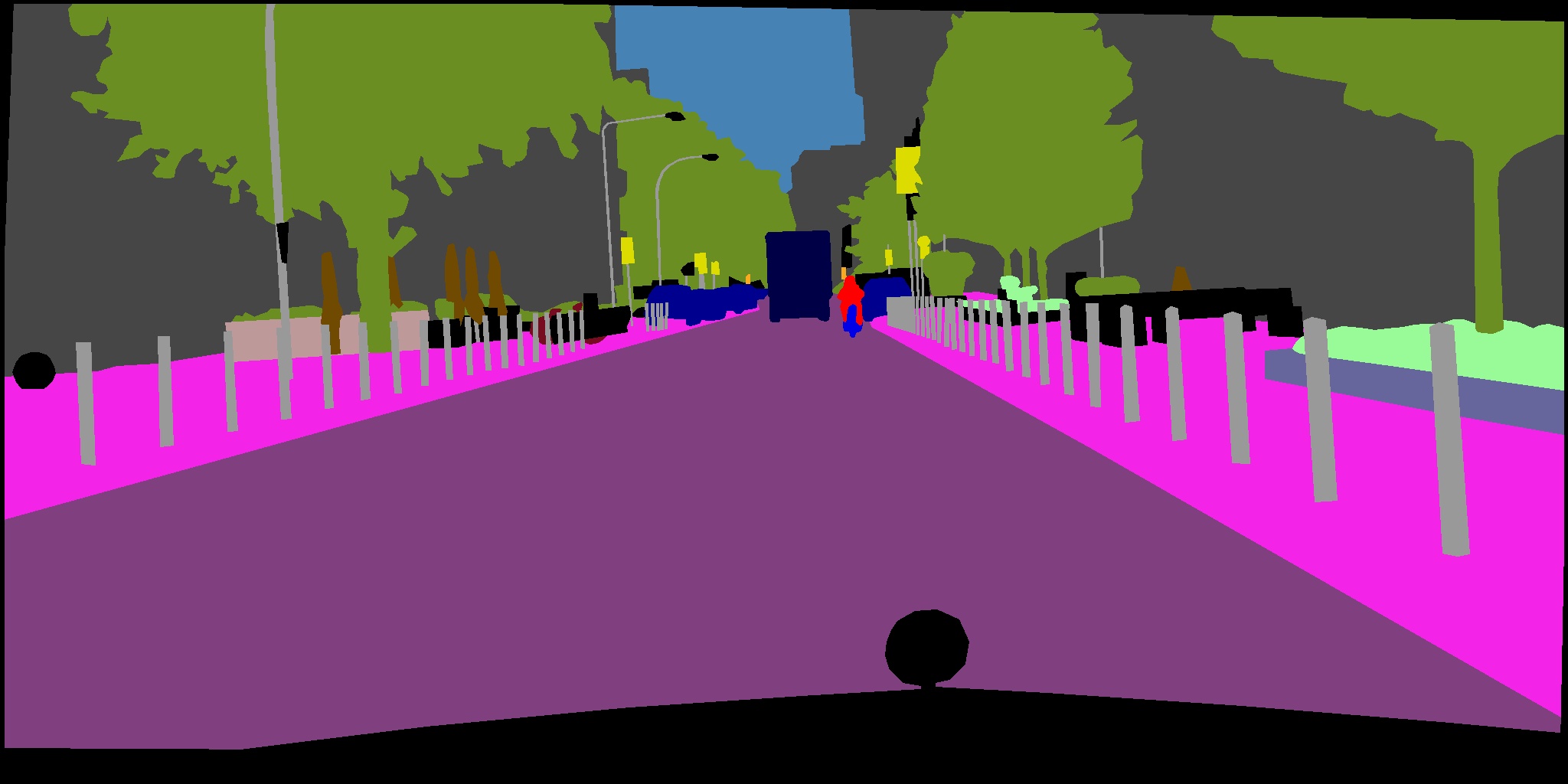}\\
            \includegraphics[width=1\linewidth]{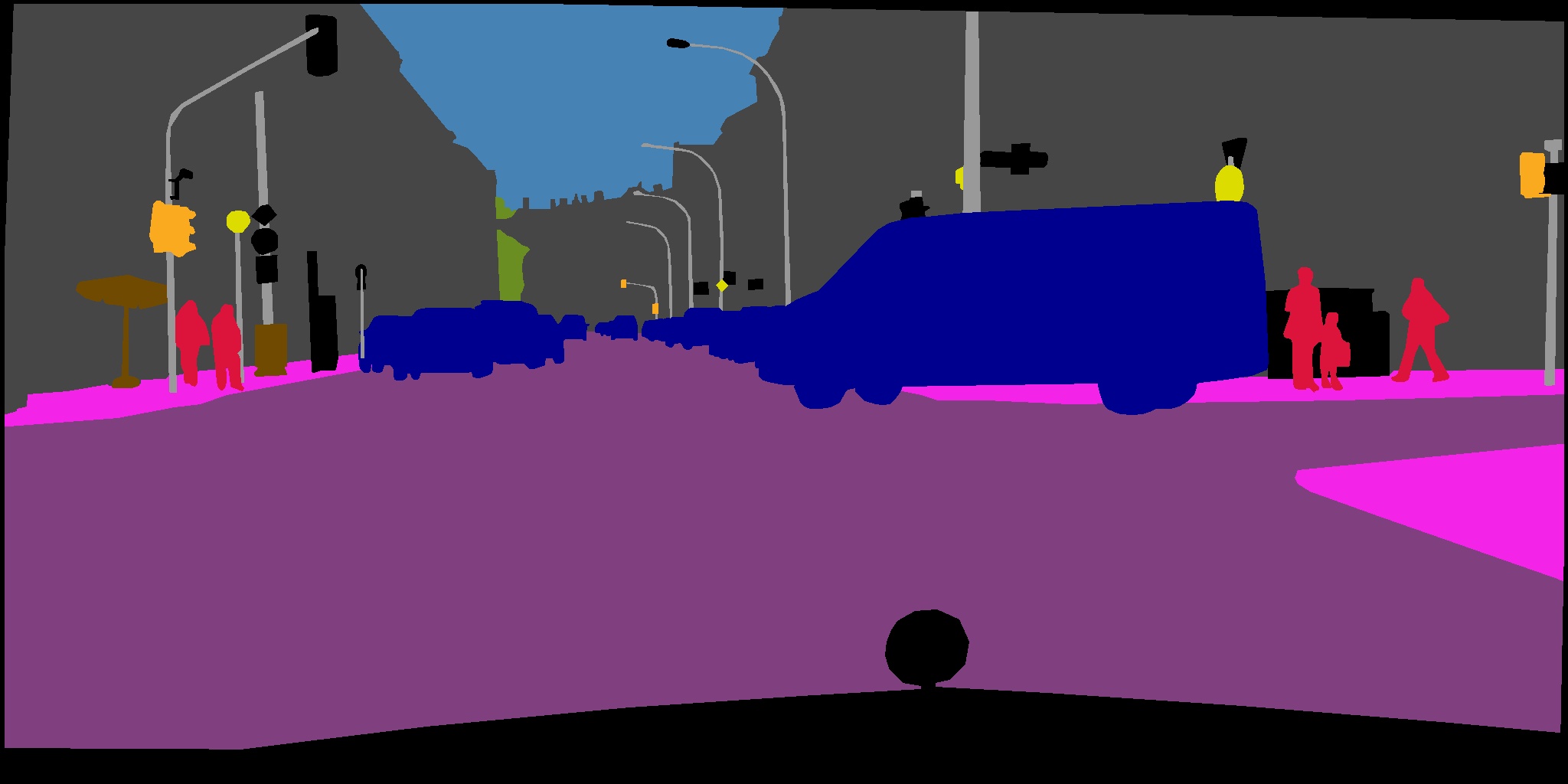}\\
            \includegraphics[width=1\linewidth]{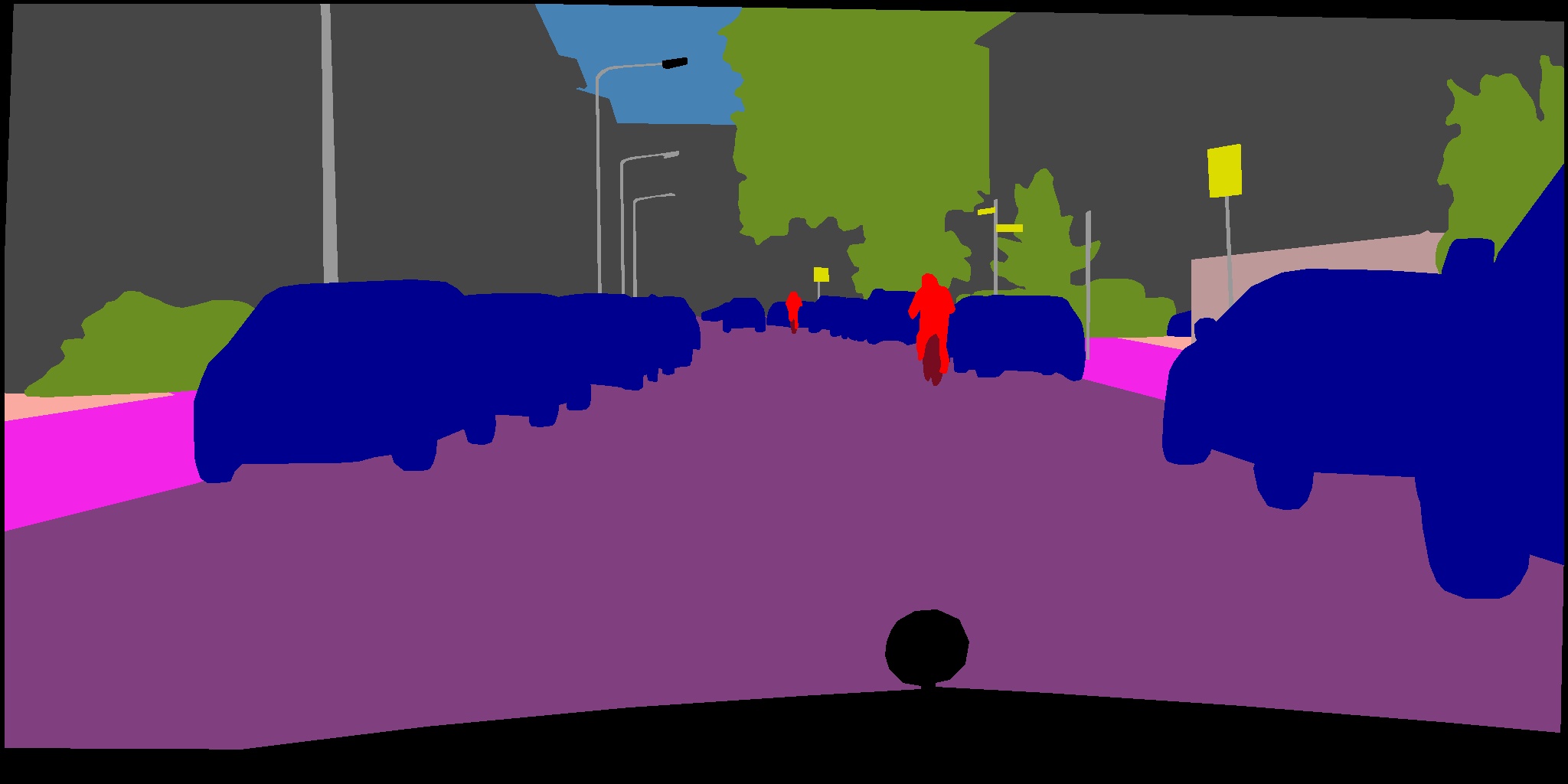}\\
            \includegraphics[width=1\linewidth]{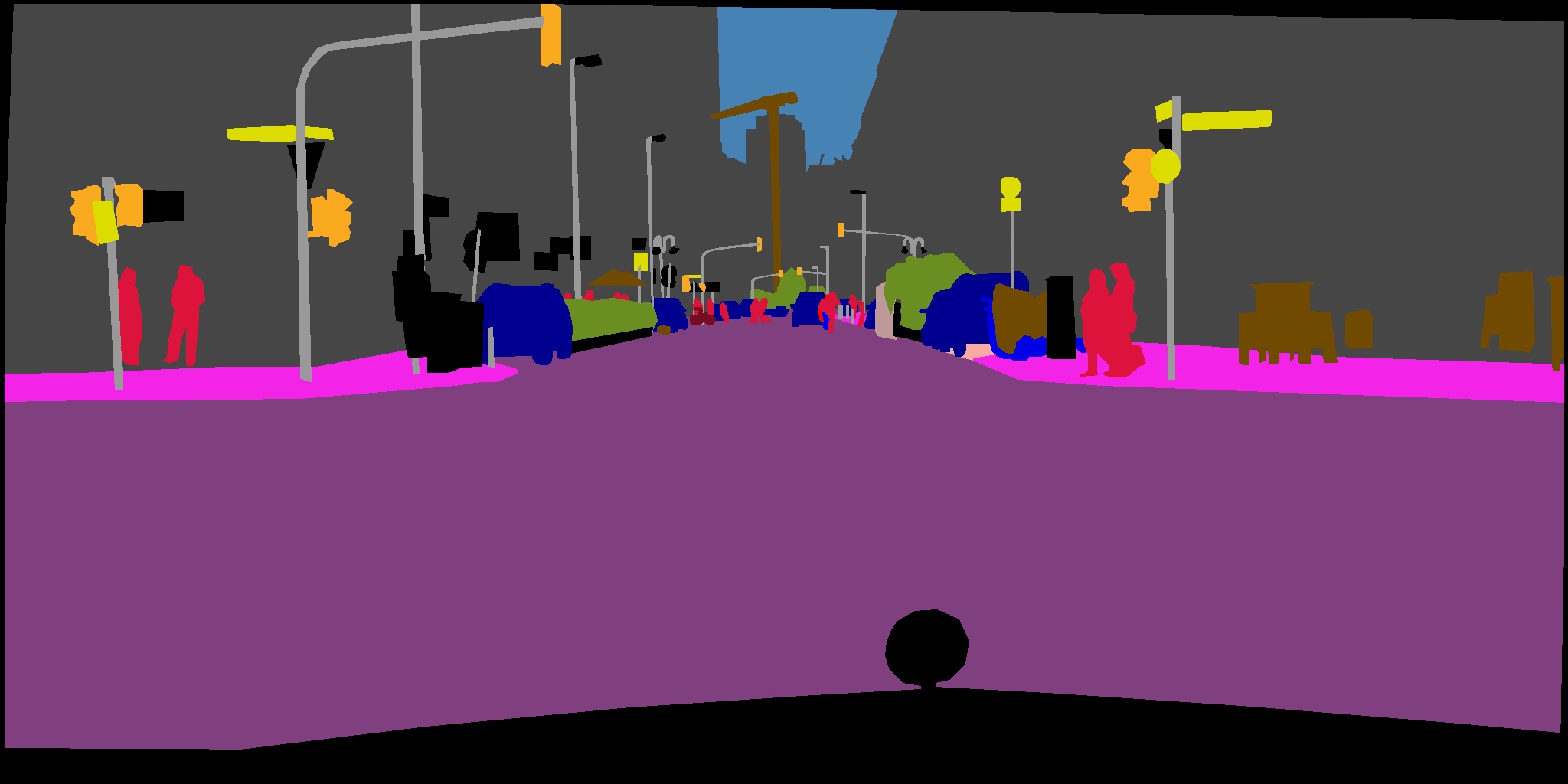}\\
            \includegraphics[width=1\linewidth]{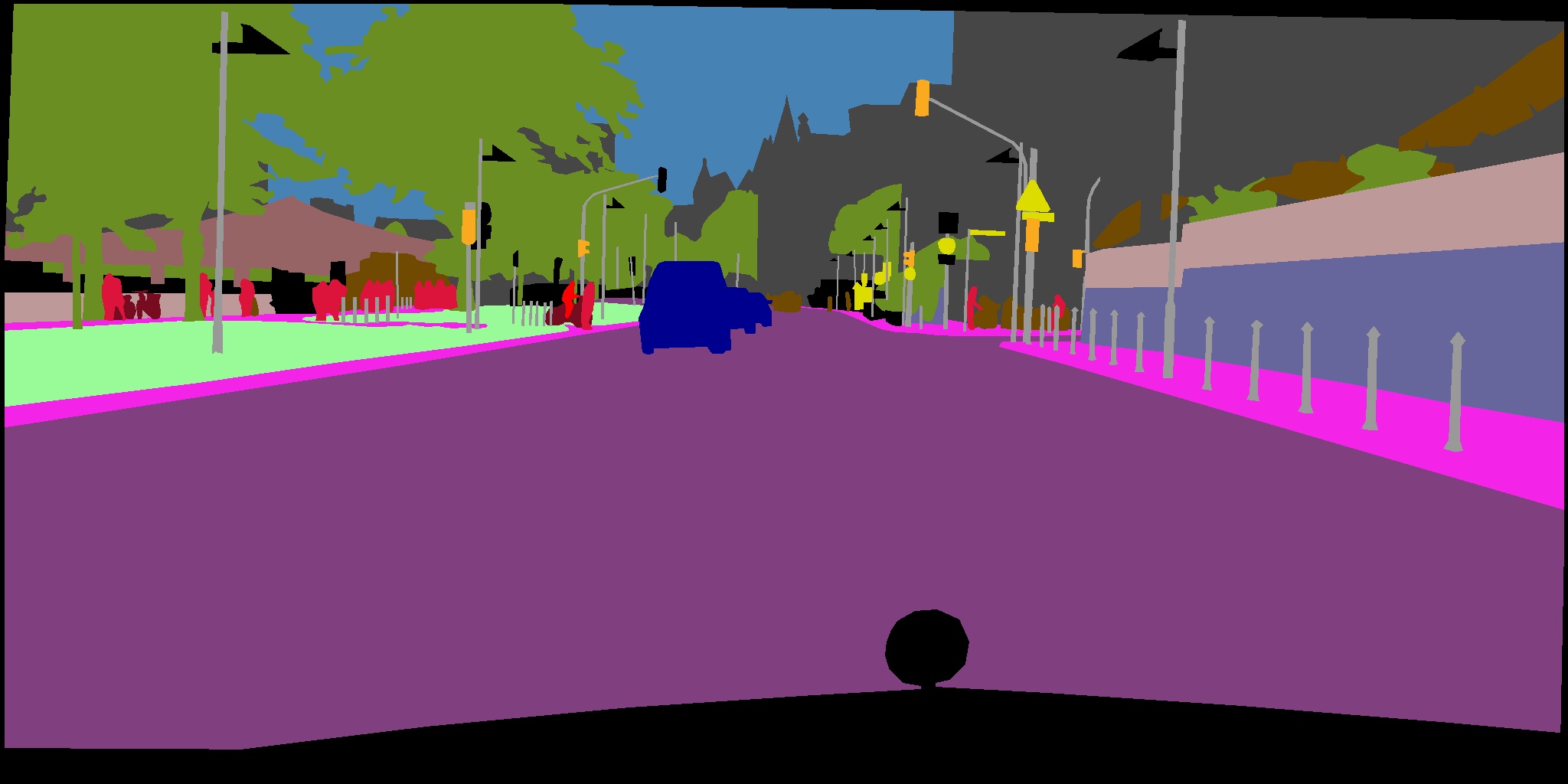}
        \end{minipage}
        \caption{Ground truth}
    \end{subfigure}
	\caption{\small  Comparison to prior work. (a) Input images from Cityscapes. (b) ProDA \cite{Zhang2021:CVPR}. (c) Results of our method. (d) Ground truth. As highlighted by the white circles and rectangles, the improvements are mainly in the ``hard classes'' (pole, traffic signs, traffic light and person \etc)}
	\label{fig:illustrate}
\end{figure*}
\mypara{Comparisons to existing work.}
\Tab~\ref{tab:gta2city} and \Fig~\ref{fig:illustrate} show the results when adapting from GTA5 to Cityscapes. Our approach outperforms existing work by a large margin in this experiment. It shows a mIoU of 60.2\% and achieves the highest IoU on 7 out of 19 classes. 6 of these classes are what can
be considered ``hard classes'' (\emph{pole}, \emph{traffic light}, \emph{person}, \emph{rider}, \emph{motorcycle}, and \emph{bicycle}) due to their small footprint in the dataset or because
of large per-class domain shift. The performance improvement of our method mostly comes from these challenging classes. 
Table \ref{tab:synthia2city} shows a similar experiment where we adapt from a different source 
domain (SYNTHIA) to Cityscapes. Our method again achieves state-of-the-art results with an mIoU of 56.5\% and 63.1\% for 16 and 13 classes, respectively. We again see a strong performance increase on hard classes, such as \emph{wall},  \emph{fence}, and \emph{motorcycle}.
Finally, we show a comparison to recent state-of-the-art methods on adaptation to a different target domain (GTA5 to Mapillary Vistas) in the supplement. Again, our method consistently outperforms prior work.

\mypara{Ablation.}\label{sec:ablation}
We show the influence of different components of our approach on the GTA5 to Cityscapes adaptation task in
Table~\ref{tab:modules}. For the pre-training step, both style transfer (\emph{transfer}) and contrastive learning (\emph{pre-training}) provide large performance improvements. These components alone improve the baseline from 37.6\% to 50.1\%. Contrastive learning further improves the mIoU from 50.1\% to 57.9\%.  Our approach without label expansion already outperforms all existing works. 
Enabling label expansion leads to an additional performance boost of more than 2.3\%.
Our ablation also shows that the label expansion step strongly benefits from contrastive learning, as contrastive learning yields more high confidence predictions particularly for hard classes (Table \ref{tab:hard_classes}).  
When we use label expansion without the contrastive feature alignment, we see a moderate improvement of 1.8\% compared to the relatively weak baseline that shows 50.1\% mIoU. When using label expansion together with contrastive feature alignment, we see an improvement from 57.9\% to 60.2\%. That is, we see a larger improvement over a significantly stronger baseline.
We also evaluate our method with different network backbones and 
provide additional ablations of various hyper-parameters in the supplementary material.
\begin{table}[ht]
	\centering
	\resizebox{0.95\linewidth}{!}{
	\begin{tabular}{c| c c c c c}
		\hline
		 condition & transfer& pre-training & contrast & label expansion & mIoU \\
		\hline
		 1 & & & & & 37.6\\
		 2 & $\surd$ & & & & 45.7\\
		 3 & $\surd$ &$\surd$ & & & 50.1\\
		 4 & $\surd$& $\surd$ &  & $\surd$ & 51.9\\
		5 & & $\surd$ & $\surd$ & $\surd$ & 57.5\\
		6 & $\surd$ &$\surd$ &$\surd$ & & 57.9\\
		\midrule
		7 & $\surd$ & $\surd$ & $\surd$ & $\surd$ & {\bf 60.2}\\
		\hline
	\end{tabular}}
    \vspace{-0.5em}
	\caption{Ablation study on GTA5$\rightarrow$Cityscapes. We find that all of our contributions improve model performance and our full model (\#7) performs best. Interestingly, label expansion is boosted through a contrastive objective (compare \#3 $\rightarrow$ \#4 vs. \#6 $\rightarrow$ \#7).}
	\label{tab:modules}
\end{table}

\mypara{Effect of contrastive learning.} 
The effect of contrastive learning on feature alignment across domains is shown in Figure~\ref{fig:feature_align}. We visualize the learned feature space of a subset of 
classes from the GTA5 and Cityscape domains using UMAP \cite{mcinnes2020umap}.
The in-domain contrastive learning concentrates samples close to their category centers while pushing the category centers apart. This leads to a well separated feature 
space for every class inside the respective domain. However, when applying the same feature 
extractor to a different domain, we observe that this separation is lost.
By adding cross-domain contrastive learning the differences between domains can be mitigated, by aligning the features that correspond to individual classes across domains.

\mypara{Effect of label expansion.} 
As illustrated qualitatively in Figure \ref{fig:pseudo_labels}, label expansion increases the number of pixels with valid pseudo-labels from hard classes (see the traffic sign and the bicycle highlighted in Figure \ref{fig:pseudo_labels} (d)). This additional data positively affects learning and boosts final mIoU from 57.9\% to 60.2\% in our GTA5$\rightarrow$Cityscapes  experiment (see Table~\ref{tab:modules}). 
It  particularly improves results on hard classes. As shown in Table \ref{tab:hard_classes},  \emph{traffic light}, \emph{sign}, and \emph{bicycle} are improved by more than 10\% whereas \emph{motorcycle} and \emph{train} are improved by 20$\sim$30\% when compared to the baseline without label expansion.

\begin{table}[t]
	\centering
	\resizebox{1\linewidth}{!}{
	\begin{tabular}{l l l l l l l l l l}
		\hline
		& \rotatebox{90}{pole} & \rotatebox{90}{light} & \rotatebox{90}{sign} &  \rotatebox{90}{person} & \rotatebox{90}{rider}  & \rotatebox{90}{train} & \rotatebox{90}{m.cycle} & \rotatebox{90}{bicycle} & mIoU \\
		\hline
       w/o expansion  & 49.2 &  48.5 & 45.8 &   70.2 &  41.4&  21.5 & 40.1&  53.4& 47.5 \\
       w/ expansion & 52.9 & 53.6 &  54.1 & 73.5 & 44.1 & 26.8 & 51.6 & 61.8 &  53.1\\
       improvements (\%) & 7.5 & 10.5  &18.1 &  4.7  & 6.5 & 24.7 & 28.7  & 15.7 &  11.8\\
		\hline
	\end{tabular}}
    \vspace{-0.5em}
	\caption{Improvements over ``hard classes'' after implementing the pseudo label expansion. }
	\label{tab:hard_classes}
\end{table}

\section{Conclusion}
We introduced contrastive learning for unsupervised domain adaptation in semantic segmentation. We leverage both in-domain contrastive samples as well as cross-domain contrastive samples that bridge the source and target domains. Our approach achieves robust class-based feature alignment to
facilitate domain adaptation. Our framework is based on a student-teacher architecture that generates pseudo-labels, which guide the selection of contrastive pairs. We introduced a label expansion strategy to discover reliable pixels from particularly hard classes.
Our method achieves state-of-the-art domain adaptation results for a variety of 
source and target domains. It significantly improves results for hard classes where only a few pixels are available in the source domain or ones that are affected by strong class-specific domain shift between the domains.



\appendix
\section{Implementation details}\label{sec:supp_implementation}

\paragraph{Confidence thresholds for generating pseudo labels.}
To determine a suitable confidence threshold for considering predictions by the teacher network as pseudo labels, we define class-dependent confidence thresholds $\tau_c$.
We set $\tau_c = min(\tau_0, \tau_p)$, where $\tau_0 = 0.9$, and $\tau_p$ represents the confidence level for which the top 10$\%$ of predictions per class and batch by the teacher network are considered confident predictions.

\paragraph{Augmentation.}
Table \ref{table:augmentations} lists the hyperparameters for the augmentations we use for generating input views.
\begin{table}[htb!]
	\centering
		\begin{tabular}{ll}
			\toprule
        Type of augmentation & Parameters\\
        \midrule
        ColorJitter & (0.4, 0.4, 0.4, 0.1)\\
        Grayscale & $p = $0.2\\
        CutOut& [0, 40]\\the 
        Scaling & [0.4, 2.5]\\
        Rotation & [-45, 45]\\
        Crop & [713, 713]\\
        Horizontal Flip& $p = $0.5 \\
		\bottomrule
		\end{tabular}
	\caption{Augmentation parameters for our domain contrastive segmentation. $p$ denotes the probability to convert a color image to {\em grayscale} or to {\em horizontally flip}. For {\em CutOut}, [0, 40] is the size in pixels of the square patch that is cut out.}
	\label{table:augmentations}
\end{table}

\section{Additional experiments}\label{sec:supp_experiments}
\subsection{Comparison to prior work}
In addition to the comparisons in the main paper, we further evaluate on the Mapillary Vistas dataset~\cite{neuhold2017mapillary}. In contrast to Cityscapes, Vistas consists of 18k training images captured by a broad range of cameras and in a more diverse set of locations. The label set of Mapillary Vistas differs from the label set of other datasets used in our experiment. Hence, we map the 65 classes of Vistas to the 19 classes of Cityscapes, following the evaluation protocol of He~\etal.~\cite{he2020segmentations}.
We compare against a number of challenging baselines in Table~\ref{tab:gta2mapillary}, including ProDA, which represents the current state-of-the art for domain adaptation~\cite{Zhang2021:CVPR}. Again, our method consistently outperforms prior work.
\begin{table}[ht]
	\centering
	\begin{tabular}{l c}
		\toprule
		 Method & mIoU \\
		 \midrule
		 Coarse-to-fine~\cite{Ma2021:CVPR}& 55.7  \\
		 ProDA~\cite{Zhang2021:CVPR}& 58.9 \\
		 \midrule
		 Ours & 62.1\\
		\bottomrule
	\end{tabular}
    \vspace{-0.5em}
	\caption{Comparison to prior work on GTA5$\rightarrow$Mapillary \cite{neuhold2017mapillary}. All methods use DeepLabV2 (ResNet101).}
	\label{tab:gta2mapillary}
\end{table}

\subsection{Controlled experiments}

\paragraph{Effect of hyperparameters.}
\begin{table}[h]
	\centering
	\begin{tabular}{c c c c}
		\hline
		 $\alpha$& $\lambda$ & $\tau_0$ & mIoU \\
		\hline
		 0.1 &  0.1 & 0.9 &60.2\\
		 \hline
		 0.05 &  - & - & 59.9\\
		 0.2 & - & - & 60.0\\
		 0.4 & - & - & 59.4\\
		 \hline
		 - & 0.05 & - & 59.7\\
		 - & 0.2 & - & 59.6\\
		 - & 0.4 & - & 59.4\\
		 \hline
		 - & - & 0.5 & 58.5\\
		 - & - & 0.7 & 59.5\\
		 - & - & 0.8 & 60.0\\
		 - & - & 0.95 & 59.3\\
		\hline
	\end{tabular}
    \vspace{1mm}
	\caption{Controlled experiment assessing the effect of hyperparameters. We vary the hyperparameters $\lambda$ and $\alpha$, which balance factors of the training objective, and the base threshold $\tau_0$ used for extracting confident predictions from the teacher network. ``$-$'' means default settings.}
	\label{tab:balance_weights}
\end{table}
We further test the sensitivity of the balance weights in the loss function. 
We analyze the impact of parameters including the balance weights $\lambda$ and $\alpha$ in the loss function, and the initial threshold $\tau_0$ for pseudo label generation. In Table \ref{tab:balance_weights}, we use different values for the loss function and different thresholds to select the pseudo labels. The results are slightly worse than the default setting when using different hyper parameters. We find that the method is not very sensitive to these user defined parameters. 

\paragraph{Effect of network architecture.}
\begin{table}[t]
	\centering
	\begin{tabular}{l l l c}
		\toprule
		 Method & Network & Backbone & mIoU \\
		 \midrule
		 Ours& PSPNet & ResNet-50 &57.8\\
		\midrule
		 Coarse-to-fine\cite{Ma2021:CVPR}& DeepLabV3 & ResNet 101 & 56.1\\
		 Ours & DeepLabV3 & ResNet-50 & 58.4\\
		 \midrule
		 ProDA\cite{Zhang2021:CVPR}& DeepLabV2 & ResNet-101 &57.5 \\
 		 Ours & DeepLabV2 & ResNet-101 & 60.2 \\
		\bottomrule
	\end{tabular}
    \vspace{-0.5em}
	\caption{Comparison of different architectures on GTA5$\rightarrow$Cityscapes.}
	\label{tab:results_architectures}
\end{table}
We evaluate our method with different network backbones and compare to baselines with the same or stronger backbones in Table \ref{tab:results_architectures}. In all conditions our work outperforms the prior work, even if our method is only trained with a weaker ResNet-50 backbone (Ours (PSPNet) and Ours vs. Coarse-to-fine)).

\section{Analysis of learned representations}
\begin{figure*}[t]
    \centering
    \includegraphics[width=0.32\linewidth]{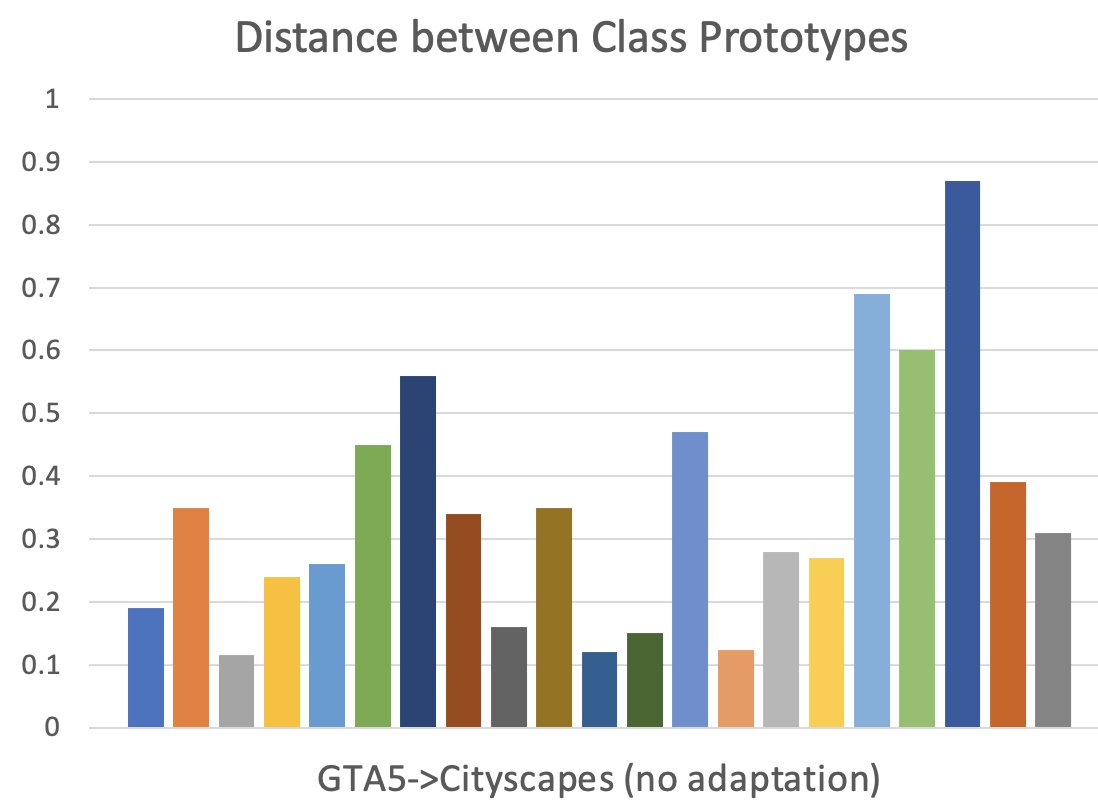}
    \includegraphics[width=0.32\linewidth]{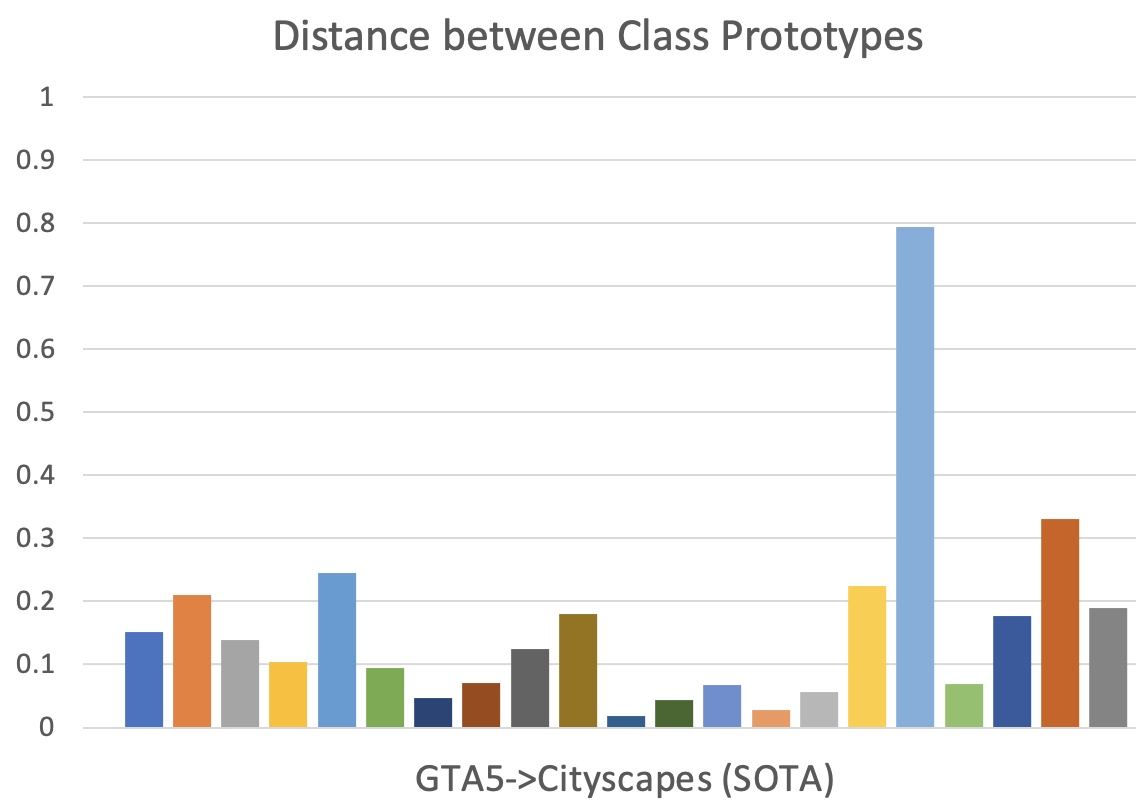}
    \includegraphics[width=0.32\linewidth]{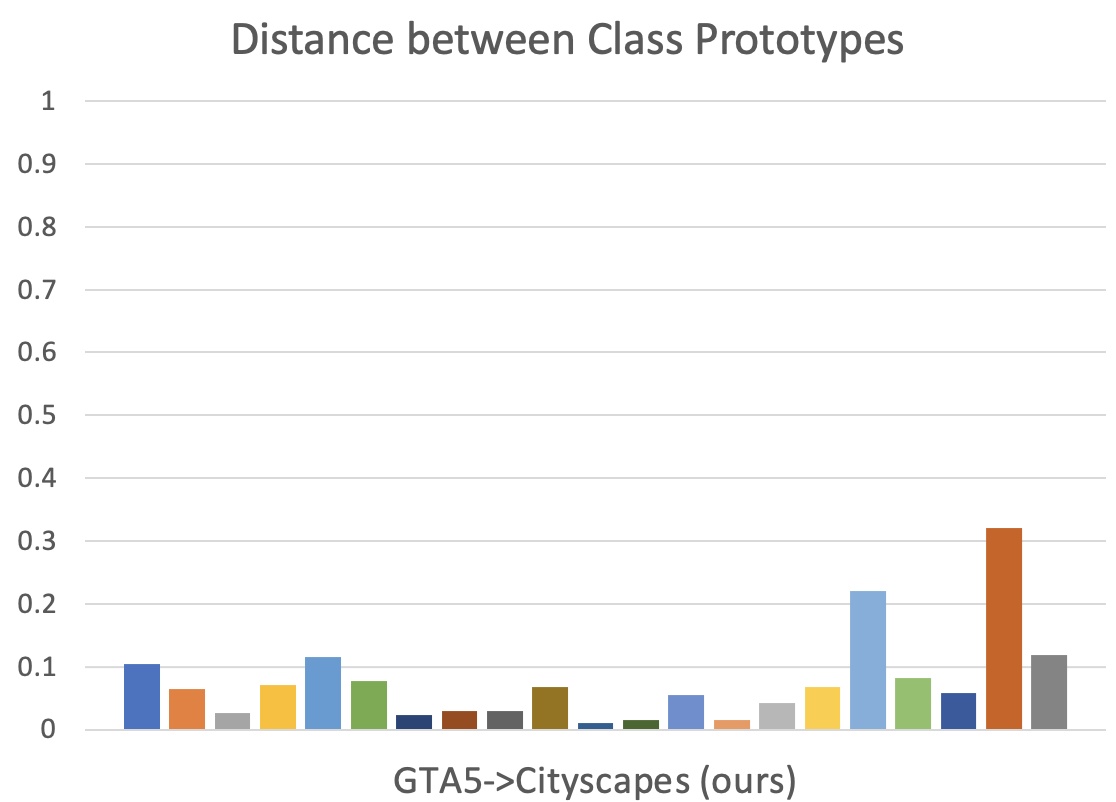}\\
    \includegraphics[width=0.8\linewidth]{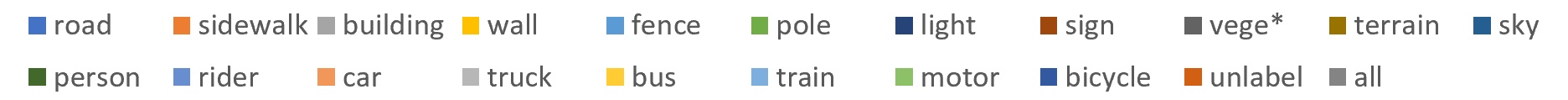}
	\caption{Analyzing feature alignment across domains for GTA $\rightarrow$ Cityscapes. We compute the $\lL_2$-distance between class prototypes across domains. A smaller distance indicates better alignment. Our learned representations (right) are consistently closer for all classes than the ones learned without domain adaptation (left) or by ProDA (middle).}
	\label{fig:feature_distance}
\end{figure*}
We analyze the alignment of class prototypes across domains for adapting GTA to Cityscapes. To that end, we unit-normalize the feature representations for each prototype and compute $\lL_2$-distances between respective class prototypes across domains. We compute these for learned representations without any domain adaptations in Figure~\ref{fig:feature_distance} (left), for the state-of-the-art ProDA~\cite{Zhang2021:CVPR} in Figure~\ref{fig:feature_distance} (middle), and for our method in Figure~\ref{fig:feature_align} (right).
We observe that the distances between prototypes across domains are consistently smaller for our method, which indicates that our learned feature representations are better aligned across domains. This is confirmed in our quantitative evaluations.

{\small
\bibliographystyle{ieee_fullname}
\bibliography{mybib}

\begin{thebibliography}{10}\itemsep=-1pt

\bibitem{Alonso2021:ICCV}
Inigo Alonso, Alberto Sabater, David Ferstl, Luis Montesano, and Ana~C Murillo.
\newblock Semi-supervised semantic segmentation with pixel-level contrastive
  learning from a class-wise memory bank.
\newblock In {\em ICCV}, 2021.

\bibitem{Araslanov2021:CVPR}
Nikita Araslanov and Stefan Roth.
\newblock Self-supervised augmentation consistency for adapting semantic
  segmentation.
\newblock In {\em CVPR}, 2021.

\bibitem{Berthelot2019:NeurIPS}
David Berthelot, Nicholas Carlini, Ian~J. Goodfellow, Nicolas Papernot, Avital
  Oliver, and Colin Raffel.
\newblock Mixmatch: {A} holistic approach to semi-supervised learning.
\newblock In {\em NeurIPS}, 2019.

\bibitem{Chaitanya2020:NeurIPS}
Krishna Chaitanya, Ertunc Erdil, Neerav Karani, and Ender Konukoglu.
\newblock Contrastive learning of global and local features for medical image
  segmentation with limited annotations.
\newblock In {\em NeurIPS}, 2020.

\bibitem{Chang2019:CVPR}
Wei{-}Lun Chang, Hui{-}Po Wang, Wen{-}Hsiao Peng, and Wei{-}Chen Chiu.
\newblock All about structure: Adapting structural information across domains
  for boosting semantic segmentation.
\newblock In {\em CVPR}, 2019.

\bibitem{Chen2021:arxiv}
Jiacheng Chen, Bin{-}Bin Gao, Zongqing Lu, Jing{-}Hao Xue, Chengjie Wang, and
  Qingmin Liao.
\newblock Scnet: Enhancing few-shot semantic segmentation by self-contrastive
  background prototypes.
\newblock {\em arXiv}, abs/2104.09216, 2021.

\bibitem{Chen2019:ICCV}
Minghao Chen, Hongyang Xue, and Deng Cai.
\newblock Domain adaptation for semantic segmentation with maximum squares
  loss.
\newblock In {\em ICCV}, 2019.

\bibitem{Chen2020:ICML}
Ting Chen, Simon Kornblith, Mohammad Norouzi, and Geoffrey~E. Hinton.
\newblock A simple framework for contrastive learning of visual
  representations.
\newblock In {\em ICML}, 2020.

\bibitem{Chen2020:NeurIPS}
Ting Chen, Simon Kornblith, Kevin Swersky, Mohammad Norouzi, and Geoffrey
  Hinton.
\newblock Big self-supervised models are strong semi-supervised learners.
\newblock In {\em NeurIPS}, 2020.

\bibitem{Chen2020:arxiv}
Xinlei Chen, Haoqi Fan, Ross~B. Girshick, and Kaiming He.
\newblock Improved baselines with momentum contrastive learning.
\newblock {\em arXiv}, abs/2003.04297, 2020.

\bibitem{Chen2017:ICCV}
Yi{-}Hsin Chen, Wei{-}Yu Chen, Yu{-}Ting Chen, Bo{-}Cheng Tsai,
  Yu{-}Chiang~Frank Wang, and Min Sun.
\newblock No more discrimination: Cross city adaptation of road scene
  segmenters.
\newblock In {\em ICCV}, 2017.

\bibitem{Chen2018:CVPR}
Yuhua Chen, Wen Li, and Luc~Van Gool.
\newblock {ROAD:} reality oriented adaptation for semantic segmentation of
  urban scenes.
\newblock In {\em CVPR}, 2018.

\bibitem{Chen2019:CVPR}
Yun{-}Chun Chen, Yen{-}Yu Lin, Ming{-}Hsuan Yang, and Jia{-}Bin Huang.
\newblock Crdoco: Pixel-level domain transfer with cross-domain consistency.
\newblock In {\em CVPR}, 2019.

\bibitem{Choi2019:ICCV}
Jaehoon Choi, Taekyung Kim, and Changick Kim.
\newblock Self-ensembling with gan-based data augmentation for domain
  adaptation in semantic segmentation.
\newblock In {\em ICCV}, 2019.

\bibitem{cordts2016cityscapes}
Marius Cordts, Mohamed Omran, Sebastian Ramos, Timo Rehfeld, Markus Enzweiler,
  Rodrigo Benenson, Uwe Franke, Stefan Roth, and Bernt Schiele.
\newblock The {C}ityscapes dataset for semantic urban scene understanding.
\newblock In {\em CVPR}, 2016.

\bibitem{DeVries2017:arxiv}
Terrance DeVries and Graham~W Taylor.
\newblock Improved regularization of convolutional neural networks with cutout.
\newblock {\em arXiv}, abs/1708.04552, 2017.

\bibitem{Dong2020:ECCV}
Jiahua Dong, Yang Cong, Gan Sun, Yuyang Liu, and Xiaowei Xu.
\newblock {CSCL}: Critical semantic-consistent learning for unsupervised domain
  adaptation.
\newblock In {\em ECCV}, 2020.

\bibitem{He2020:CVPR}
Kaiming He, Haoqi Fan, Yuxin Wu, Saining Xie, and Ross Girshick.
\newblock Momentum contrast for unsupervised visual representation learning.
\newblock In {\em CVPR}, 2020.

\bibitem{he2020segmentations}
Yang He, Shadi Rahimian, Bernt Schiele, and Mario Fritz.
\newblock Segmentations-leak: Membership inference attacks and defenses in
  semantic image segmentation.
\newblock In {\em ECCV}, pages 519--535. Springer, 2020.

\bibitem{Hoffman2018:ICLR}
Judy Hoffman, Eric Tzeng, Taesung Park, Jun-Yan Zhu, Phillip Isola, Kate
  Saenko, Alexei Efros, and Trevor Darrell.
\newblock Cycada: Cycle-consistent adversarial domain adaptation.
\newblock In {\em ICML}, 2018.

\bibitem{Huang2018:ECCV}
Xun Huang, Ming-Yu Liu, Serge Belongie, and Jan Kautz.
\newblock Multimodal unsupervised image-to-image translation.
\newblock In {\em ECCV}, 2018.

\bibitem{Kang2020:NeurIPS}
Guoliang Kang, Yunchao Wei, Yi Yang, Yueting Zhuang, and Alexander~G.
  Hauptmann.
\newblock Pixel-level cycle association: {A} new perspective for domain
  adaptive semantic segmentation.
\newblock In {\em NeurIPS}, 2020.

\bibitem{Khosla2020:NeurIPS}
Prannay Khosla, Piotr Teterwak, Chen Wang, Aaron Sarna, Yonglong Tian, Phillip
  Isola, Aaron Maschinot, Ce Liu, and Dilip Krishnan.
\newblock Supervised contrastive learning.
\newblock In {\em NeurIPS}, 2020.

\bibitem{Kim2020:CVPR}
Myeongjin Kim and Hyeran Byun.
\newblock Learning texture invariant representation for domain adaptation of
  semantic segmentation.
\newblock In {\em CVPR}, 2020.

\bibitem{BDL}
Yunsheng Li, Lu Yuan, and Nuno Vasconcelos.
\newblock Bidirectional learning for domain adaptation of semantic
  segmentation.
\newblock In {\em ICCV}, 2019.

\bibitem{Liu2021:arxiv}
Weizhe Liu, David Ferstl, Samuel Schulter, Lukas Zebedin, Pascal Fua, and
  Christian Leistner.
\newblock Domain adaptation for semantic segmentation via patch-wise
  contrastive learning.
\newblock {\em arXiv}, abs/2104.11056, 2021.

\bibitem{liu2021bapa}
Yahao Liu, Jinhong Deng, Xinchen Gao, Wen Li, and Lixin Duan.
\newblock Bapa-net: Boundary adaptation and prototype alignment for
  cross-domain semantic segmentation.
\newblock In {\em ICCV}, 2021.

\bibitem{Lv2020}
Fengmao Lv, Tao Liang, Xiang Chen, and Guosheng Lin.
\newblock Cross-domain semantic segmentation via domain-invariant interactive
  relation transfer.
\newblock In {\em CVPR}, 2020.

\bibitem{Ma2021:CVPR}
Haoyu Ma, Xiangru Lin, Zifeng Wu, and Yizhou Yu.
\newblock Coarse-to-fine domain adaptive semantic segmentation with photometric
  alignment and category-center regularization.
\newblock In {\em CVPR}, 2021.

\bibitem{Marsden2021:arxiv}
Robert~A. Marsden, Alexander Bartler, Mario D{\"{o}}bler, and Bin Yang.
\newblock Contrastive learning and self-training for unsupervised domain
  adaptation in semantic segmentation.
\newblock {\em arXiv}, abs/2105.02001, 2021.

\bibitem{mcinnes2020umap}
Leland McInnes, John Healy, and James Melville.
\newblock Umap: Uniform manifold approximation and projection for dimension
  reduction.
\newblock {\em arXiv}, abs/1802.03426, 2018.

\bibitem{Mei2020:ECCV}
Ke Mei, Chuang Zhu, Jiaqi Zou, and Shanghang Zhang.
\newblock Instance adaptive self-training for unsupervised domain adaptation.
\newblock In {\em ECCV}, 2020.

\bibitem{Musto2020:BMVC}
Luigi Musto and Andrea Zinelli.
\newblock Semantically adaptive image-to-image translation for domain
  adaptation of semantic segmentation.
\newblock In {\em BMVC}, 2020.

\bibitem{neuhold2017mapillary}
Gerhard Neuhold, Tobias Ollmann, Samuel Rota~Bulo, and Peter Kontschieder.
\newblock The mapillary vistas dataset for semantic understanding of street
  scenes.
\newblock In {\em ICCV}, 2017.

\bibitem{intra_domain}
Fei Pan, Inkyu Shin, Francois Rameau, Seokju Lee, and In~So Kweon.
\newblock Unsupervised intra-domain adaptation for semantic segmentation
  through self-supervision.
\newblock In {\em CVPR}, 2020.

\bibitem{gta5}
Stephan~R Richter, Vibhav Vineet, Stefan Roth, and Vladlen Koltun.
\newblock Playing for data: Ground truth from computer games.
\newblock In {\em ECCV}, 2016.

\bibitem{synthia}
German Ros, Laura Sellart, Joanna Materzynska, David Vazquez, and Antonio~M
  Lopez.
\newblock The synthia dataset: A large collection of synthetic images for
  semantic segmentation of urban scenes.
\newblock In {\em CVPR}, 2016.

\bibitem{Subhani2020:ECCV}
M~Naseer Subhani and Mohsen Ali.
\newblock Learning from scale-invariant examples for domain adaptation in
  semantic segmentation.
\newblock In {\em ECCV}, 2020.

\bibitem{Tang2012:NIPS}
Kevin~D. Tang, Vignesh Ramanathan, Li Fei{-}Fei, and Daphne Koller.
\newblock Shifting weights: Adapting object detectors from image to video.
\newblock In {\em NIPS}, 2012.

\bibitem{Tarvainen2017:NIPS}
Antti Tarvainen and Harri Valpola.
\newblock Mean teachers are better role models: Weight-averaged consistency
  targets improve semi-supervised deep learning results.
\newblock In {\em NIPS}, 2017.

\bibitem{truong2021bimal}
Thanh-Dat Truong, Chi~Nhan Duong, Ngan Le, Son~Lam Phung, Chase Rainwater, and
  Khoa Luu.
\newblock Bimal: Bijective maximum likelihood approach to domain adaptation in
  semantic scene segmentation.
\newblock In {\em ICCV}, 2021.

\bibitem{Tsai2018:CVPR}
Yi{-}Hsuan Tsai, Wei{-}Chih Hung, Samuel Schulter, Kihyuk Sohn, Ming{-}Hsuan
  Yang, and Manmohan Chandraker.
\newblock Learning to adapt structured output space for semantic segmentation.
\newblock In {\em CVPR}, 2018.

\bibitem{Tsai2019:ICCV}
Yi{-}Hsuan Tsai, Kihyuk Sohn, Samuel Schulter, and Manmohan Chandraker.
\newblock Domain adaptation for structured output via discriminative patch
  representations.
\newblock In {\em ICCV}, 2019.

\bibitem{Van2021:ICCV}
Wouter Van~Gansbeke, Simon Vandenhende, Stamatios Georgoulis, and Luc Van~Gool.
\newblock Unsupervised semantic segmentation by contrasting object mask
  proposals.
\newblock In {\em ICCV}, 2021.

\bibitem{Vu2019:CVPR}
Tuan{-}Hung Vu, Himalaya Jain, Maxime Bucher, Matthieu Cord, and Patrick
  P{\'{e}}rez.
\newblock {ADVENT:} adversarial entropy minimization for domain adaptation in
  semantic segmentation.
\newblock In {\em CVPR}, 2019.

\bibitem{Wang2020:ECCV}
Haoran Wang, Tong Shen, Wei Zhang, Ling{-}Yu Duan, and Tao Mei.
\newblock Classes matter: {A} fine-grained adversarial approach to cross-domain
  semantic segmentation.
\newblock In {\em ECCV}, 2020.

\bibitem{Wang2021:AAAI}
Kaihong Wang, Chenhongyi Yang, and Margrit Betke.
\newblock Consistency regularization with high-dimensional non-adversarial
  source-guided perturbation for unsupervised domain adaptation in
  segmentation.
\newblock In {\em AAAI}, 2021.

\bibitem{Wang2021:ICCV}
Wenguan Wang, Tianfei Zhou, Fisher Yu, Jifeng Dai, Ender Konukoglu, and Luc
  Van~Gool.
\newblock Exploring cross-image pixel contrast for semantic segmentation.
\newblock In {\em ICCV}, 2021.

\bibitem{Wang2021:CVPRa}
Xinlong Wang, Rufeng Zhang, Chunhua Shen, Tao Kong, and Lei Li.
\newblock Dense contrastive learning for self-supervised visual pre-training.
\newblock In {\em CVPR}, 2021.

\bibitem{wang2021uncertainty}
Yuxi Wang, Junran Peng, and ZhaoXiang Zhang.
\newblock Uncertainty-aware pseudo label refinery for domain adaptive semantic
  segmentation.
\newblock In {\em ICCV}, 2021.

\bibitem{Wang2020:CVPR}
Zhonghao Wang, Mo Yu, Yunchao Wei, Rog{\'{e}}rio Feris, Jinjun Xiong, Wen{-}Mei
  Hwu, Thomas~S. Huang, and Honghui Shi.
\newblock Differential treatment for stuff and things: {A} simple unsupervised
  domain adaptation method for semantic segmentation.
\newblock In {\em CVPR}, 2020.

\bibitem{Wei2020:arxiv}
Longhui Wei, Lingxi Xie, Jianzhong He, Jianlong Chang, Xiaopeng Zhang, Wengang
  Zhou, Houqiang Li, and Qi Tian.
\newblock Can semantic labels assist self-supervised visual representation
  learning?
\newblock {\em arXiv}, abs/2011.08621, 2020.

\bibitem{dagan3}
Zuxuan Wu, Xintong Han, Yen-Liang Lin, Mustafa Gokhan~Uzunbas, Tom Goldstein,
  Ser Nam~Lim, and Larry~S Davis.
\newblock Dcan: Dual channel-wise alignment networks for unsupervised scene
  adaptation.
\newblock In {\em ECCV}, 2018.

\bibitem{Xiao2021:ICCV}
Tete Xiao, Colorado~J. Reed, Xiaolong Wang, Kurt Keutzer, and Trevor Darrell.
\newblock Region similarity representation learning.
\newblock In {\em ICCV}, 2021.

\bibitem{Xie2021:CVPR}
Zhenda Xie, Yutong Lin, Zheng Zhang, Yue Cao, Stephen Lin, and Han Hu.
\newblock Propagate yourself: Exploring pixel-level consistency for
  unsupervised visual representation learning.
\newblock In {\em CVPR}, 2021.

\bibitem{Yang2020:ECCV}
Jinyu Yang, Weizhi An, Sheng Wang, Xinliang Zhu, Chaochao Yan, and Junzhou
  Huang.
\newblock Label-driven reconstruction for domain adaptation in semantic
  segmentation.
\newblock In {\em ECCV}, 2020.

\bibitem{Yang2021:WACV}
Jinyu Yang, Weizhi An, Chaochao Yan, Peilin Zhao, and Junzhou Huang.
\newblock Context-aware domain adaptation in semantic segmentation.
\newblock In {\em WACV}, 2021.

\bibitem{Yang2021:ICCV}
Jinyu Yang, Chunyuan Li, Weizhi An, Hehuan Ma, Yuzhi Guo, Yu Rong, Peilin Zhao,
  and Junzhou Huang.
\newblock Exploring robustness of unsupervised domain adaptation in semantic
  segmentation.
\newblock In {\em ICCV}, 2021.

\bibitem{Yang2020:CVPRb}
Yanchao Yang, Dong Lao, Ganesh Sundaramoorthi, and Stefano Soatto.
\newblock Phase consistent ecological domain adaptation.
\newblock In {\em CVPR}, 2020.

\bibitem{Yang2020:CVPRa}
Yanchao Yang and Stefano Soatto.
\newblock {FDA}: Fourier domain adaptation for semantic segmentation.
\newblock In {\em CVPR}, 2020.

\bibitem{Zhang2021:CVPR}
Pan Zhang, Bo Zhang, Ting Zhang, Dong Chen, Yong Wang, and Fang Wen.
\newblock Prototypical pseudo label denoising and target structure learning for
  domain adaptive semantic segmentation.
\newblock In {\em CVPR}, 2021.

\bibitem{CAG}
Qiming Zhang, Jing Zhang, Wei Liu, and Dacheng Tao.
\newblock Category anchor-guided unsupervised domain adaptation for semantic
  segmentation.
\newblock In {\em NeurIPS}, 2019.

\bibitem{Zhang2020:NeurIPS}
Xiao Zhang and Michael Maire.
\newblock Self-supervised visual representation learning from hierarchical
  grouping.
\newblock In {\em NeurIPS}, 2020.

\bibitem{Zhang2020:CVPR}
Yiheng Zhang, Zhaofan Qiu, Ting Yao, Chong{-}Wah Ngo, Dong Liu, and Tao Mei.
\newblock Transferring and regularizing prediction for semantic segmentation.
\newblock In {\em CVPR}, 2020.

\bibitem{pspnet}
Hengshuang Zhao, Jianping Shi, Xiaojuan Qi, Xiaogang Wang, and Jiaya Jia.
\newblock Pyramid scene parsing network.
\newblock In {\em CVPR}, 2017.

\bibitem{Zhao2021:ICCV}
Xiangyun Zhao, Raviteja Vemulapalli, Philip Mansfield, Boqing Gong, Bradley
  Green, Lior Shapira, and Ying Wu.
\newblock Contrastive learning for label-efficient semantic segmentation.
\newblock In {\em ICCV}, 2021.

\bibitem{Zheng2021:IJCV}
Zhedong Zheng and Yi Yang.
\newblock Rectifying pseudo label learning via uncertainty estimation for
  domain adaptive semantic segmentation.
\newblock {\em IJCV}, 129(4):1106--1120, 2021.

\bibitem{Zhu2017:ICCV}
Jun-Yan Zhu, Taesung Park, Phillip Isola, and Alexei~A Efros.
\newblock Unpaired image-to-image translation using cycle-consistent
  adversarial networks.
\newblock In {\em ICCV}, 2017.

\bibitem{Zoph2020:NeurIPS}
Barret Zoph, Golnaz Ghiasi, Tsung-Yi Lin, Yin Cui, Hanxiao Liu, Ekin~D Cubuk,
  and Quoc~V Le.
\newblock Rethinking pre-training and self-training.
\newblock In {\em NeurIPS}, 2020.

\bibitem{Zou2018:ECCV}
Yang Zou, Zhiding Yu, B.~V. K.~Vijaya Kumar, and Jinsong Wang.
\newblock Unsupervised domain adaptation for semantic segmentation via
  class-balanced self-training.
\newblock In {\em ECCV}, 2018.

\end{thebibliography}
}

\end{document}